\documentclass[conference]{IEEEtran}

\usepackage[utf8]{inputenc} 
\usepackage[T1]{fontenc}    
\usepackage{hyperref}       
\usepackage{url}            
\usepackage{booktabs}       
\usepackage{amsfonts}       
\usepackage{nicefrac}       
\usepackage{microtype}      
\usepackage{graphicx}
\usepackage{subcaption}
\usepackage{booktabs} 
\usepackage{amsmath, amsthm, amssymb}
\usepackage{algorithm}
\usepackage{algorithmic}
\usepackage{tabularx}
\usepackage{multirow}
\usepackage{bm}

\usepackage{enumitem}

\usepackage[export]{adjustbox}
\usepackage{xspace}
\newcommand{\mypar}[1]{\medskip\noindent\textbf{#1}\xspace}

\newcommand{\citep}{\cite}

\begin{document}

\def\fix{linear transfer learning}
\def\e2e{end-to-end transfer learning}
\def\sys{Bullseye Polytope}
\def\sysshortThree{BP-3x}
\def\sysshortFive{BP-5x}
\def\sysshortOne{BP}

\def\cpHoursSingleTransfer{\(\sim\)17}
\def\cpMinsSingleTransfer{1,002}
\def\spMinsSingleTransfer{47}
\def\threespMinsSingleTransfer{88}
\def\fivespMinsSingleTransfer{141}
\def\spFasterCPSingleTransfer{21} 
\def\threespFasterCPSingleTransfer{11} 
\def\fivespFasterCPSingleTransfer{7} 
\def\spImproveSingleTransfer{4.32\%}
\def\threespImproveSingleTransfer{7.44\%}
\def\fivespImproveSingleTransfer{8.38\%}

\def\spImproveSingleTransferToDiffTrainingSet{5.82\%}
\def\threespImproveSingleTransferToDiffTrainingSet{8.56\%}
\def\fivespImproveSingleTransferToDiffTrainingSet{9.27\%}

\def\cpHoursSingleEnd{\(\sim\)20}
\def\cpMinsSingleEnd{1,180}
\def\spMinsSingleEnd{15}
\def\threespMinsSingleEnd{98}
\def\spFasterCPSingleEnd{36} 
\def\threespFasterCPSingleEnd{12} 
\def\spImproveSingleEnd{18.25\%}
\def\threespImproveSingleEnd{26.75\%}
\def\oneToThreeImproveSingleEnd{8.5\%}

\def\cpMinsMultiTransfer{421}
\def\spMinsMultiTransfer{27}
\def\threespMinsMultiTransfer{22}
\def\fivespMinsMultiTransfer{45}
\def\spFasterCPMultiTransfer{16} 

\def\spFasterCPGeneral{10-36}

\def\spAttackAccMultipleTransferValidation{49.56\%} 
\def\spAttackAccMultipleEndValidation{31.17\%} 

\newcommand\norm[1]{\left\lVert#1\right\rVert}
\def\phiPoisons{\space \(\{\phi(x^{(j)}_{p})\}_{j=1}^k\)}
\def\phiBases{\space \(\{\phi(x^{(j)}_{b})\}_{j=1}^k\)}
\def\phiTarget{\space \(\phi(x_t)\)}

\def\knndef{Deep k-NN}
\def\l2def{\(l_2\textrm{-norm}\) centroid}

\newcolumntype{C}[1]{>{\hsize=#1\hsize\centering\arraybackslash}X}
\newcolumntype{L}[1]{>{\hsize=#1\hsize\arraybackslash}X}
\newcolumntype{R}[1]{>{\hsize=#1\hsize\RaggedLeft\arraybackslash}X}

\thispagestyle{plain}
\pagestyle{plain}

\title{\sys{}: A Scalable Clean-Label Poisoning Attack with Improved Transferability}

\author{
\IEEEauthorblockN{Hojjat Aghakhani,
Dongyu Meng,
Yu-Xiang Wang,
Christopher Kruegel,
and Giovanni Vigna
}
\IEEEauthorblockA{University of California, Santa Barbara}
\IEEEauthorblockA{\{hojjat, dmeng, yuxiangw, chris, vigna\}@cs.ucsb.edu}
}

\maketitle

\newtheorem{prop}{Proposition}

\begin{abstract}
A recent source of concern for the security of neural networks is the emergence of clean-label dataset poisoning attacks, wherein correctly labeled poison samples are injected into the training dataset.
While these poison samples look legitimate to the human observer, they contain malicious characteristics that trigger a targeted misclassification during inference.
We propose a scalable and transferable clean-label poisoning attack against transfer learning, which creates poison images with their center close to the target image in the feature space.
Our attack, \sys{}, improves the attack success rate of the current state-of-the-art by 26.75\% in end-to-end transfer learning, while increasing attack speed
by a factor of 12.
We further extend \sys{} to a more practical attack model by including multiple images of the same object (e.g., from different angles) when crafting the poison samples.
We demonstrate that this extension improves attack transferability by over 16\% to unseen images (of the same object) without using extra poison samples.\footnote{Accepted at \texttt{EuroS\&P 2021}.}
\end{abstract}

\section{Introduction}
\label{sec:intro}
Machine-learning-based systems are being increasingly deployed in security-critical applications, such as face recognition~\citep{parkhi2015deep, sun2014deep}, fingerprint identification~\citep{wang2014fingerprint}, and cybersecurity~\citep{suciu2018does}, as well as applications with a high cost of failure such as autonomous driving~\citep{chen2015deepdriving}. 
The possibility of generating adversarial examples in deep neural networks has raised serious doubt on the security of these systems~\citep{goodfellow2014explaining, biggio2013evasion, szegedy2013intriguing}.
In these \emph{evasion} attacks, a targeted input is perturbed by imperceptible amounts at test time to trigger misclassification by a trained network.
But neural networks are also vulnerable to malicious manipulation during the \textit{training} process.
As neural networks require large datasets for training, it is common practice to use training samples collected from other, often untrusted, sources (e.g., the Internet), and it is expensive to have these datasets carefully vetted. 
While neural networks are strong enough to learn powerful models in the presence of \emph{natural} noise, they are vulnerable to carefully crafted \emph{malicious} noise introduced deliberately by adversaries. 
In particular, gathering data from untrusted sources makes neural networks susceptible to \emph{data poisoning attacks}, where an adversary injects data into the training set to manipulate or degrade the system performance.


\begin{figure}[t]
	\begin{subfigure}{0.32\linewidth}
		\centering
        \includegraphics[width=0.7\linewidth]{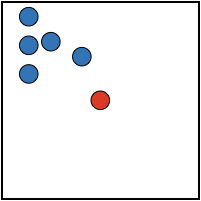}
        \vspace{-0.1cm}
		\caption{Original images.}
    \end{subfigure}
    \begin{subfigure}{0.32\linewidth}
    	\centering
        \includegraphics[width=0.7\linewidth]{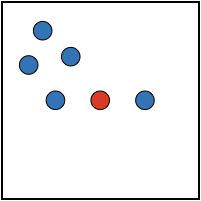}
        \vspace{-0.1cm}
		\caption{Convex Polytope}
    \end{subfigure}
    \begin{subfigure}{0.32\linewidth}
    	\centering
        \includegraphics[width=0.7\linewidth]{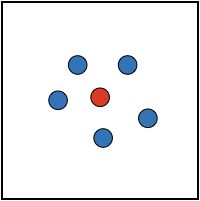}
        \vspace{-0.1cm}
		\caption{\sys{}}
    \end{subfigure}
    \hfill
    \vspace{-0.1cm}
    \caption{Simplified representation of poison samples in a two-dimensional feature space. The blue circles are poison samples and the red circle is the target. Convex Polytope moves poison samples until the target is inside their convex hull, making no further refinements to move the target away from the polytope boundary, whereas \sys{} enforces that the target resides close to the center.}
	\label{fig:example}
	\vspace{-0.4cm}
\end{figure}

Our work focuses on \emph{clean-label poisoning attacks}, a branch of poisoning attacks wherein the attacker does not have any control over the labeling process.
In this threat model, the poison samples are created by introducing imperceptible (yet malicious) alterations that will result in model misbehavior in response to specific target inputs.
These perturbations are small enough to justify the original images' labels in the eye of a domain expert.
The stealth of the attack increases its success rate in real-world scenarios compared to other types of data-poisoning attacks, as the poison data (1) will not be identified by human labelers, and (2) does not degrade test accuracy except for misclassification of particular target samples.

Clean-label poisoning on \emph{transfer learning} was first studied in a \emph{white-box} setting~\cite{shafahi2018poison}, where the attacker leverages complete knowledge of the pre-trained network \(\phi\) that the victim employs to either (1) extract features for training a (linear) classifier (\emph{\fix{}}) or (2) fine-tune on a similar task (\emph{\e2e{}}).
The \emph{Feature Collision} attack~\cite{shafahi2018poison} selects a base image \(x_b\) from the intended misclassification class, and creates a poison sample, \(x_p\), by adding small (bounded) adversarial perturbations to \(x_b\) that brings it close to the target image \(x_t\) in the feature space, i.e., \(\phi(x_t) \approx \phi(x_p) \). This triggers misclassification of \(x_t\) to the targeted class by any linear classifier that is trained on the features of a dataset containing \(x_p\).
This approach fails when the feature extractor \(\phi\) is unknown to the attacker.
To mitigate such limitation, Zhu et al. proposed \emph{Convex Polytope}~\cite{zhu2019transferable}, which, instead of finding poison samples close to the target, finds a set of poison samples that form a convex polytope around it, increasing the probability that the target lies within (or at least close to) this ``attack zone'' in the victim's feature space.
Convex Polytope relies on the fact that every \textit{linear} classifier that classifies a set of points into label \(l\) will classify every point in the convex hull of these points as label \(l\).

As we will show later, Convex Polytope suffers from one inherent flaw. 
The target feature vector tends to be close to the boundary of the attack zone, potentially hampering the attack transferability.
Furthermore, the Convex Polytope algorithm is very slow.
For example, crafting a set of five poison samples for a single target takes \cpHoursSingleTransfer{} GPU-hours on average. 

To address these limitations, we propose \sys{}, which refines the constraints of Convex Polytope such that the target is pushed toward the ``center'' of the attack zone (i.e., the convex hull of poison samples).
The geometrical comparison of \sys{} and Convex Polytope is shown in Figure~\ref{fig:example}.
\sys{} improves both the transferability and speed of the attack. 
When the victim adopts \fix{}, our method improves the attack success rate by \threespImproveSingleTransfer{} on average, while being \threespFasterCPSingleTransfer{}x faster.
In \e2e{}, \sys{} outperforms Convex Polytope by \threespImproveSingleEnd{} on average, while being \threespFasterCPSingleEnd{}x faster.
For some victim models, the attack success rate of \sys{} is \(\sim50\%\) higher than Convex Polytope.
In a weaker threat model, where the adversary has limited knowledge of the training set of the victim's feature extractor \(\phi\), \sys{} provides a \fivespImproveSingleTransferToDiffTrainingSet{} higher attack success rate in \fix{}.

We also extend \sys{} to a more \textit{practical} threat model. 
Current clean-label poisoning attacks are designed to target only one image at a time, rendering them ineffective against unpredictable variations in real-world image acquisition.
Such attacks disregard the following major point: to succeed in real-world scenarios, the attack needs to cope with a spectrum of test inputs.
By including a larger number of target images (of the same object) when crafting the poison samples, we are able to obtain an attack transferability of \spAttackAccMultipleTransferValidation{} against \emph{unseen} images (of the same object), 
without increasing the number of poison samples. 
This is an improvement  of over 16\%, compared to the single-target mode, when testing against the same set of images (in linear transfer learning).

We further evaluate \sys{} against \l2def{} and \knndef{} defenses~\cite{peri2020deep}, which are shown to be effective against poisoning attacks on transfer learning.
These defenses employ neighborhood conformity tests to sanitize the training data.
Our evaluation shows that \sys{} is much more resilient than Convex Polytope against less aggressive defense configurations.
To completely mitigate the attacks, \knndef{} and \l2def{} defenses need to remove 5\% and 10\% of the training data, respectively, of which 1\% are the poison data.
We show that increasing the number of poison samples makes the \l2def{} defense completely ineffective, as it needs to aggressively prune the dataset, which, in turn, degrades the model's performance. 
This gives our attack a major advantage, as, unlike Convex Polytope, \sys{} can incorporate more poison samples into the attack process, with virtually no cost in attack-execution time.
As we will show later, Convex Polytope scales poorly as the number of poison samples increases.
In particular, running the Convex Polytope attack for 800 iterations to craft ten poison samples takes 603 minutes on GPU, while Bullseye Polytope takes only seven minutes.

The \knndef{} defense is able to completely mitigate the attack by increasing the neighborhood size until poison samples cannot become a majority, but it suffers from low detection precision (20\%). 
On the other hand, if the number of poison samples is larger than the number of samples in the target object's original class, the majority test can be overwhelmed, leaving many poison samples undetected. 
Furthermore, in some applications, the target object does not belong to one of the classes in the training set, but rather is an unclassified object that the adversary aims to ``smuggle in.'' 
In this case, poison samples are not likely to have nearby neighbors in the fine-tuning set from a single class other than the poison class.
Therefore, to fully mitigate the attack, the \knndef{} defense needs to adopt a much larger neighborhood size, which results in discarding a higher number of clean samples.


Concurrent to our work, a recent study was published on arXiv~\cite{schwarzschild2020just}. That study develops standardized benchmarks for data poisoning and backdoor attacks to promote fair comparison.
Interestingly, the authors already include our work as presented in this paper. The results for linear transfer learning settings demonstrate that \sys{} outperforms all other attacks.
Especially in the white-box setting, the independent third-party study showed that our attack achieved more than 50\% higher success rates across experiments compared to the runner-up.
The study also benchmarks {\it from-scratch training scenarios}, where the victim's network is trained from random initialization on the poisoned dataset.
This is a much more challenging scenario for attacks that are designed for transfer learning settings (like \sys{}). However, it is a scenario that is specifically taken into consideration by another attack, \textit{Witches' Brew} (WiB)~\cite{geiping2020witches}, which was also recently published on arXiv (and parallel to this work).
The from-scratch benchmarks are evaluated on two datasets: CIFAR-10~\cite{krizhevsky2012imagenet} and TinyImageNet~\cite{le2015tiny}.
On the former dataset, WiB demonstrated a success rate of 26\%, while all other attacks (including \sys{}) succeeded less than 3\% of the time.
Interestingly, however, for the TinyImageNet benchmark, our attack achieved the highest success rate (44\%), 12\% higher than the runner-up (WiB), while other attacks failed most of the times.

To some readers, \sys{} might appear as a simple extension of prior work, such as Convex Polytope.
We argue that this would be myopic — compelling ideas often appear simple in hindsight.
Our experiments show that \sys{} is not only more successful than current state-of-the-art poisoning attacks on transfer learning, but, perhaps more importantly, it is also an order of magnitude faster.
This performance improvement is significant, as it unlocks our practical ability to build defenses against this class of attacks with higher detection precision.
When creating solutions to detect poisoning attacks, researchers have to experiment with ideas and parameters and perform statistical evaluations.
These experiments take a significant amount of time, even when deploying substantial amounts of resources in the cloud. The proposed technique in this paper cuts down this time by a factor of 10, enabling a much faster cycle of experimentation.
We also make all source code as well as poison samples available, which can be found at \texttt{github.com/ucsb-seclab/BullseyePoison}.

\section{Threat Model}
\label{sec:threat}
In our threat model, we assume that the victim employs transfer learning, where a model trained for one task is reused as part of a different model for a second task.
Transfer learning is shown to be a common practice, as it obtains high-quality models without incurring the cost of training a model from scratch~\cite{gu2017badnets}.
We consider two transfer learning approaches that the victim may adopt; \emph{\fix{}} and \emph{\e2e{}}.
In the former, a pre-trained but \textbf{frozen} network acts as a feature extractor \(\phi\), and an application-specific \textbf{linear} classifier is fine-tuned on \(\phi(\Gamma)\), where \(\Gamma\) is the \emph{fine-tuning training set}.
In \e2e{}, the feature extractor and linear classifier are trained jointly on \(\Gamma\), and, therefore, the feature extractor is altered during fine-tuning.
In both scenarios, the attacker injects a small number of poison samples into \(\Gamma\), obtained by imperceptibly perturbing some of the original samples.
The attacker does not have any control over the labeling process, therefore, the poison samples remain correctly labeled according to their original class.
We consider both \emph{black-box} and \emph{gray-box} settings.
The attacker has no access to the victim model in the black-box setting. 
In the gray-box setting, only the victim network's architecture is known.
We assume that the attacker knows the training set that is used to build \(\phi\).\footnote{Note that the attacker has no knowledge of \(\Gamma\) (other than the added poisons).}
The attacker uses this training set for training \emph{substitute networks}, which will be used to craft poison samples.
Unless explicitly stated, by ``attack transferability'' we mean the transferability of the poison samples' characteristics (i.e., targeted misclassification) to the victim's (fine-tuned) model.
We do further evaluation in more limited settings where the adversary has no or partial knowledge of the \textit{training set} of \(\phi\).


\section{Related Work}
\label{sec:relatedwork}
\mypar{Data Poisoning Attacks.} A well-studied portion of data-poisoning attacks aims to use malicious data to degrade the test accuracy of a model~\cite{nelson2008exploiting, biggio2012poisoning, xiao2012adversarial, mei2015using, burkard2017analysis}.
While such attacks are shown to be successful, they are easy to detect, as the performance of a model can always be assessed by testing the model on a private, trusted set of samples.
Another important branch of data-poisoning attacks, known as \emph{backdoor attacks}~\cite{gu2017badnets}, fools models by imprinting a small number of training examples with a specific pattern (\emph{trigger}) and changing their labels to a different target label. 
During inference, the attacker achieves misclassification by injecting the trigger into targeted examples.
This strategy relies on the assumption that the labels of the poison data will not be inspected.
To avoid injecting wrong labels, clean-label~\cite{turner2018clean} and hidden-trigger~\cite{saha2020hidden} backdoor attacks are proposed, where poison samples are crafted with optimization procedures.
In general, similar to evasion attacks, backdoor attacks present the following shortcoming: they require the modification of test samples during inference to enable misclassification.

\mypar{Clean-label Poisoning Attacks.} A recent branch of data-poisoning attacks has no control over the labeling process.
The first clean-label poisoning attack is Feature Collision~\cite{shafahi2018poison}, which mainly targets linear transfer learning, where the adversary has complete knowledge of the feature extractor network \(\phi\) employed by the victim.
Feature Collision suffers from one major problem; it tends to fail in black-box settings~\cite{zhu2019transferable}.
To mitigate such limitations, Zhu et al. proposed the Convex Polytope attack~\cite{zhu2019transferable}, which crafts a set of poison samples that contain the target's feature vector within their convex hull.
In particular, this attack outperforms Feature Collision by 20\% on average in terms of success rate. 
As we will show in Section~\ref{sec:experiments}, Convex Polytope suffers from two shortcomings;
(1) \emph{Speed}: Convex Polytope is significantly slow.
(2) \emph{Robustness}: The target's feature vector tends to be close to the boundary of the polytope formed by the poison samples, leaving the full potential for attack transferability untapped.

To mitigate such limitations, we design \sys{} by crafting poison samples centered around the target image in the feature space.
As we will show later, our attack accelerates poison construction by an order of magnitude compared to Convex Polytope, while achieving higher attack success rates in both transfer learning setups.
We further improve the attack robustness by incorporating multiple images of a target object. 
Current clean-label poisoning attacks are designed to target only one image at a time, rendering  them  ineffective against unpredictable variations in real-world image acquisition.
We show that the resulting attack is effective on \textit{unseen} images of the target while maintaining good baseline test accuracy on non-targeted images. 
To the best of our knowledge, \sys{} is the first clean-label poisoning attack being proposed for a multi-target threat model, which is an important feature for practical implementations on real-world systems.

Concurrent to our work, a recent paper~\cite{geiping2020witches} -- published on arXiv -- proposed a clean-label poisoning attack, named WiB, against from-scratch training scenarios, where the victim's model is trained from random initialization on the poisoned dataset.
Such a setting is more challenging for previous clean-label poisoning attacks and \sys{}, as they are designed for transfer learning scenarios.
However, as we will show later, our attack outperforms WiB in some experiments.
This is quite interesting, as unlike WiB, \sys{} is not originally designed for from-scratch training scenarios.  


\mypar{Defenses Against Clean-label Poisoning.}
Parallel to this work, a recent study by Peri et al.~\cite{peri2020deep} proposed defenses against clean-label poisoning attacks, i.e., Feature Collision~\cite{shafahi2018poison} and Convex Polytope~\cite{zhu2019transferable}. 
They adopted defenses that are shown to be effective against both evasion and backdoor attacks~\cite{papernot2018deep, steinhardt2017certified}.\footnote{A detailed discussion of defenses against evasion and backdoor attacks is provided in the Appendix~\ref{appendix:defensesRelatedWork}.}
For the Feature Collision attack, they observed that a \knndef{} based method applied to the penultimate layer (i.e., the feature layer) of the neural network outperforms other types of defenses, such as adversarial training or \l2def{} defenses.
In the Convex Polytope attack, \knndef{} and \l2def{} defenses demonstrate comparable resilience, however, the \knndef{} defense removes fewer clean samples from the training data.
In this work, we evaluate \sys{} and Convex Polytope against both \knndef{} and \l2def{} defenses. 
As we will show in Section~\ref{sec:eval-defenses}, \sys{} is generally more robust than Convex Polytope against less aggressive defense configurations.


\section{Background}
\label{sec:bg}
As discussed earlier, Feature Collision fails when the victim's feature extractor \(\phi\) is unknown to the attacker. To mitigate such limitation, Zhu et al.~\cite{zhu2019transferable} proposed Convex Polytope (CP), which crafts a set of poison samples that contain the target within their convex hull.
CP exploits the following mathematical guarantee: if the victim's linear classifier associates the poison samples with the targeted class, it will label any point inside their convex hull as the targeted class.
CP creates a larger ``attack zone'' in the feature space, thus increasing the chance of transferability, as argued by the authors.
In particular, CP solves the following optimization problem:
\begin{align}
\underset{\{c^{(i)}\},\{x_p^{(j)}\}}{\textrm{minimize}} \; &\frac{1}{2m} \sum_{i=1}^m \frac{\norm{\phi^{(i)}(x_t) - \sum_{j=1}^{k} c_j^{(i)}\phi^{(i)}(x_p^{(j)})}^2}{\norm{\phi^{(i)}(x_t)}^2}\notag\\
\textrm{subject to} \; &\sum_{j=1}^k c_j^{(i)} = 1, c_j^{(i)} \geq 0, \forall i,j, \notag\\
&\norm{x_p^{(j)} - x_b^{(j)}}_{\infty} \leq \epsilon \:, \forall j,
\label{eq:cvxLoss}
\end{align}
where \(x_b^{(j)}\) is the original image of the \(j\)-th poison sample, and \(\epsilon\) determines the maximum allowed perturbation. 
Eq.~\ref{eq:cvxLoss} finds a set of poison samples \(\{x^{(j)}_{p}\}_{j=1}^k\) such that the target \(x_t\) lies inside, or at least close to, the convex hull of the poison samples in the feature spaces defined by \(m\) substitute networks \(\{\phi^{(i)}\}_{i=1}^{m}\).
In the \(i\)-th substitute network, the target feature vector \(\phi^{(i)}(x_t)\) is ideally a convex combination of the feature vectors of poison images, i.e.,  \(\phi^{(i)}(x_t)=\sum_{j=1}^{k} c_j^{(i)}\phi^{(i)}(x_p^{(j)})\), where \(c_j^{(i)}\) determines the \(j\)-th poison's coefficient. 
To solve the non-convex problem in Eq.~\ref{eq:cvxLoss} (i.e., find the optimal poison samples), CP repeats the following steps for 4,000 iterations:

\begin{enumerate}[leftmargin=1.5\parindent]
    \item Freezing \(\{x^{(j)}_{p}\}_{j=1}^k\), use forward-backward splitting~\cite{goldstein2014field} to optimize the coefficients for each individual network \(\{c^{(i)}\}\).
    \item Given \(\{c^{(i)}\}\), optimize \(\{x^{(j)}_{p}\}_{j=1}^k\) using one gradient step.
    \item Clip \(\{x^{(j)}_{p}\}_{j=1}^k\) to the \(\epsilon\)-ball around the base images \(\{x^{(j)}_{b}\}_{j=1}^k\).
\end{enumerate}

\mypar{Poor Scalability of Convex Polytope.}
We observed that when using 18 substitute networks, solving Eq.~\ref{eq:cvxLoss} for five poison samples takes \cpHoursSingleTransfer{} GPU-hours on average.\footnote{This is the exact same setting used in the original paper~\cite{zhu2019transferable}.}
Of this time, step one alone takes \(\sim\)15 hours.
We list the details of step one in the Appendix (Algorithm 1).
Within this process, we noticed two major time-consuming operations: (1) checking whether the new coefficients result in a smaller loss compared to the old coefficients (this is done in every iteration of coefficient optimization), and (2) projection onto the probability simplex, which happens whenever the new coefficients satisfy the above condition.
While we believe that there is room for improvement of this algorithm, e.g., by checking the condition every few steps rather than each step, we did not make any such changes in order to avoid degradation of the attack success rate, and to allow for a fair comparison.

\section{\sys{}}
\label{sec:method}

Apart from scalability, CP has an inherent flaw: as soon as the target crosses the boundary into the interior of the convex polytope, there is no incentive to refine further and move the target deeper inside the attack zone (Figure~\ref{fig:example}).
Therefore, the target will lie close to the boundary of the resulting poison polytope, which reduces robustness and generalizability.
We design \sys{} (\sysshortOne{}) based on the insight that, by fixing the relative position of the target with respect to the poison samples' convex hull, we speed up the attack while also improving its robustness.
Instead of searching for coefficients by optimization, which is neither efficient nor effective, \sysshortOne{} \textit{predetermines} the \(k\) coefficients as equal, i.e., \(\frac{1}{k}\), to enforce that the target resides close to the ``center'' of the poison samples' polytope.\footnote{Our notion of center coincides with the \textit{center of mass} of the poison set.}
\sysshortOne{} then solves the special case of:
\begin{align}
\underset{\{x_p^{(j)}\}}{\textrm{minimize}} \; &\frac{1}{2m} \sum_{i=1}^m \frac{\norm{\phi^{(i)}(x_t) - \frac{1}{k}\sum_{j=1}^{k} \phi^{(i)}(x_p^{(j)})}^2}{\norm{\phi^{(i)}(x_t)}^2}\notag\\
\textrm{subject to} \; & \norm{x_p^{(j)} - x_b^{(j)}}_{\infty} \leq \epsilon \:, \forall j.
\label{eq:simplecvxLoss}
\end{align}
As we show later, \sysshortOne{} indeed improves attack transferability by effectively pushing the target toward the center of the attack zone.
Also, by precluding the most time-consuming step of computing coefficients, \sysshortOne{} is an order of magnitude faster than CP.
It should be noted that, while \sysshortOne{} seems to be a special case of CP, the objective loss of Eq.~\ref{eq:simplecvxLoss} has a significant difference with respect to Eq.~\ref{eq:cvxLoss}.
That is, the closer the target gets to the polytope's center, the smaller the loss becomes, which is not true for Eq.~\ref{eq:cvxLoss}.
For this reason, the solution of Eq.~\ref{eq:simplecvxLoss} (\sysshortOne{}) is not necessarily a special case of Eq.~\ref{eq:cvxLoss} (CP), since an optimizer that uses Eq.~\ref{eq:cvxLoss} might never find such a solution.
Although CP initially sets the k coefficients as equals (i.e., \(\frac{1}{k}\)), we observed that the coefficients become skewed from the very beginning. 
This happens because at each step of optimizing the coefficients, the solution of Eq.~\ref{eq:cvxLoss} is skewed towards poison samples that are closer to the target.

\mypar{Kernel Embedding-view of \sys{}.}
Besides improved computational efficiency, our approach in Eq.~\ref{eq:simplecvxLoss} can be viewed as optimizing \emph{distribution} of poison samples via its mean embedding. Informally speaking, when $\phi$ is a sufficiently descriptive feature map,\footnote{For example, $\phi(x) = k(x,\cdot)$ for a \emph{characteristic} reproducing kernel $k$, e.g., the Gaussian-RBF kernel $k(x,y) = e^{-\|x-y\|^2}$.} then $\mathbb{E}_{x\sim P}[\phi(x)] = \mathbb{E}_{x\sim Q}[\phi(x)]$ \emph{if and only if} distributions \(P\) and \(Q\) are identical (see, e.g., ~\citep[Theorem 1]{smola2007hilbert}; also see a recent survey of kernel mean embedding ~\citep{muandet2017kernel}). 
Deep neural networks are closely related to kernel methods \citep{neal1996bayesian,rahimi2008random,jacot2018neural}. The pre-trained network is a powerful feature extractor, thus is often viewed as an even better descriptor of the input feature $x$ than kernels for prediction purposes. 
As a result, if  $\{x_p^{(j)}\}_{j=1}^{k}$ are drawn from a distribution $P$, then Eq.~\ref{eq:simplecvxLoss} is \emph{essentially} optimizing this distribution using the \emph{plug-in estimator} of mean embedding: $\frac{1}{k}\sum_{j=1}^{k} \phi(x_p^{(j)})$.

\mypar{Deep Sets.} \sys{} is also backed by the more recent approach of \emph{deep sets}~\citep{zaheer2017deep},  which establishes that for \emph{any} function $f$ of a set of poison samples $x_p^{(1)},...,x_p^{(k)}$ that enjoys \emph{permutation invariance} admits a decomposition: $f= \rho(\frac{1}{k}\sum_{j=1}^{k} \phi(x_p^{(j)}))$ for some function $\rho,\phi$. Notice that due to the random reshuffling steps in training machine learning models, the learned prediction function (i.e., classifier) is \emph{permutation invariant} by construction with respect to the training dataset (containing the set of poison samples). That is, the classifier's prediction \(f\) can be decomposed to $\rho(\frac{1}{k}\sum_{j=1}^{k} \phi(x_p^{(j)}))$ --- a function of the mean embedding.  Thus, our simplification from Eq.~\ref{eq:cvxLoss} to Eq.~\ref{eq:simplecvxLoss} that optimizes the mean embedding rather than a more general convex combination is arguably \emph{without loss of generality} (See Appendix~\ref{appendix:DeepSets} for a more detailed discussion). 



\subsection{Improved Transferability via Multi-Draw Dropout}
Attack transferability improves when we increase the number of substitute networks for crafting poison samples.
While it is impractical to ensemble a large number of networks due to memory and time constraints, introducing dropout randomization provides some of the diversification afforded by a larger ensemble.
With dropout, the substitute network \(\phi^{(i)}\) provides a different feature vector for the same poison sample each time.
This randomization was observed to result in a much higher variance in the (training) loss of Eq.~\ref{eq:simplecvxLoss} compared to that of Eq.~\ref{eq:cvxLoss}.
Since the solution space of Eq.~\ref{eq:simplecvxLoss} is much more restrictive than Eq.~\ref{eq:cvxLoss}, and moves around for different realizations of dropout, gradient descent has a harder time converging for Eq.~\ref{eq:simplecvxLoss}.
We use averaging over multiple draws to alleviate this problem. 
In each iteration, we compute the feature vector of poison samples \(R\) times for each network, and use their average in optimizing Eq.~\ref{eq:simplecvxLoss}. 
Of course, increasing \(R\) results in higher attack execution time, but even a modest choice of \(R\!=\!3\) is enough to achieve an \oneToThreeImproveSingleEnd{} higher success rate compared to when \(R\!=\!1\) is used for \e2e{}.
Even in this case, \sysshortOne{} is \threespFasterCPSingleEnd{} times faster than CP.

\subsection{Multi-target Mode} 
We further improve the robustness of \sysshortOne{} by incorporating multiple images of a target object.
This is similarly achieved by simply replacing $\phi(x_t)$ in \eqref{eq:simplecvxLoss} with a \emph{mean embedding of the distribution} of the targets $\frac{1}{N_{\mathrm{im}}}\sum_{j=1}^n\phi(x_t^{(j)})$ where $x_t^{(1)},...,x_t^{(N_{\mathrm{im}})}$ are drawn i.i.d from a target distribution that captures the natural variations in lighting conditions, observation angles, and other unpredictable stochasticity in real-world image acquisition.
To say it differently, instead of attacking one individual instance, we are now attacking a distribution of instances by creating a set of poison samples that \emph{match} the target distribution in terms of the mean embedding as much as possible. In Section~\ref{sec:eval-multi}, we demonstrate that the resulting attack is highly effective not only on the ``training'' instances of the targets but also \emph{generalizes} to 
\textit{unseen} images of the target
, while maintaining good baseline test accuracy on images of non-targeted objects.
In contrast, current clean-label poisoning attacks only work with one image at a time, rendering them ineffective in more realistic attack scenarios.

\subsection{End-to-End Transfer Learning}
In \e2e{}, the victim retrains both the feature extractor and the linear classifier, altering the feature space in the process.
This causes unpredictability in the attack zone, even in the white-box setting.
To tackle this issue, inspired by Zhu et al.~\cite{zhu2019transferable}, we jointly apply \sysshortOne{} to multiple layers of the network, crafting poison samples that satisfy Eq.~\ref{eq:simplecvxLoss} on the feature space created by each layer.
This adds to the complexity of the problem, which is especially problematic for the already slow CP algorithm.

\section{Experiments}
\label{sec:experiments}
\begin{figure*}
     \centering
     \vspace{-0.4cm}
     \begin{subfigure}[b]{0.4\textwidth}
         \centering
         \includegraphics[width=\textwidth]{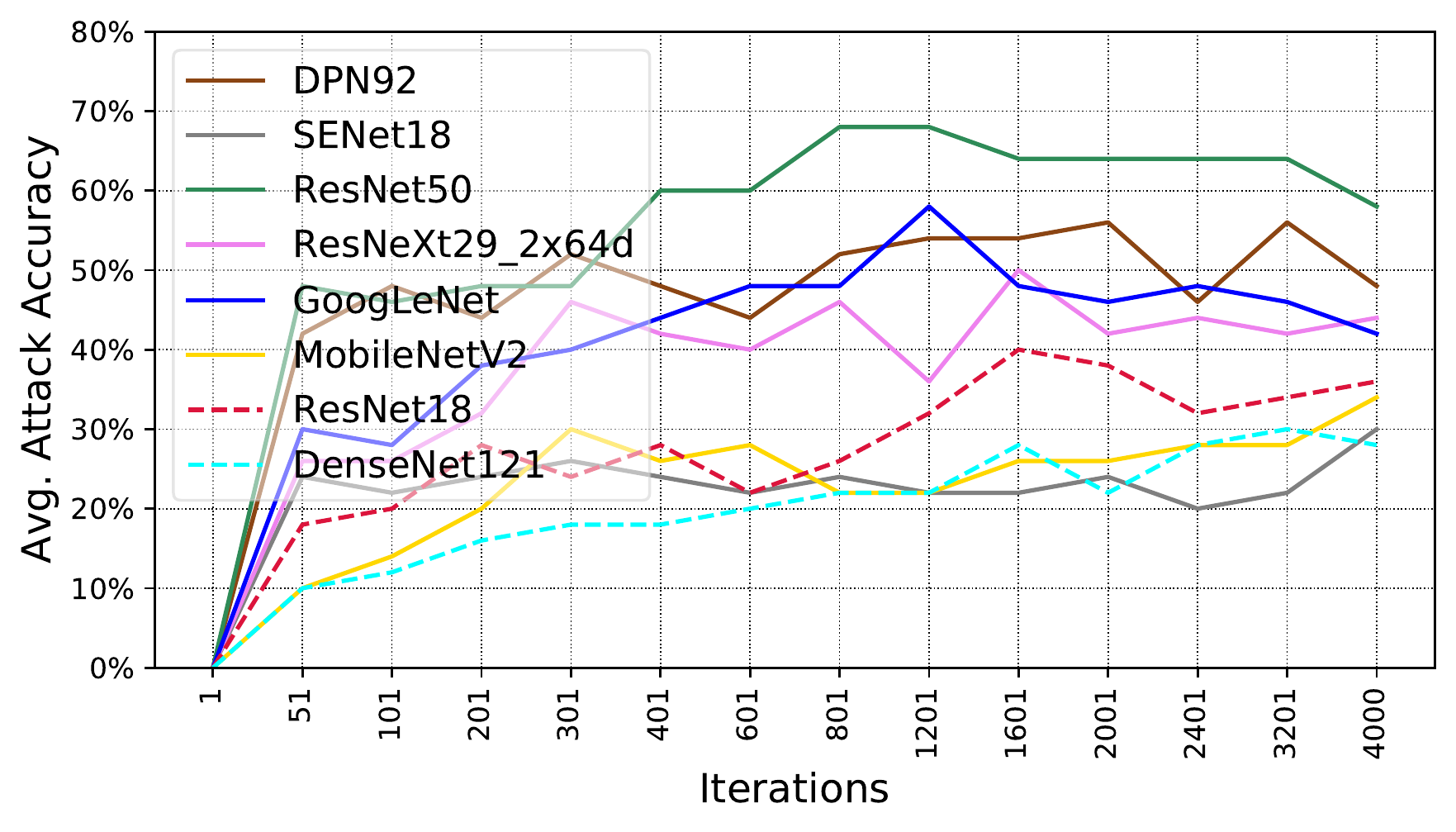}
         \vspace{-0.6cm}
         \caption{CP}
         \label{fig:single-tr-CP-attacksuccrate}
     \end{subfigure}
     \hfill
     \begin{subfigure}[b]{0.4\textwidth}
         \centering
         \includegraphics[width=\textwidth]{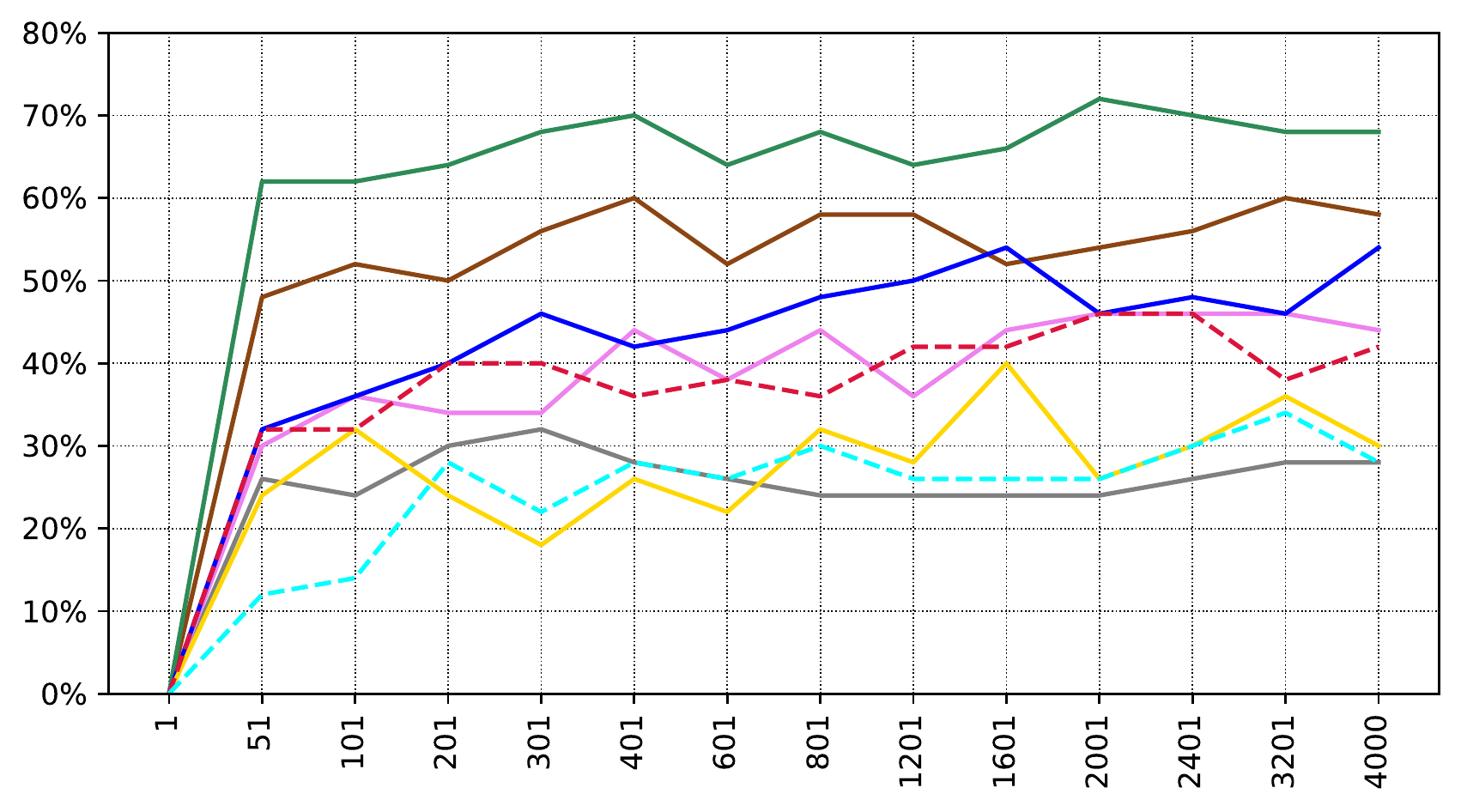}
         \vspace{-0.6cm}
         \caption{\sysshortOne{}}
         \label{fig:single-tr-SP-attacksuccrate}
     \end{subfigure}
     
     \begin{subfigure}[b]{0.4\textwidth}
         \centering
         \includegraphics[width=\textwidth]{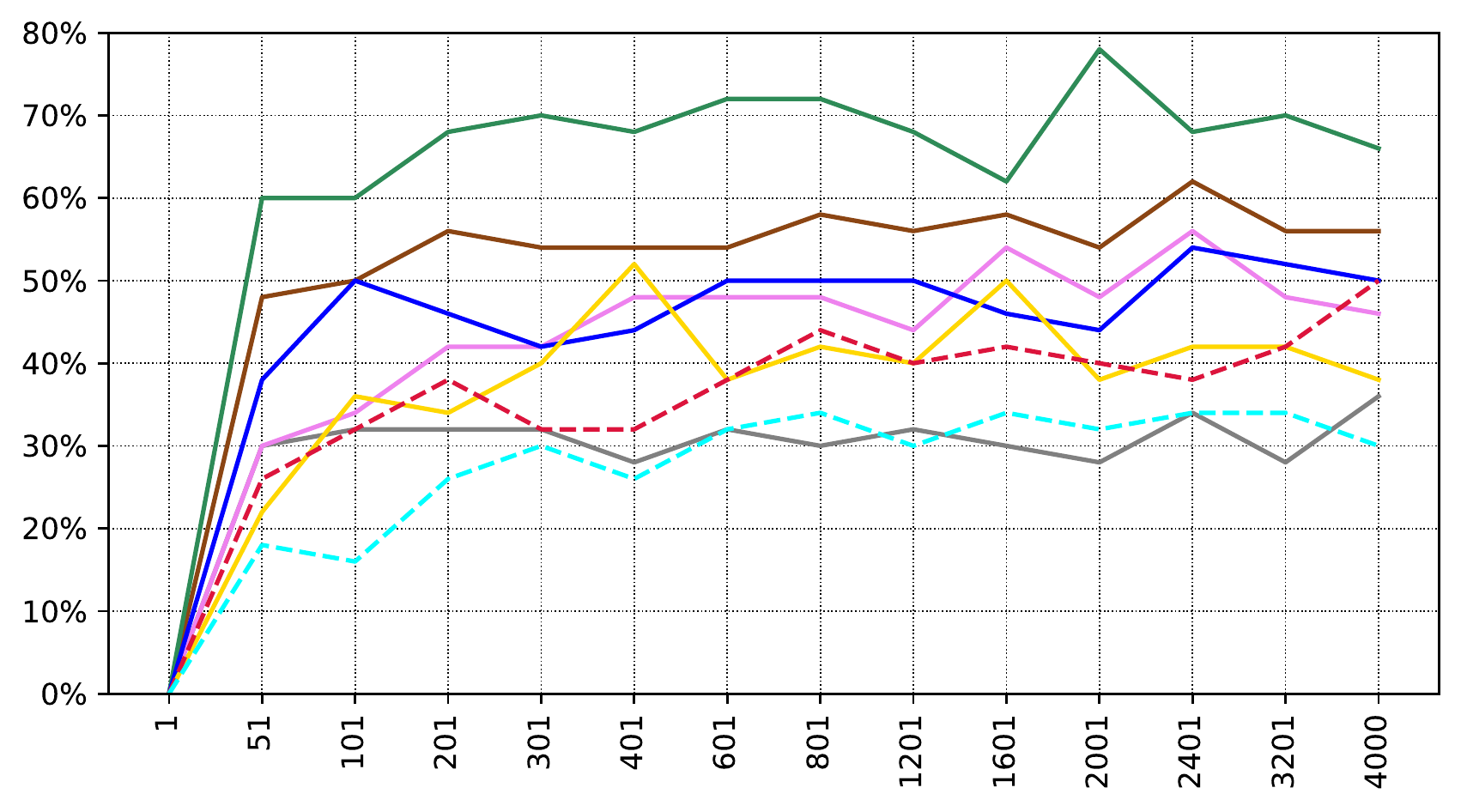}
         \vspace{-0.6cm}
         \caption{\sysshortThree{}}
         \label{fig:single-tr-SP3-attacksuccrate}
     \end{subfigure}
     \hfill
     \begin{subfigure}[b]{0.4\textwidth}
         \centering
         \includegraphics[width=\textwidth]{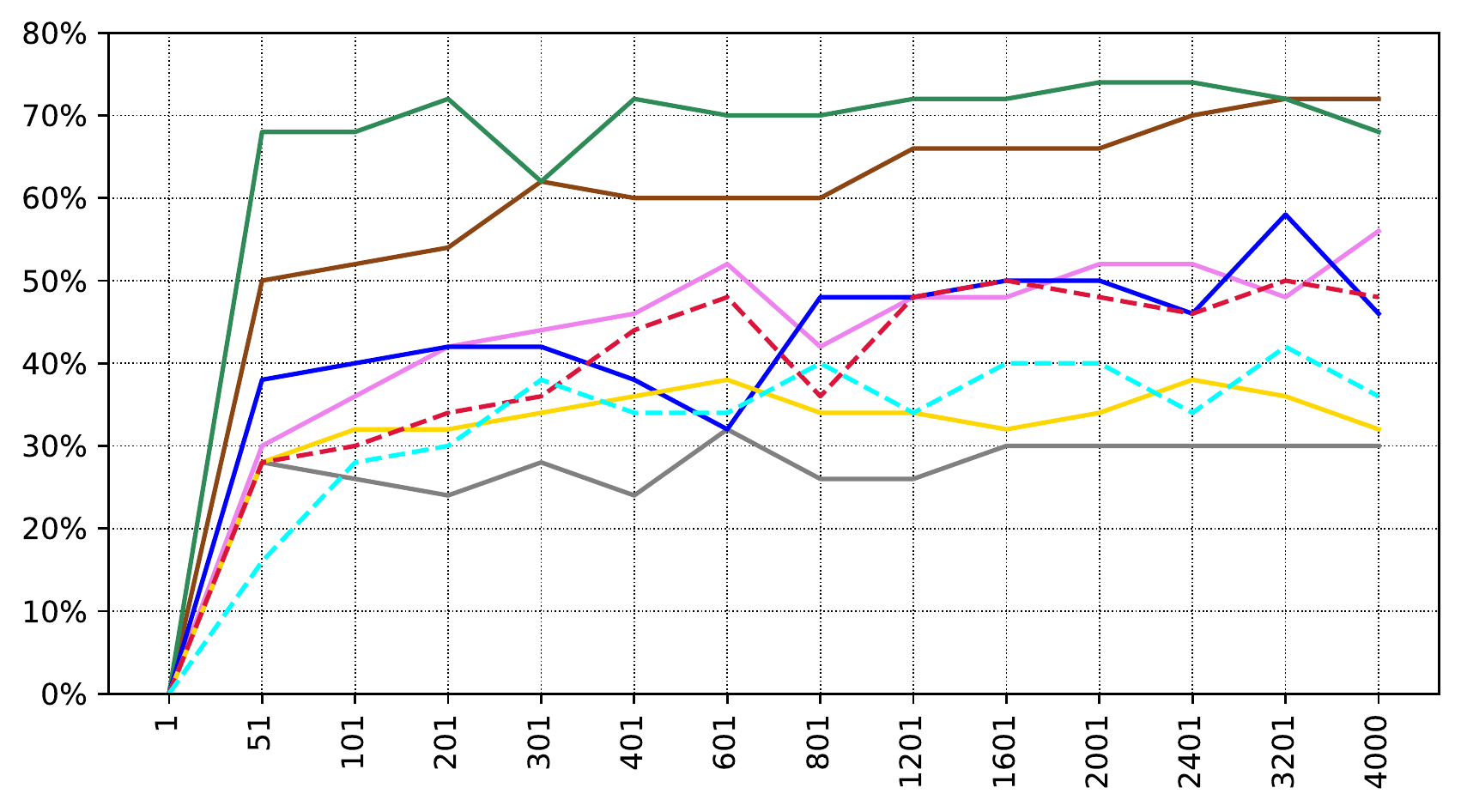}
         \vspace{-0.6cm}
         \caption{\sysshortFive{}}
         \label{fig:single-tr-SP5-attacksuccrate}
     \end{subfigure}
     \caption{Linear transfer learning - success rates of CP, \sysshortOne{}, \sysshortThree{}, and \sysshortFive{} on victim models. Notice \texttt{ResNet18} and \texttt{DenseNet121} are the black-box setting.}
     \label{fig:single-tr-attacksuccrate}
\end{figure*}
\begin{figure*}[t]
    \vspace{-0.3cm}
	\centering
	\begin{subfigure}[b]{0.4\textwidth}
    	\includegraphics[width=\textwidth]{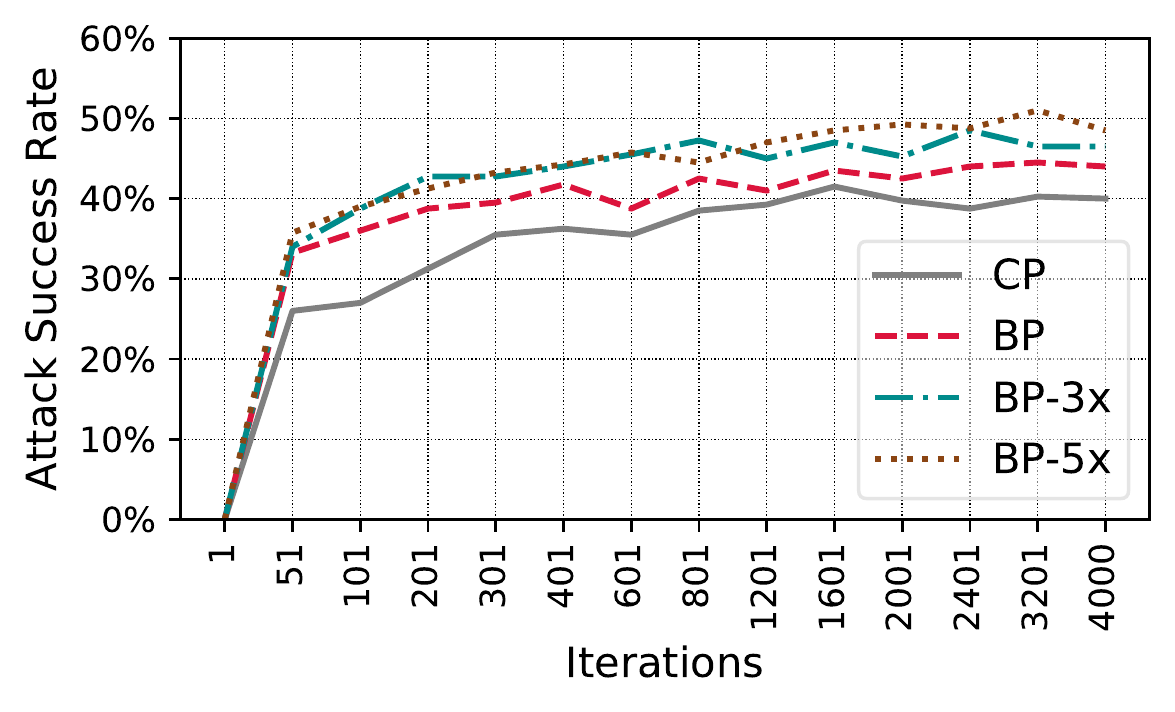}
    	\vspace{-0.7cm}
    	\caption{Linear transfer learning}
    	\label{fig:single-tr-meanVictim}
    \end{subfigure}
    \hfill
    \begin{subfigure}[b]{0.4\textwidth}
    	\centering
    	\includegraphics[width=\textwidth]{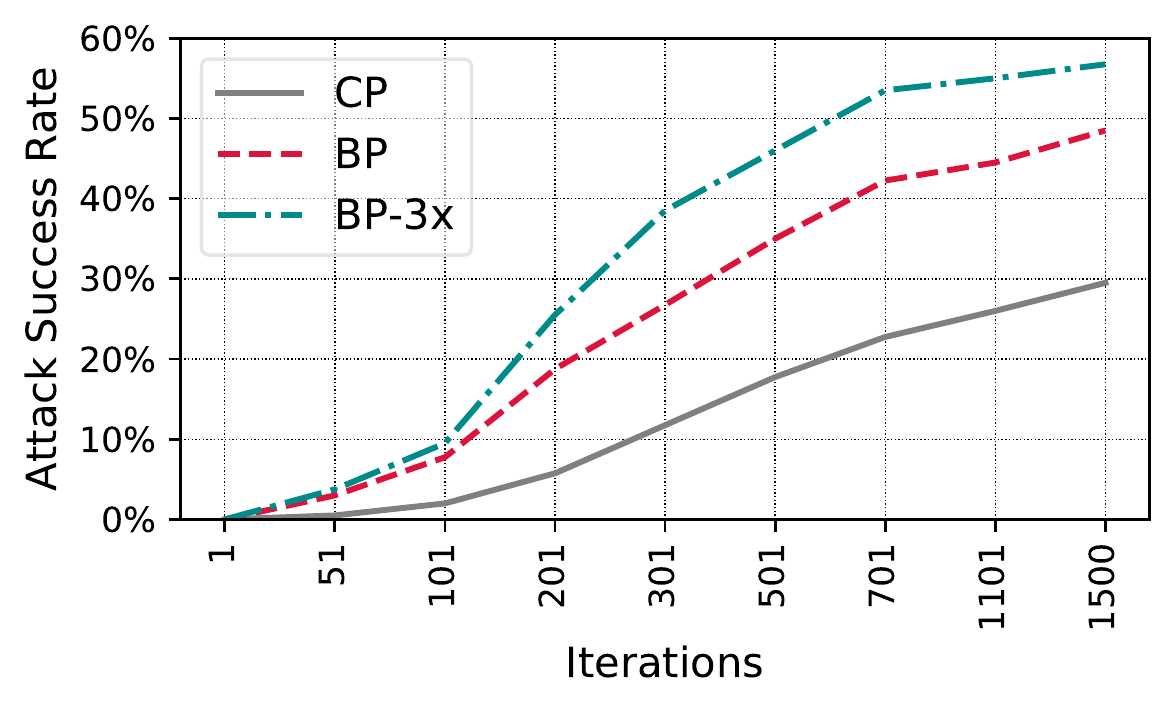}
    	\vspace{-0.7cm}
    	\caption{End-to-end transfer learning}
    	\label{fig:single-end-meanVictim}
    \end{subfigure}
    \vspace{-0.1cm}
    \caption{Attack success rates of CP, \sysshortOne{}, \sysshortThree{}, and \sysshortFive{}, averaged over all eight victim models.}
    \vspace{-0.3cm}
\end{figure*}
\begin{figure*}
     \centering
     \vspace{-0.4cm}
     \begin{subfigure}[b]{0.3\textwidth}
         \centering
         \includegraphics[width=\textwidth]{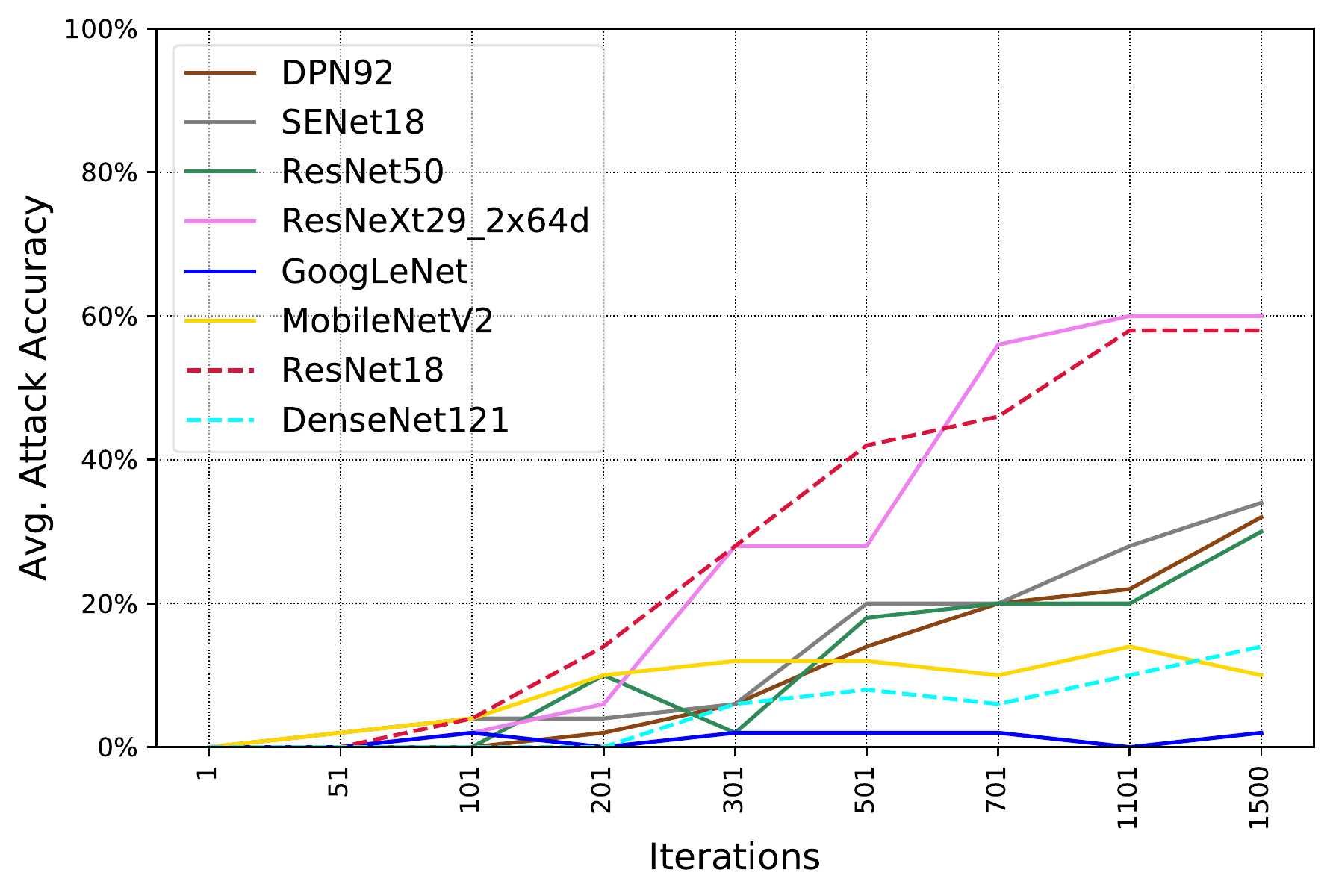}
         \vspace{-0.6cm}
         \caption{CP}
     \end{subfigure}
     \begin{subfigure}[b]{0.3\textwidth}
         \centering
         \includegraphics[width=\textwidth]{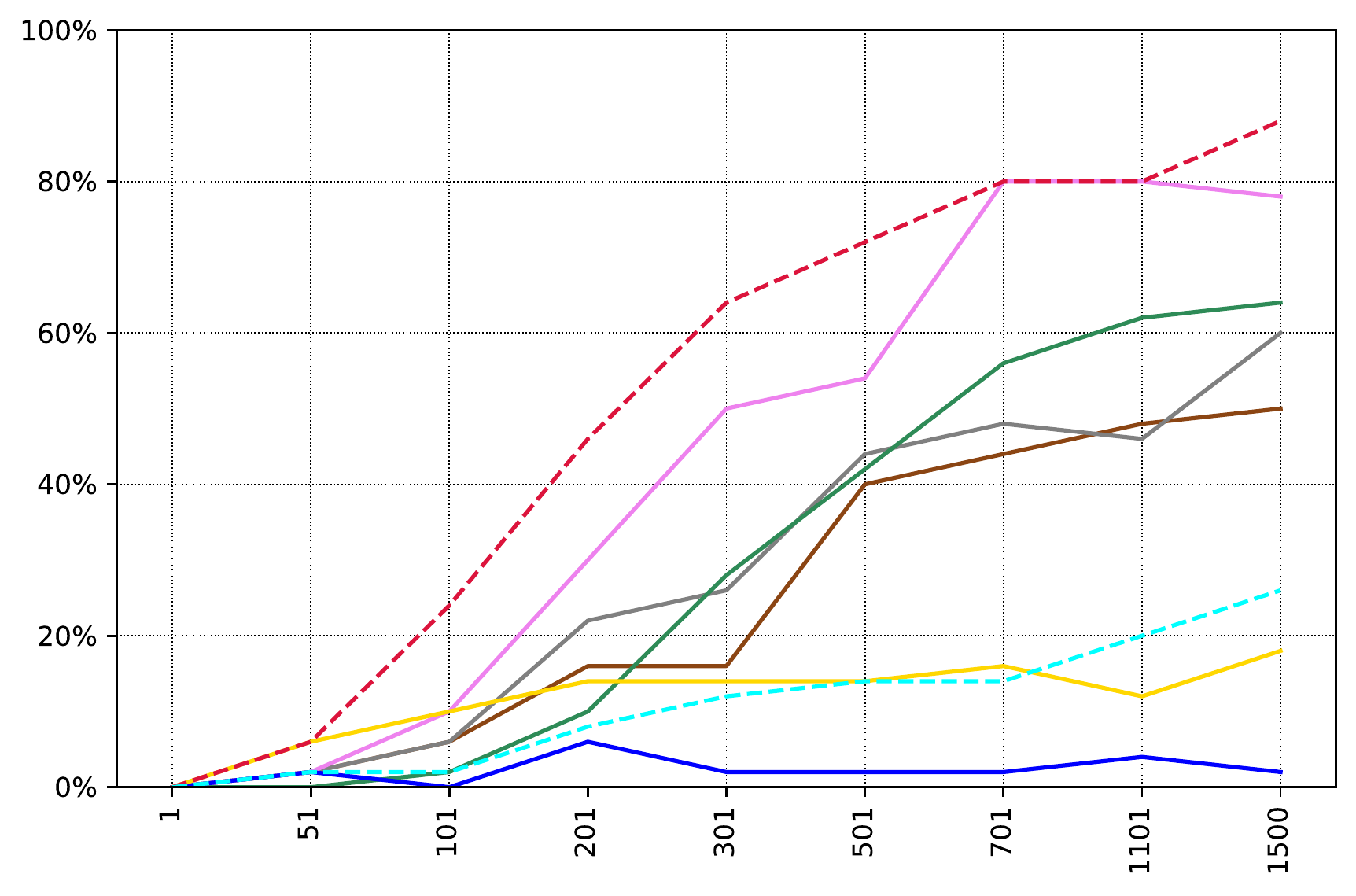}
         \vspace{-0.6cm}
         \caption{\sysshortOne{}}
     \end{subfigure}
     \begin{subfigure}[b]{0.3\textwidth}
         \centering
         \includegraphics[width=\textwidth]{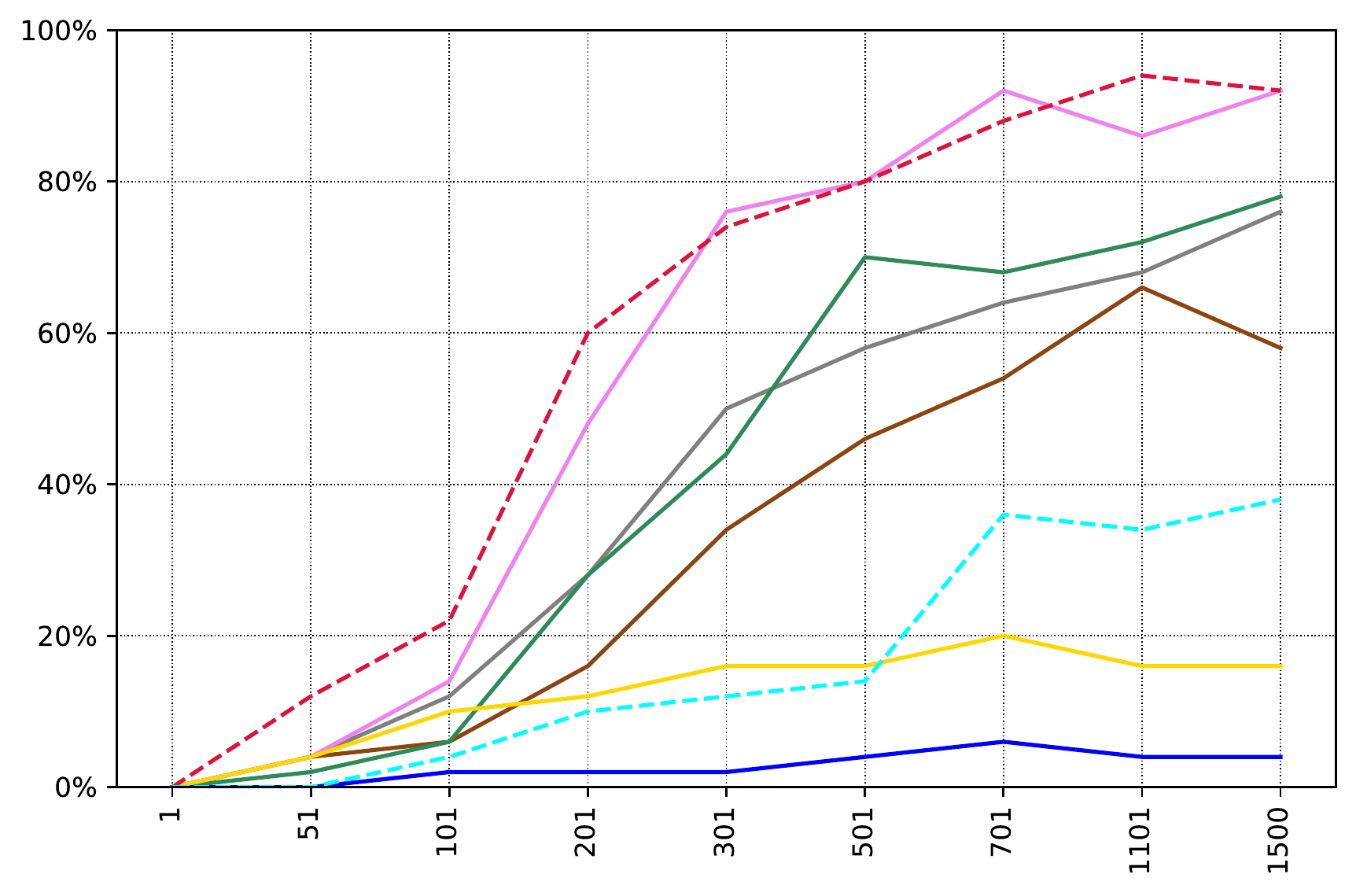}
         \vspace{-0.6cm}
         \caption{\sysshortThree{}}
     \end{subfigure}
     
     \caption{End-to-end transfer learning - success rates of CP, \sysshortOne{}, and \sysshortThree{} on victim models. Notice \texttt{MobileNetV2}, \texttt{GoogLeNet}, \texttt{ResNet18} and \texttt{DenseNet121} are the black-box setting.}
     \label{fig:single-tr-attacksuccrate-end2end}
\end{figure*}

\begin{figure*}[h]
\vspace{-0.3cm}
\centering
\begin{minipage}{0.48\textwidth}
  	\centering
     	\begin{subfigure}[b]{0.7\textwidth}
     		\centering
     		\includegraphics[width=\textwidth]{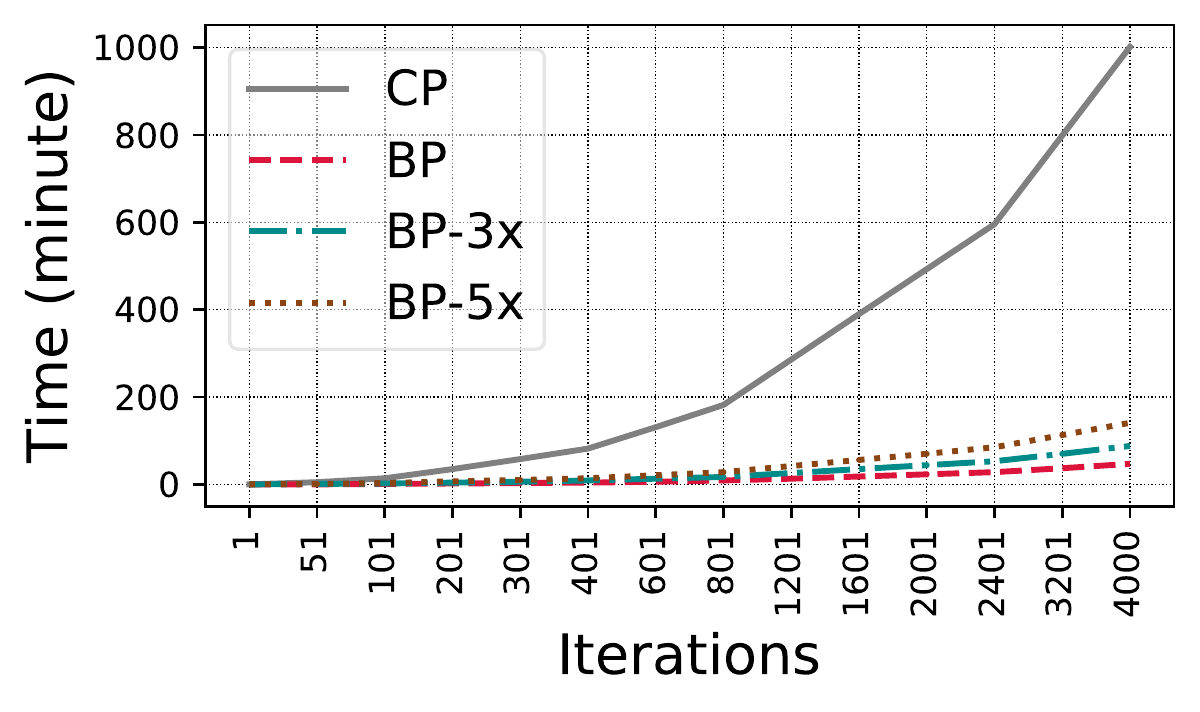}
     		\vspace{-0.6cm}
     		\caption{Linear transfer learning}
     		\label{fig:single-tr-time}
     	\end{subfigure}

     	\begin{subfigure}[b]{0.7\textwidth}
			\centering     	
     		\includegraphics[width=\textwidth]{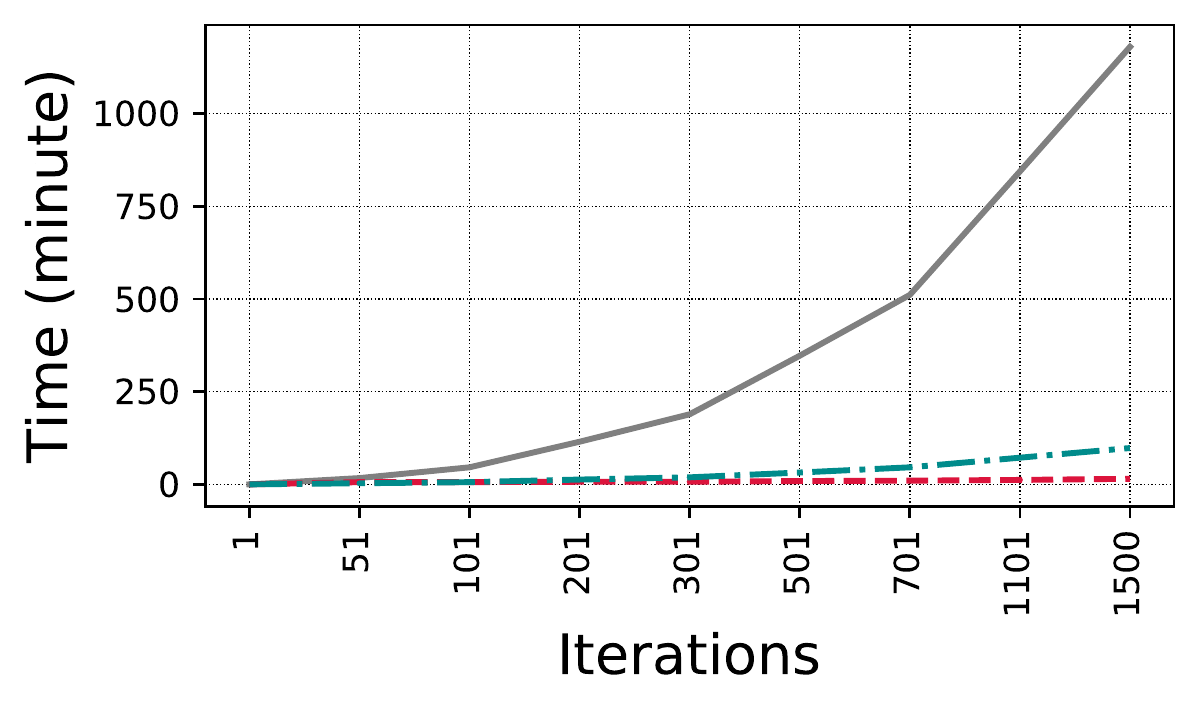}
     		\vspace{-0.6cm}
     		\caption{End-to-end transfer learning}
     		\label{fig:single-end-time}
     	\end{subfigure}
     	\vspace{-0.1cm}
	 	\caption{Attack execution time.}
\end{minipage}
\hfill
\begin{minipage}{.48\textwidth}
  	\centering
		\begin{subfigure}[b]{0.7\textwidth}
    		\includegraphics[width=\textwidth]{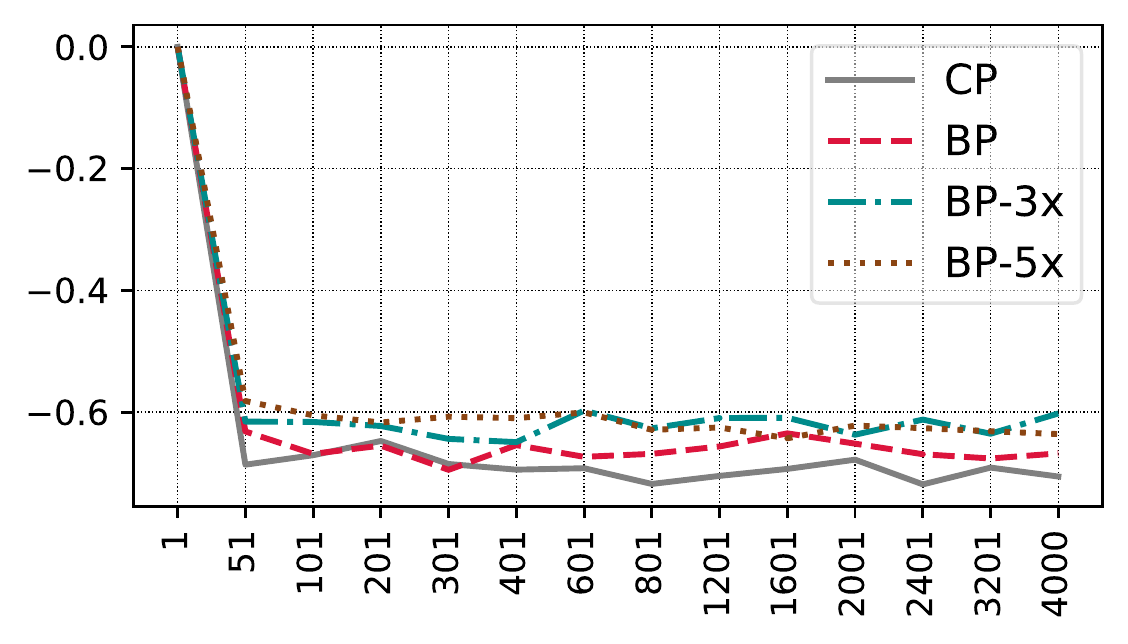}
    		\vspace{-0.6cm}
    		\caption{Linear transfer learning}
    		\label{fig:single-tr-meanVictim-cleantestacc}
    	\end{subfigure}
    	
    	\begin{subfigure}[b]{0.7\textwidth}
    		\centering
    		\includegraphics[width=\textwidth]{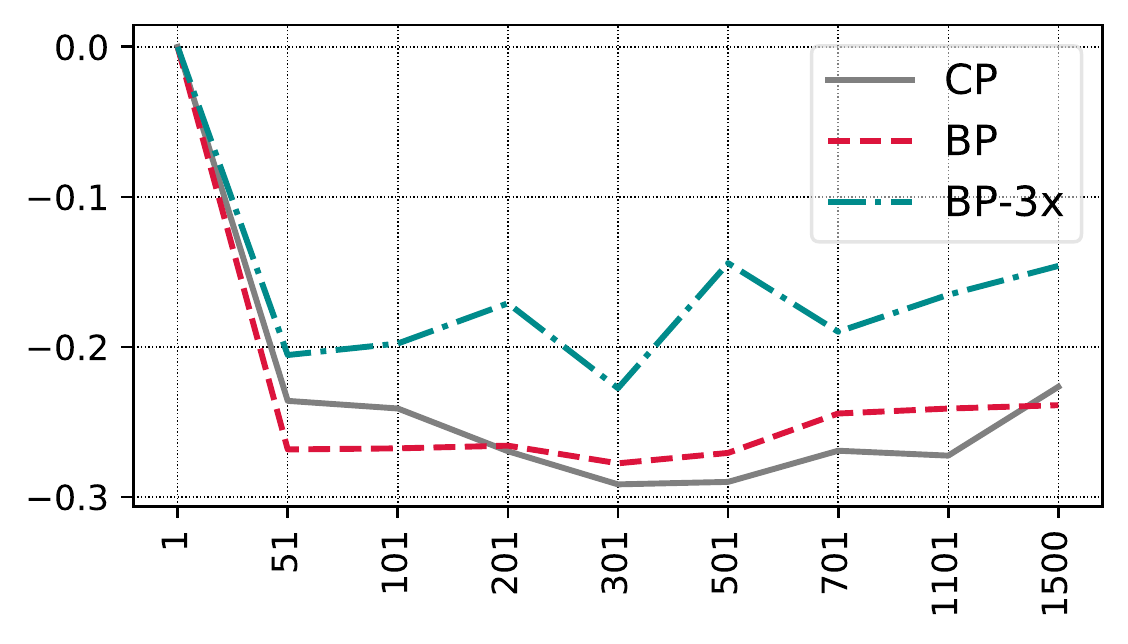}
    		\vspace{-0.6cm}
    		\caption{End-to-end transfer learning}
    		\label{fig:single-end-meanVictim-cleantestacc}
    	\end{subfigure}
    	\vspace{-0.1cm}
    	\caption{Average baseline test accuracy variation.}
\end{minipage}
\vspace{-0.3cm}

\end{figure*}

We first evaluate \sysshortOne{} in single-target mode and compare against CP, and then demonstrate its transferability on unseen images of the target object (multi-target mode).
\sysshortThree{} and \sysshortFive{} represent the case where multi-draw dropout is enabled, with \(R\) set to 3 and 5, respectively.
Unless stated otherwise, we use the same settings as used by Zhu et al.~\cite{zhu2019transferable} to provide a fair comparison.
We also study the effect of the perturbation budget \(\epsilon\) and the number of poison samples on the attack success rate through ablation studies.
Furthermore, we evaluate both \sysshortOne{} and CP against defenses that are proposed by a recent study~\cite{peri2020deep}.
In the end, we further evaluate \sysshortOne{} using standard benchmarks that are developed in a recent study~\cite{schwarzschild2020just}. 
We ran all the attacks using \texttt{NVIDIA Titan RTX} graphics cards.

\subsection{Single-target Mode}

\mypar{Datasets.}
We use the CIFAR-10 dataset.
If not explicitly stated, all the substitute and victim models are trained using the first 4,800 images from each of the 10 classes.
In all experiments, we use the standard test set from CIFAR-10 to evaluate the \emph{baseline test accuracy} of the poisoned models and compare them with their unpoisoned counterparts.
The attack targets, base images of poison samples, and victim's fine-tuning set are selected from the remaining 2,000 images of the dataset.
We assume that the victim models are fine-tuned on a training set consisting of the first 50 images from each class, i.e., the \emph{fine-tuning dataset}, containing a total of 500 images.
Zhu et al.~\cite{zhu2019transferable} randomly selected ``ship'' as the misclassification class, and ``frog'' as the target's image class.
We assume the same choice for comparison fairness.
Specifically, the attacker crafts clean-label poison samples from ship images to cause a particular frog image to be misclassified as a ship.
We craft the poison images \(x_p^{(j)}\) from the first five images of the ship class in the fine-tuning dataset.
We run CP and \sysshortOne{} attacks with 50 different target images of the frog class (indexed from 4,851 to 4,900) to collect performance statistics.
Thus we ensure that target images, training set, and fine-tuning set are mutually exclusive subsets.
We set an \(\ell_\infty\) perturbation budget of \(\epsilon=0.1\).

\mypar{Linear Transfer Learning.}
For substitute networks, we use SENet18~\cite{hu2018squeeze}, ResNet50~\cite{he2016deep}, ResNeXt29-2x64d~\cite{xie2017aggregated}, DPN92~\cite{chen2017dual}, MobileNetV2~\cite{sandler2018mobilenetv2}, and GoogLeNet~\cite{szegedy2015going}.
Each network architecture is trained with dropout probabilities of 0.2, 0.25, and 0.3, which results in a total of 18 substitute models.
To evaluate the attacks under gray-box settings, we use the aforementioned architectures (although trained with a different random seed).
For black-box settings, we use two new architectures, ResNet18~\cite{he2016deep} and DenseNet121~\cite{huang2017densely}.
Dropout remains activated when crafting the poison samples to improve attack transferability.
However, all eight victim models are trained without dropout, and dropout is disabled during evaluation.
We perform both CP and \sysshortOne{} for 4,000 iterations with the same hyperparameters used by CP.
The only difference is that \sysshortOne{} forces the coefficients to be uniform, i.e., \(c_j^{(i)}\!=\!\frac{1}{5}\).
We use Adam~\cite{kingma2014adam} with a learning rate of 0.1 to fine-tune the victim models on the poisoned dataset for 60 epochs.

Figure~\ref{fig:single-tr-attacksuccrate} shows the progress of CP, \sysshortOne{}, \sysshortThree{}, and \sysshortFive{} over the number of iterations of the attack against each individual victim model.
Figure~\ref{fig:single-tr-meanVictim} shows the attack progress wherein the attack success rate is averaged over eight victim models.
In general, \sysshortOne{} outperforms CP and converges faster.
In particular, on average over all iterations, \sysshortThree{} and \sysshortFive{} demonstrate \threespImproveSingleTransfer{} and \fivespImproveSingleTransfer{} higher attack success rates than CP.
Both CP and \sysshortOne{} hardly affect the baseline test accuracy of models (Figure~\ref{fig:single-tr-meanVictim-cleantestacc}).\footnote{BP has slightly less severe effect on the baseline test accuracy.}
\sysshortOne{} is almost \spFasterCPSingleTransfer{} times faster than CP, as it excludes the computation-heavy step of optimizing the coefficients.
Figure~\ref{fig:single-tr-time} shows the attack execution time based on the number of iterations.
Running CP for 4,000 iterations takes \cpMinsSingleTransfer{} minutes on average, while \sysshortOne{} takes only \spMinsSingleTransfer{} minutes.
\sysshortThree{} and \sysshortFive{} take \threespMinsSingleTransfer{} and \fivespMinsSingleTransfer{} minutes, respectively.
It is worth noting that \sysshortOne{} needs fewer iterations than CP to achieve the same attack success rate for some victim models (Figure~\ref{fig:single-tr-attacksuccrate}).


\mypar{End-to-end Transfer Learning.} In this mode, the victim feature extractor is altered during the fine-tuning process, which results in a (slightly) different feature space.
This causes the conventional CP attack to have a success rate of less than 5\%.
To tackle this problem, CP creates convex polytopes in different layers of the substitute models.
We follow the same strategy for \sysshortOne{}, 
this time limiting each attack to 1,500 iterations to meet time and resource constraints.
For substitute networks, we use SENet18, ResNet50, ResNeXt29-2x64d, and DPN92, with dropout values of 0.2, 0.25, and 0.3 (a total of 12 substitute models).
For gray-box testing, we evaluate the attacks against these four architectures.
In the black-box setting, MobileNetV2, GoogLeNet, ResNet18, and DenseNet121 are used as victim networks.
We use Adam with a learning rate of \(10^{-4}\) to fine-tune the victim models on the poisoned dataset for 60 epochs.

Similar to what we observed for \fix{}, but with a wider margin, \sysshortOne{} presents higher attack transferability than CP, especially in the black-box setting.
Figure~\ref{fig:single-tr-attacksuccrate-end2end} shows the progress of CP, \sysshortOne{}, and \sysshortThree{} over the number of iterations of the attack against each individual victim model.
Here we report attack success rates after 1,500 iterations. 
\sysshortOne{} and \sysshortThree{} improve average attack transferability (over victim models) by \spImproveSingleEnd{} and \threespImproveSingleEnd{}, respectively (Figure~\ref{fig:single-end-meanVictim}).
Figure~\ref{fig:single-end-individualVictim} in the Appendix shows attack success rates against each individual victim model.
\sysshortOne{} and \sysshortThree{} have 10-30\% and 10-50\% higher attack transferability than CP, respectively (except against GoogLeNet).
Poor transferability against GoogLeNet is also reported for CP~\cite{zhu2019transferable}.
Since the GoogLeNet architecture differs significantly from the substitute models, it is, therefore, more difficult for the ``attack zone'' to survive \e2e{}.
For other black-box models (MobileNetV2, ResNet18, and DenseNet121), \sysshortOne{} and \sysshortThree{} improve attack transferability by \(\sim\)18\% and \(\sim\)24\%, respectively.
Both CP and \sysshortOne{} have hardly any effect on the baseline test accuracy of models (Figure~\ref{fig:single-end-meanVictim-cleantestacc}).
As Figure ~\ref{fig:single-end-time} shows, \sysshortOne{} and \sysshortThree{} take \spMinsSingleEnd{} and \threespMinsSingleEnd{} minutes, while CP takes \cpMinsSingleEnd{} minutes, which is \spFasterCPSingleEnd{}x slower.

It is worth noting that we found multi-draw dropout not beneficial to CP in the experiments. 
Since using multi-draw dropout makes CP (much) slower, with no gain in attack success rate, to fairly compare the execution time of BP with CP, multi-draw dropout is always disabled for CP.

\begin{figure*}[t]
	\centering
	\vspace{-0.4cm}
	\begin{subfigure}{0.4\textwidth}
    	\includegraphics[width=0.9\textwidth]{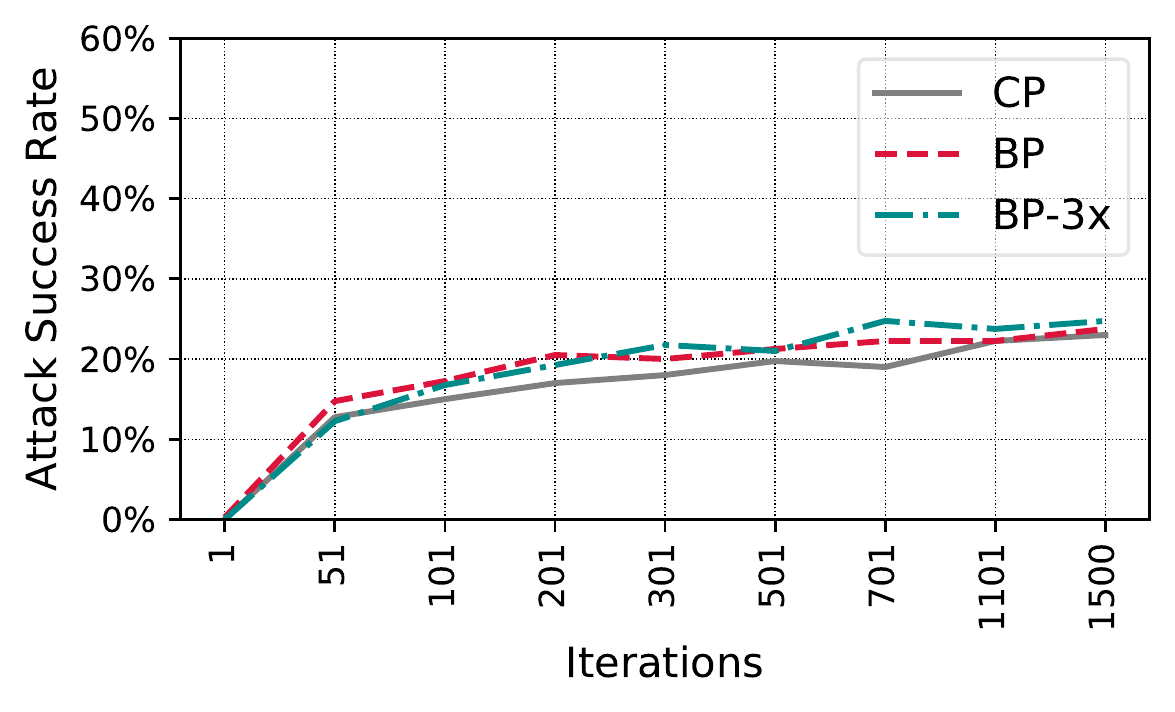}
    	\vspace{-0.2cm}
    	\caption{Zero overlap}
    	\label{fig:diff0-single-tr-meanVictim}
    \end{subfigure}
    \hfill
    \begin{subfigure}{0.4\textwidth}
    	\centering
    	\includegraphics[width=0.9\textwidth]{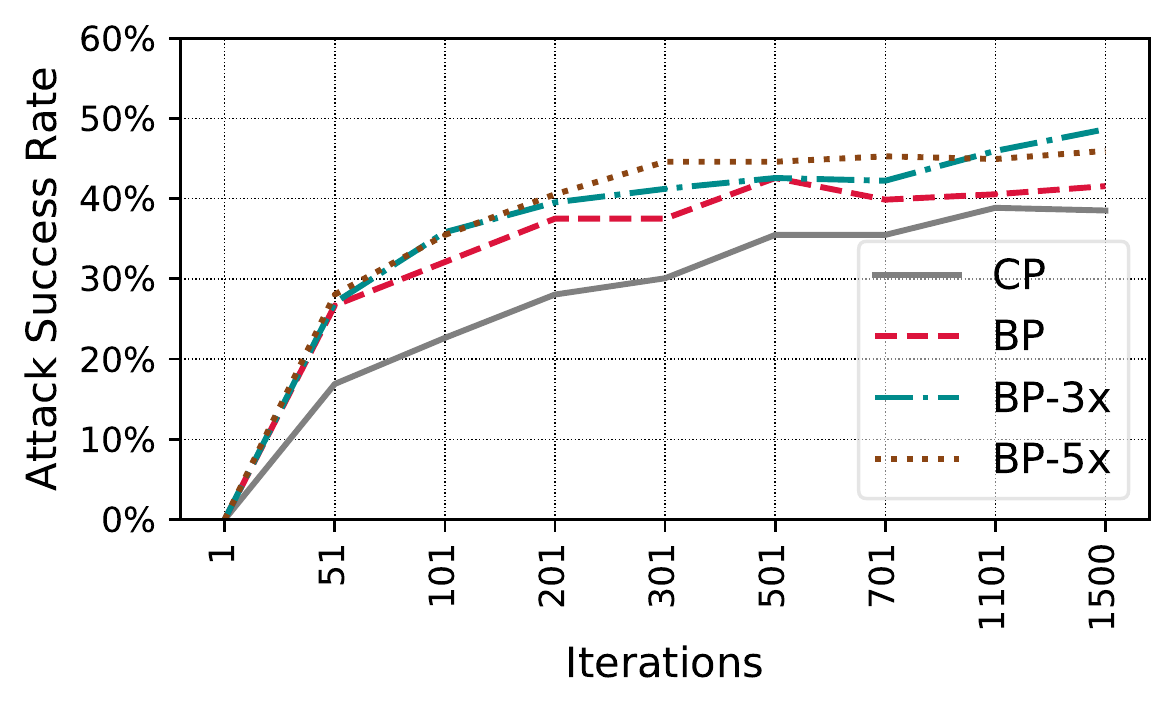}
    	\vspace{-0.2cm}
    	\caption{50\% overlap}
    	\label{fig:diff50-single-tr-meanVictim}
    \end{subfigure}
    \vspace{-0.1cm}
    \caption{Comparison of CP, \sysshortOne{}, and \sysshortThree{} in linear transfer learning, with zero and 50\% overlap between training sets of the substitute networks and the victim's network.}
    \label{fig:diff-single-tr-attacksuccrate}
    \vspace{-0.3cm}
\end{figure*}

\mypar{Transferability to Unseen Training Sets.} 
Until now, we have assumed that the substitute models are trained on the same training set (\(\Psi\)) on which the victim's feature extractor network is trained.
In this section, we evaluate CP and \sysshortOne{} using substitute models that are trained on a training set that has (1) \textbf{zero} or (2) \textbf{50\%} overlap with \(\Psi\).
Such a setting is more realistic compared to when the attacker has complete knowledge of \(\Psi\). 
We use the same setting as in linear transfer learning except for the following changes:
(i) We train the victim models on the first 2,400 images of each class;
(ii) In the zero overlap setting, we train substitute models on samples indexed from 2,401 to 4,800 for each class;
(ii) For the 50\% overlap setting, we train substitute models on samples indexed from 1,201 to 3,600 for each class.
Figure~\ref{fig:diff-single-tr-attacksuccrate} shows the attack success rates (averaged over victims) for both zero overlap and 50\% overlap setups.
When we have 50\% overlap, \sysshortOne{}, \sysshortThree{}, and \sysshortFive{} demonstrate \spImproveSingleTransferToDiffTrainingSet{}, \threespImproveSingleTransferToDiffTrainingSet{}, and \fivespImproveSingleTransferToDiffTrainingSet{} higher attack success rates compared to CP (on average over all iterations), with BP converging significantly faster than CP.
For the zero overlap setup, BP provides hardly any improvement over CP.
They both achieve much lower attack success rates of 20-25\%.
It should be noted that the zero overlap scenario is much more restricted than what is usually assumed in threat models for poisoning attacks.
The victim's network, training set, and even the fine-tuning training set (except for, of course, the poison samples) are all unseen to the adversary.
All attacks hardly affect the baseline test accuracy (Figure~\ref{fig:diff-single-tr-meanVictim-cleantestacc} in the Appendix).

\begin{figure*}
     \centering
     \vspace{-0.4cm}
     \begin{subfigure}[b]{0.13\textwidth}
         \centering
         \includegraphics[width=0.9\textwidth, frame]{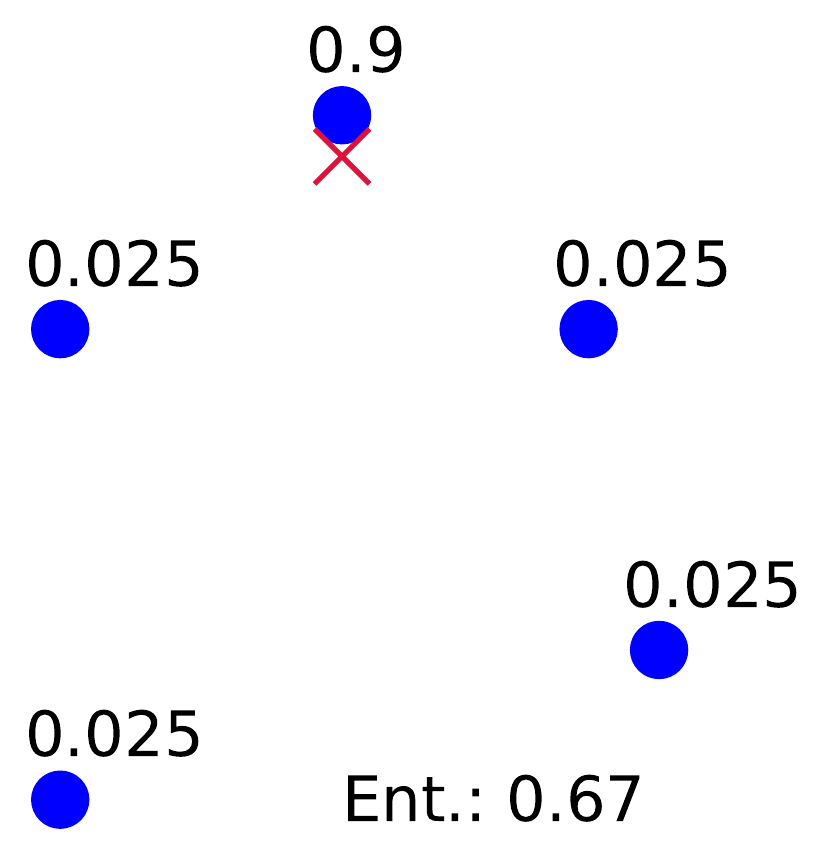}
     \end{subfigure}
     \hfill
     \begin{subfigure}[b]{0.13\textwidth}
         \centering
         \includegraphics[width=0.9\textwidth, frame]{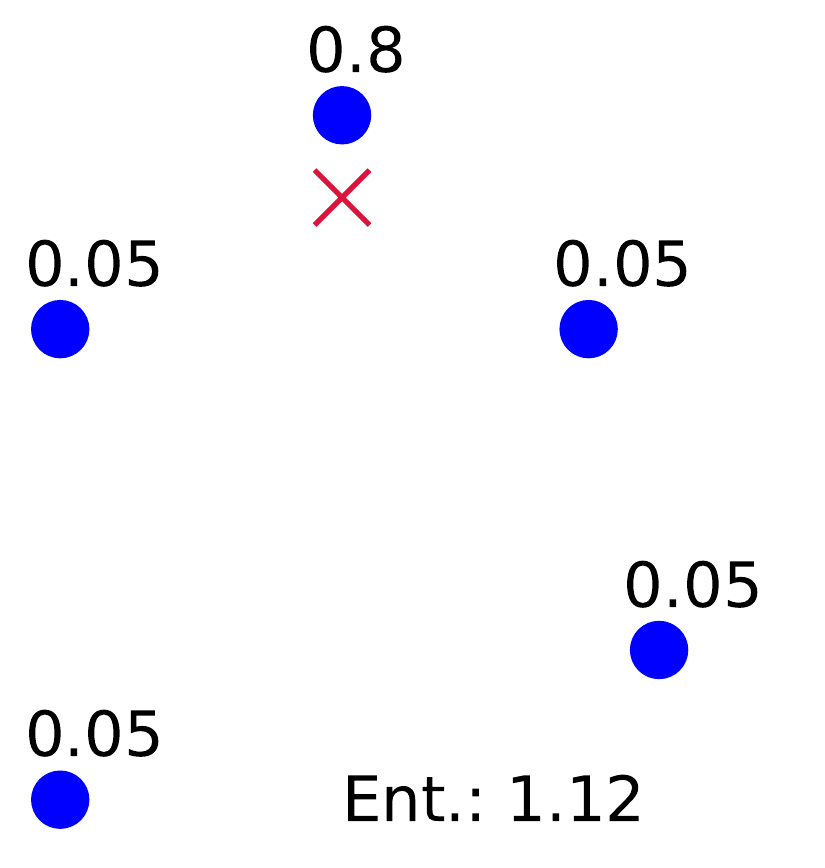}
     \end{subfigure}
     \hfill
     \begin{subfigure}[b]{0.13\textwidth}
         \centering
         \includegraphics[width=0.9\textwidth, frame]{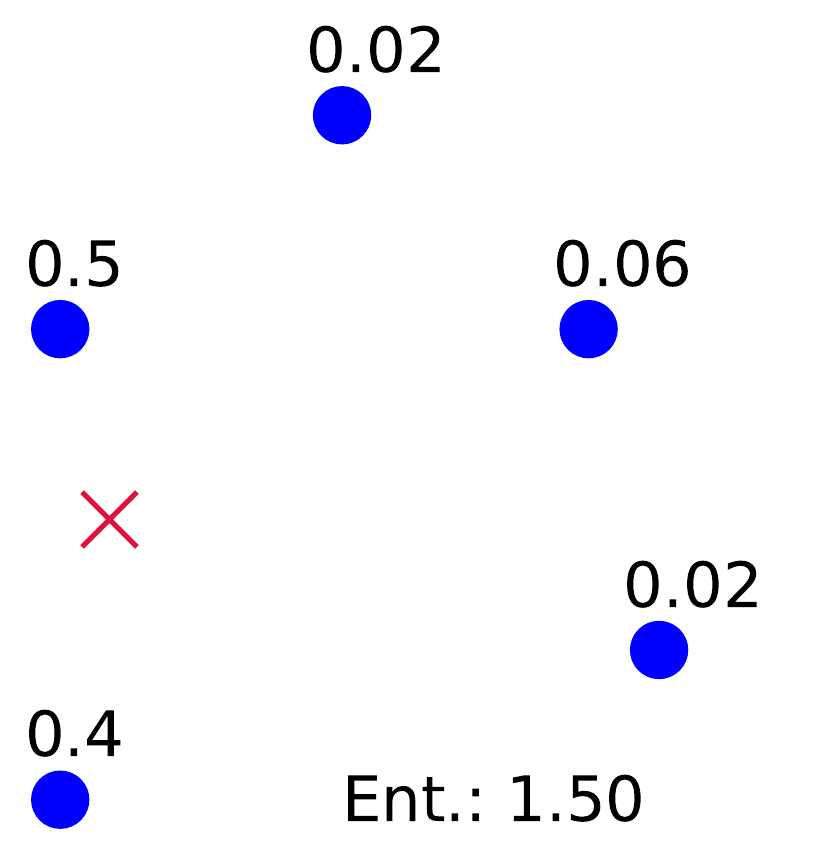}
     \end{subfigure}
     \hfill
     \begin{subfigure}[b]{0.13\textwidth}
         \centering
         \includegraphics[width=0.9\textwidth, frame]{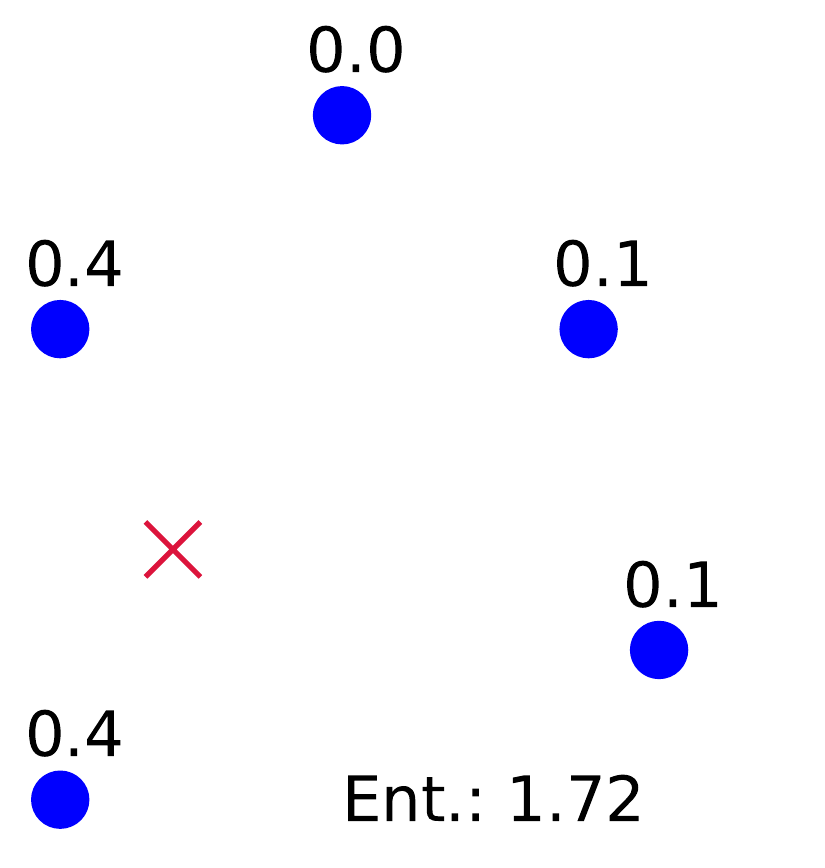}
     \end{subfigure}
     \hfill
     \begin{subfigure}[b]{0.13\textwidth}
         \centering
         \includegraphics[width=0.9\textwidth, frame]{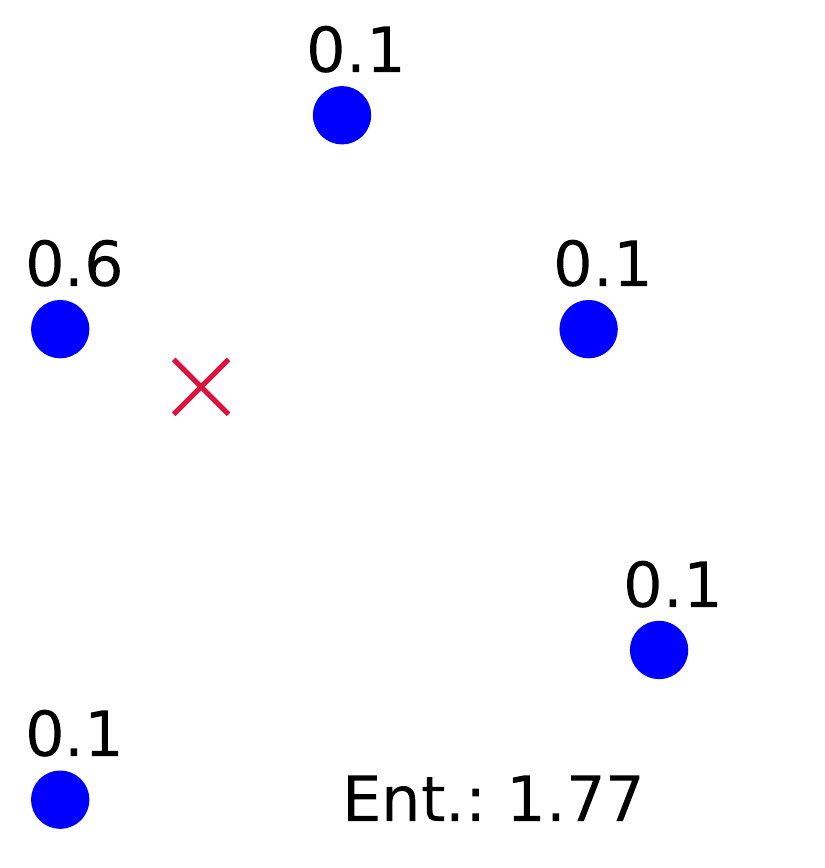}
     \end{subfigure}
     
     \begin{subfigure}[b]{0.13\textwidth}
         \centering
         \includegraphics[width=0.9\textwidth, frame]{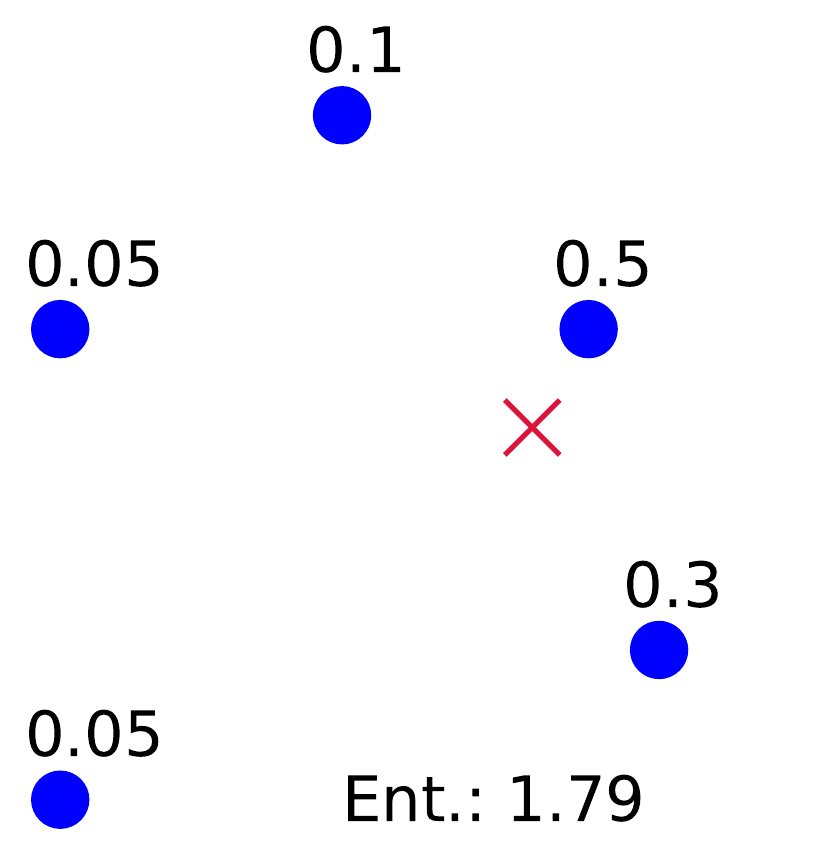}
     \end{subfigure}
     \hfill
     \begin{subfigure}[b]{0.13\textwidth}
         \centering
         \includegraphics[width=0.9\textwidth, frame]{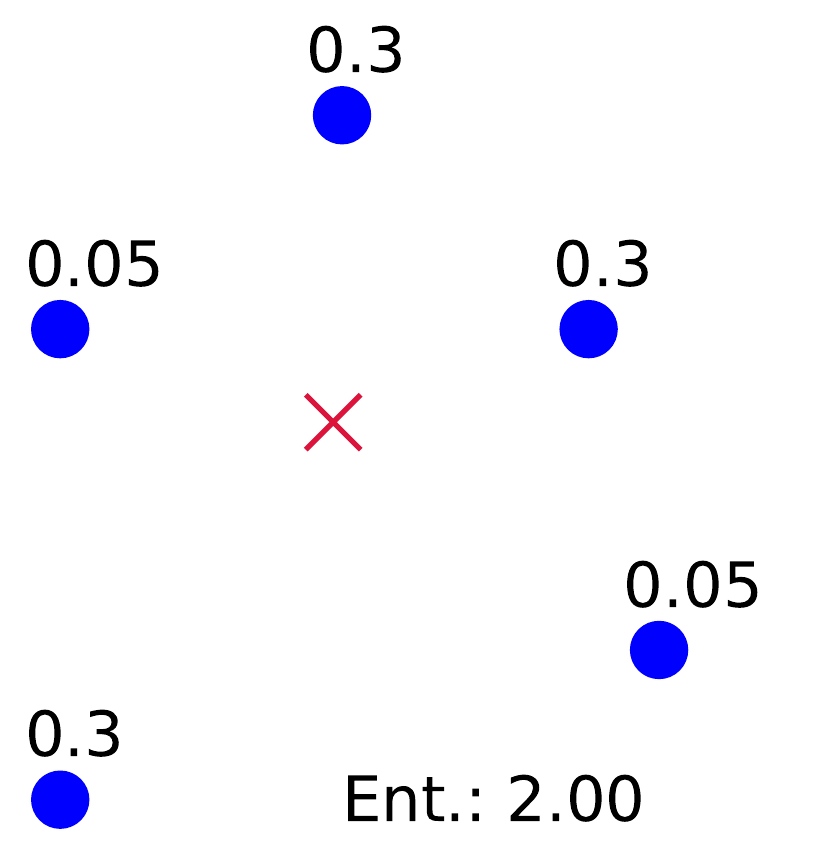}
     \end{subfigure}
     \hfill
     \begin{subfigure}[b]{0.13\textwidth}
         \centering
         \includegraphics[width=0.9\textwidth, frame]{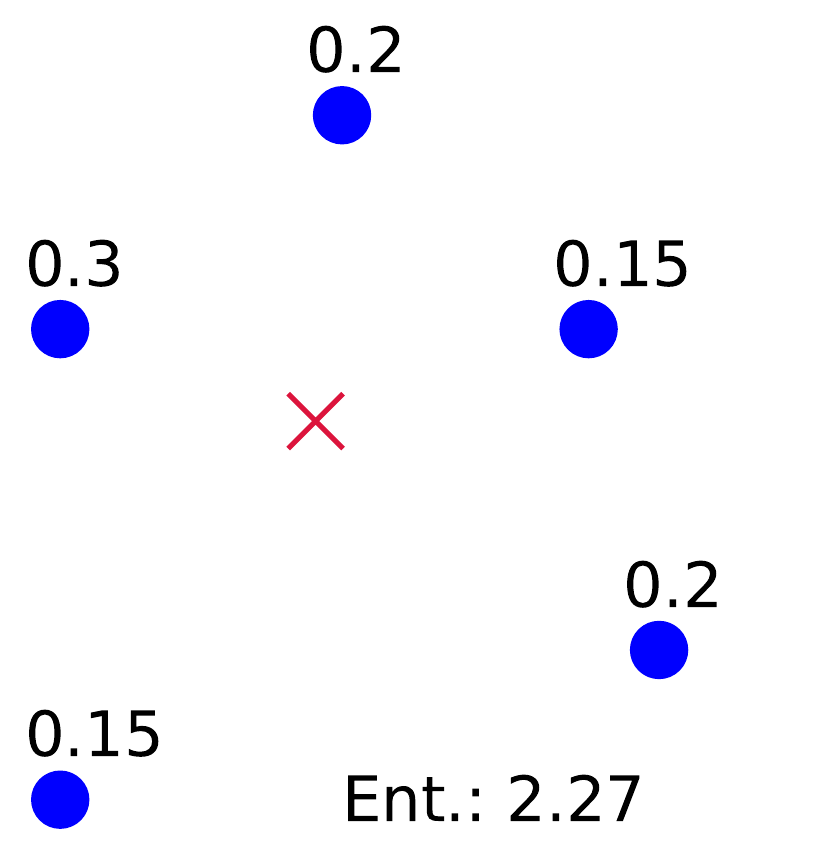}
     \end{subfigure}
      \hfill
     \begin{subfigure}[b]{0.13\textwidth}
         \centering
         \includegraphics[width=0.9\textwidth, frame]{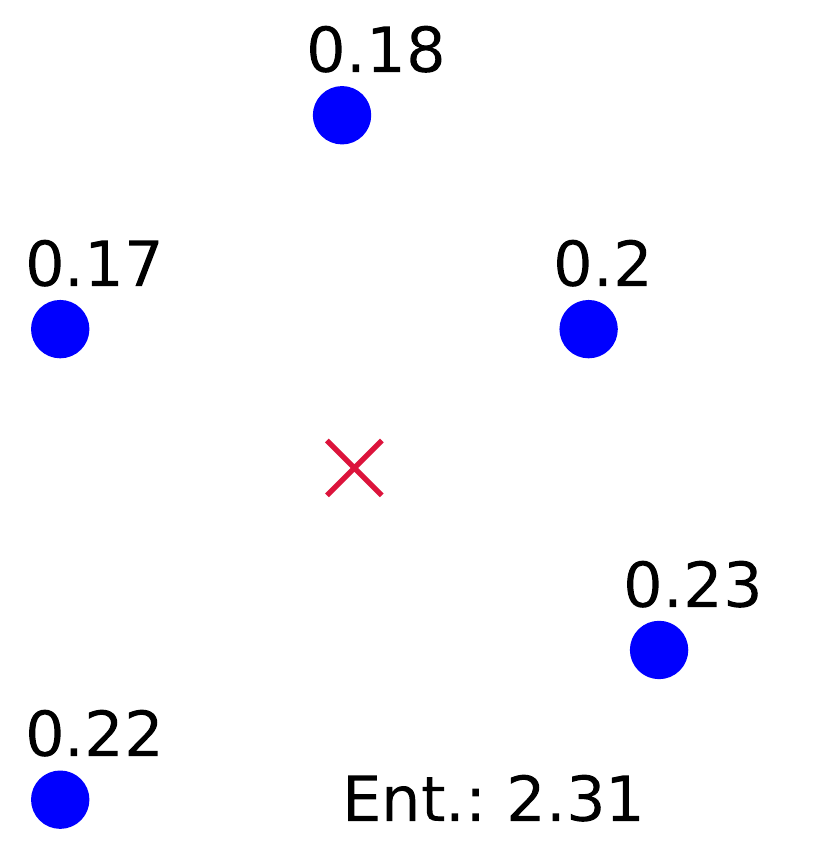}
     \end{subfigure}
      \hfill
     \begin{subfigure}[b]{0.13\textwidth}
         \centering
         \includegraphics[width=0.9\textwidth, frame]{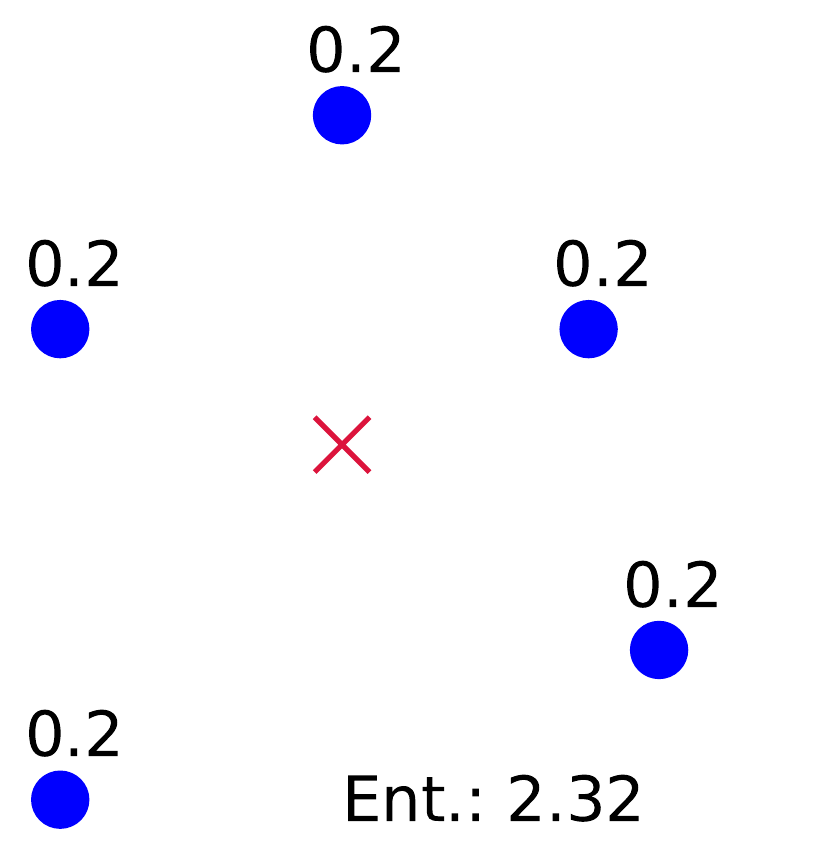}
     \end{subfigure}
     \caption{Nine alternatives of \sys{} with different sets of nonuniform coefficients. The blue circles are poison samples with their coefficients written next to them, and the red cross is the target. The entropy of the coefficients increases from left to right. Note that the bottom right represents \sysshortOne{}.
     }
\label{fig:examples}
\vspace{-0.3cm}
\end{figure*}
\begin{figure*}[t]
	\centering
	\begin{subfigure}[b]{0.32\textwidth}
    	\includegraphics[width=\textwidth]{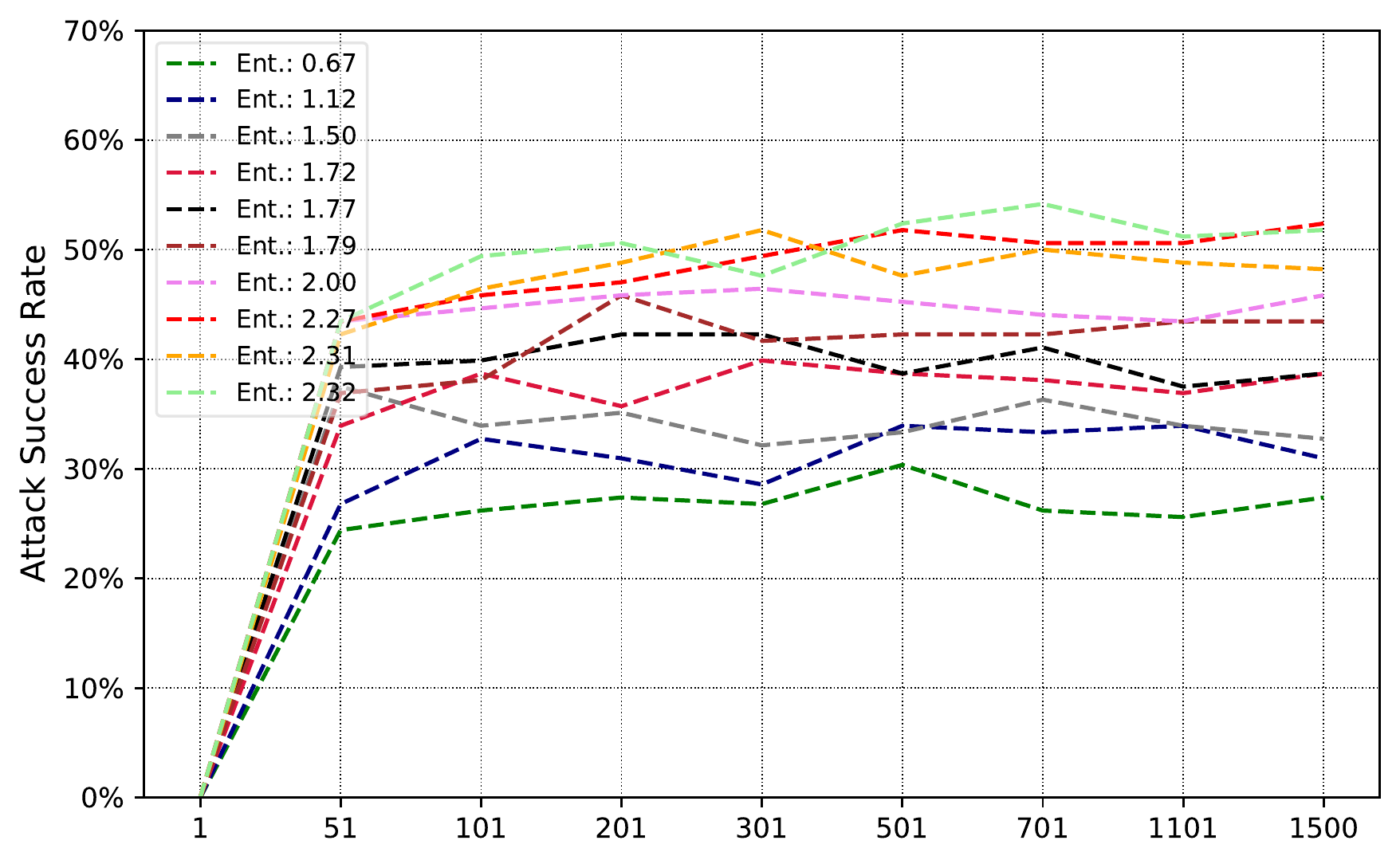}
    	\caption{Averaged over victim models}
    	\label{fig:diff-coeffs-single-tr-meanVictim}
    \end{subfigure}
    \begin{subfigure}[b]{0.32\textwidth}
    	\includegraphics[width=\textwidth]{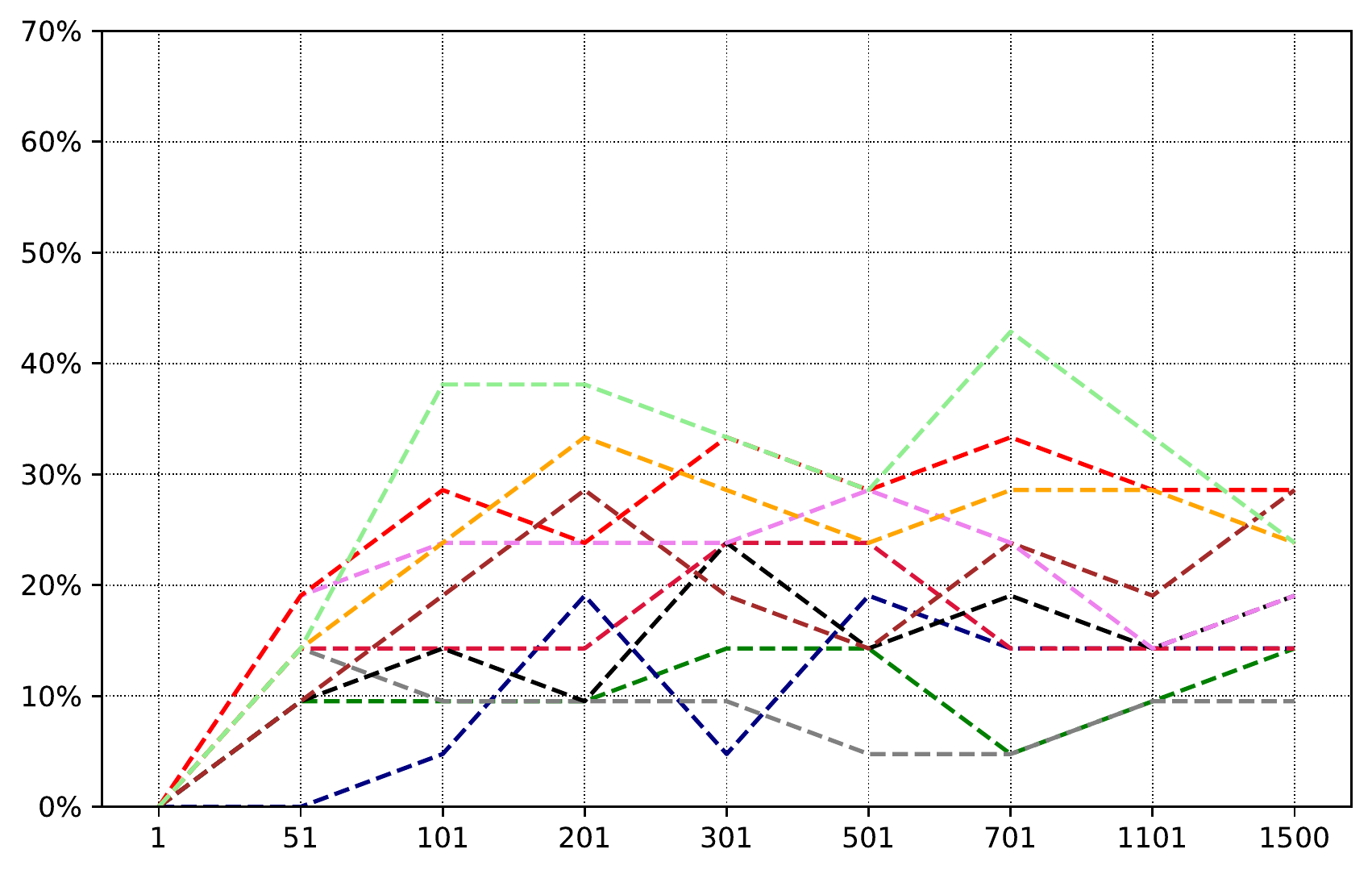}
    	\caption{DenseNet121}
    	\label{fig:diff-coeffs-single-tr-DenseNet121}
    \end{subfigure}
    \begin{subfigure}[b]{0.32\textwidth}
    	\includegraphics[width=\textwidth]{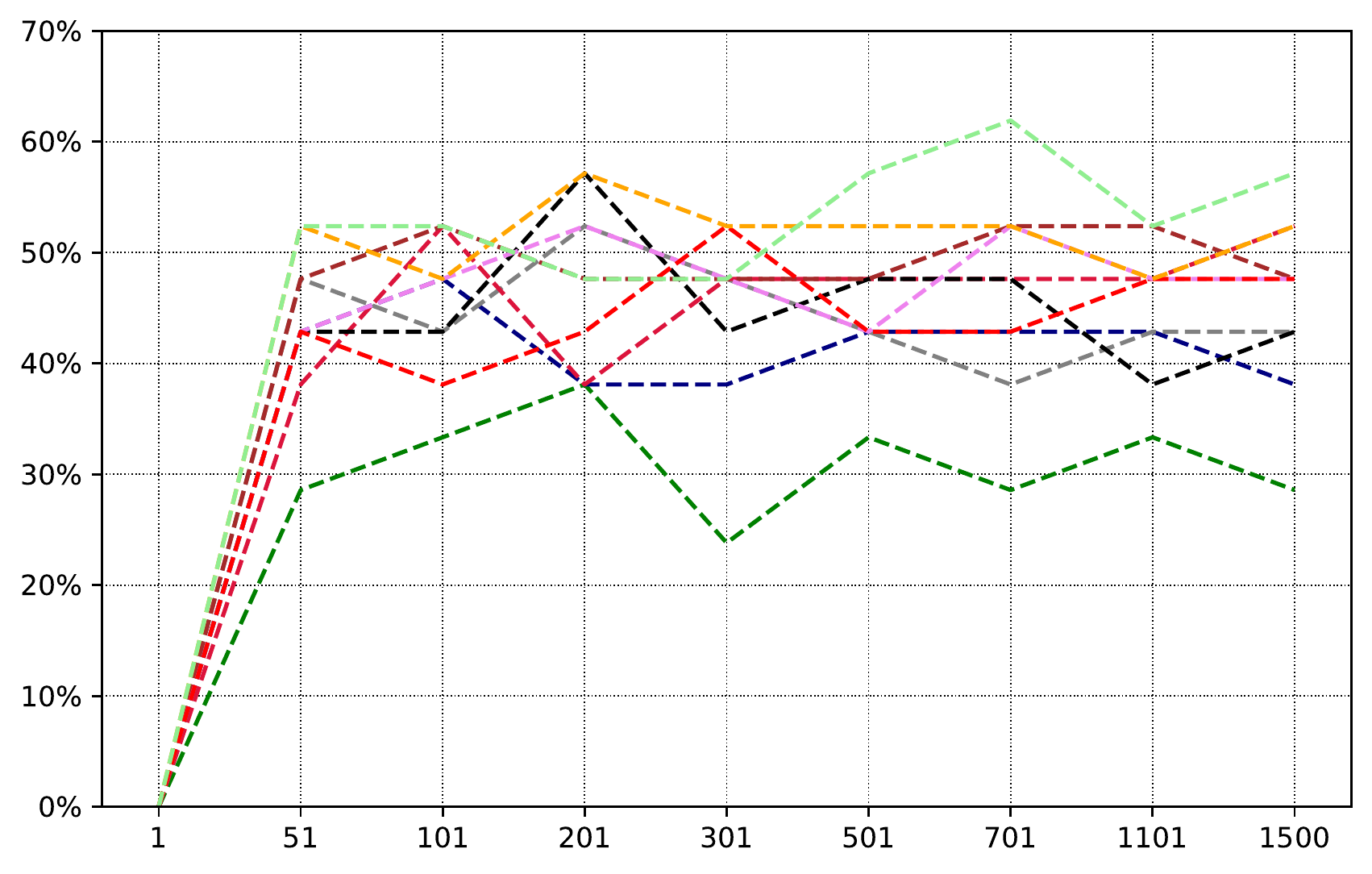}
    	\caption{ResNet18}
    	\label{fig:diff-coeffs-single-tr-ResNet18}
    \end{subfigure}
    \caption{Comparison between \sysshortOne{} and the other nine alternatives.}
    \label{fig:diff-coeffs-single-tr}
    \vspace{-0.4cm}
\end{figure*}
\begin{figure}
    	\centering
        \includegraphics[width=0.8\linewidth]{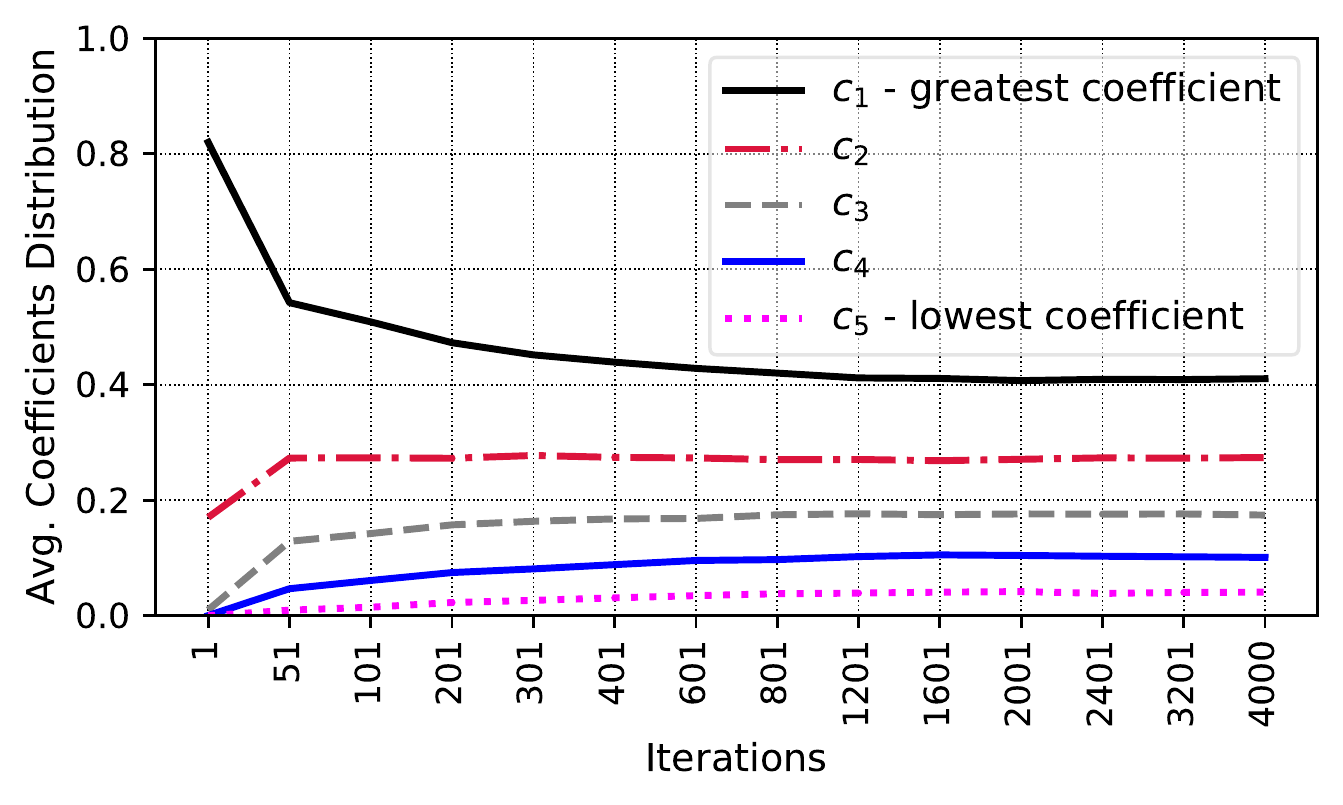}
        \vspace{-0.2cm}
		\caption{Distribution of poison samples' coefficients defined in Eq.~\ref{eq:cvxLoss} (averaged over all targets and victim networks). The coefficients are sorted, \(c_1\) denotes the highest coefficient, and \(c_5\) denotes the lowest coefficient. In Convex Polytope, the coefficient distribution is more skewed with some poison samples having a relatively small contribution to the attack.}
		\label{fig:single-tr-CP-coeffsdist}
		\vspace{-0.3cm}
\end{figure}

\mypar{Effectiveness of the Bullseye Idea.} 
We have argued that the effectiveness (robustness and transferability) of BP stems from the fact that predetermining the convex coefficients as uniform weights draws the target to the ``center'' of the attack zone, increasing its distance from the poison polytope boundary. In order to evaluate this claim quantitatively, we run the attack with different sets of \textit{nonuniform} coefficients, to see if the improvement is truly due to target centering (i.e., the ``bullseye'' idea) or simply from ``fixing'' the coefficients instead of searching for them. 
We evaluate \sysshortOne{} against nine alternatives \(\{\mathrm{BP'_t}\}_{t=1}^9\), each with a different set of positive predefined coefficients that satisfy \(\sum_{j=1}^k\! c_j\!=\!1\).
Figure~\ref{fig:examples} depicts a geometrical example for each set (sorted from left to right based on the entropy of the coefficient vector), with \sysshortOne{} having the highest possible entropy of \(\log_2 5 \simeq 2.32\).
As Figure~\ref{fig:diff-coeffs-single-tr} shows, variations of \sysshortOne{} with higher coefficient entropy generally demonstrate higher attack success rates compared to those with smaller entropy, especially in the black-box setting.
This finding indicates that predetermining the coefficients to uniform weights (\sysshortOne{}) is preferable to simply fixing them to some other plausible values.
This backs our intuition behind \sysshortOne{} that the further the target is from the polytope boundary, the lower its chances of jumping out of the attack zone in the victim's feature space.
In fact, the average entropy of coefficients in CP roughly converges to 1.70, which means the coefficient distribution is more skewed, with some poison samples having a relatively small contribution to the attack.
Figure~\ref{fig:single-tr-CP-coeffsdist} shows the mean values of the (sorted) coefficients to provide a sense of the coefficient distributions used by CP. 

\mypar{Different Pairs of <original class,  poison class>.}
Until now, the adversarial goal in all experiments was to make the victim's model identify an image of a frog (original class) as a ship (poison class). 
For comparison fairness, we have followed Zhu et al.~\cite{zhu2019transferable} for this selection of the original and poison classes.
To assess the impact of selecting different original and poison classes on the performance of the attack, we evaluate \sysshortThree{} for all 90 pairs of <original class,  poison class>; each with 5 different target images (indexed from 4,851 to 4,855 in the original class), resulting in a total of 450 attack instances.
We focus on linear transfer learning and limit each attack to 800 iterations to meet time and resource constraints.
On average, against all eight victim networks, \sysshortThree{} achieved a success rate of 40.83\%.
In the original setting of <frog, ship>, \sysshortThree{} showed a success rate of 47.25\% (Figure~\ref{fig:single-tr-meanVictim}).
See Appendix~\ref{appendix:experimentDetails} for the attack performance against individual victim networks as well as a comparison of different class pairs.

\subsection{Multi-target Mode}
\label{sec:eval-multi}

\begin{figure*}[t]
     \centering
     \vspace{-0.4cm}
     \begin{subfigure}[b]{0.32\textwidth}
         \centering
         \includegraphics[width=\textwidth]{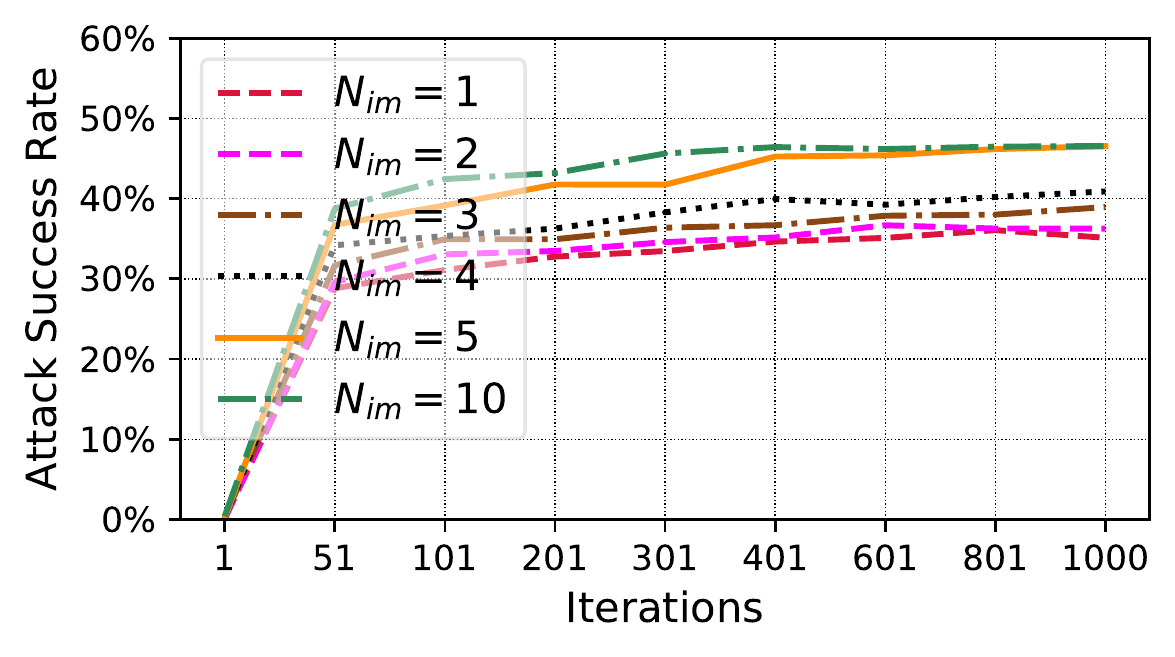}
         \vspace{-0.6cm}
         \caption{CP}
         \label{fig:multiple-tr-CP-attacksuccrate}
     \end{subfigure}
     \hfill
     \begin{subfigure}[b]{0.32\textwidth}
         \centering
         \includegraphics[width=\textwidth]{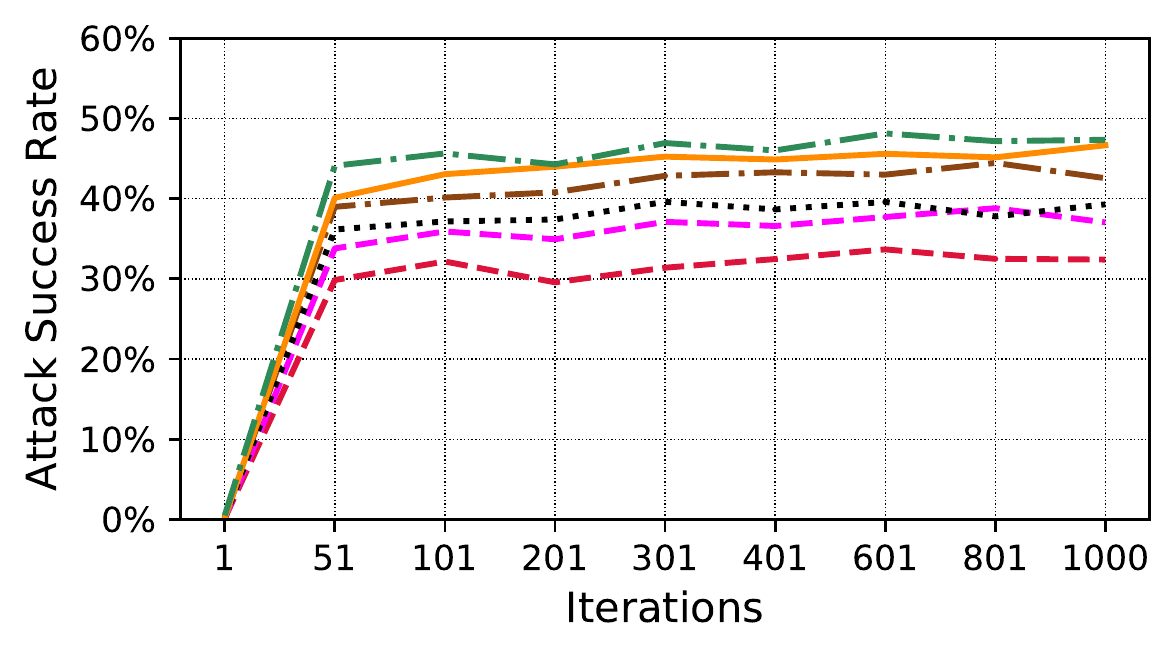}
         \vspace{-0.6cm}
         \caption{\sysshortOne{}}
         \label{fig:multiple-tr-BP-attacksuccrate}
     \end{subfigure}
     \hfill
     \begin{subfigure}[b]{0.32\textwidth}
         \centering
         \includegraphics[width=\textwidth]{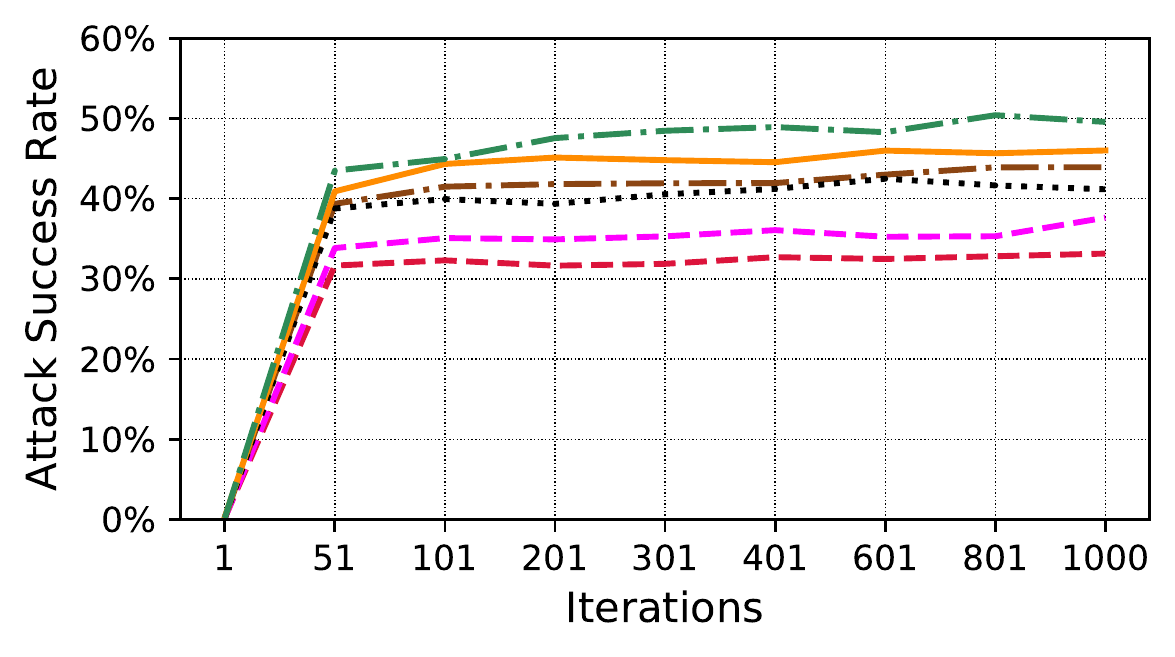}
         \vspace{-0.6cm}
         \caption{\sysshortThree{}}
         \label{fig:multiple-tr-BP3-attacksuccrate}
     \end{subfigure}
     \vspace{-0.2cm}
     \caption{Attack transferability to unseen angles in \fix{}.}
     \label{fig:multiple-tr-attacksuccrate}
\end{figure*}

\begin{figure*}[t]
    \vspace{-0.3cm}
     \centering
     \begin{subfigure}[b]{0.32\textwidth}
         \centering
         \includegraphics[width=\textwidth]{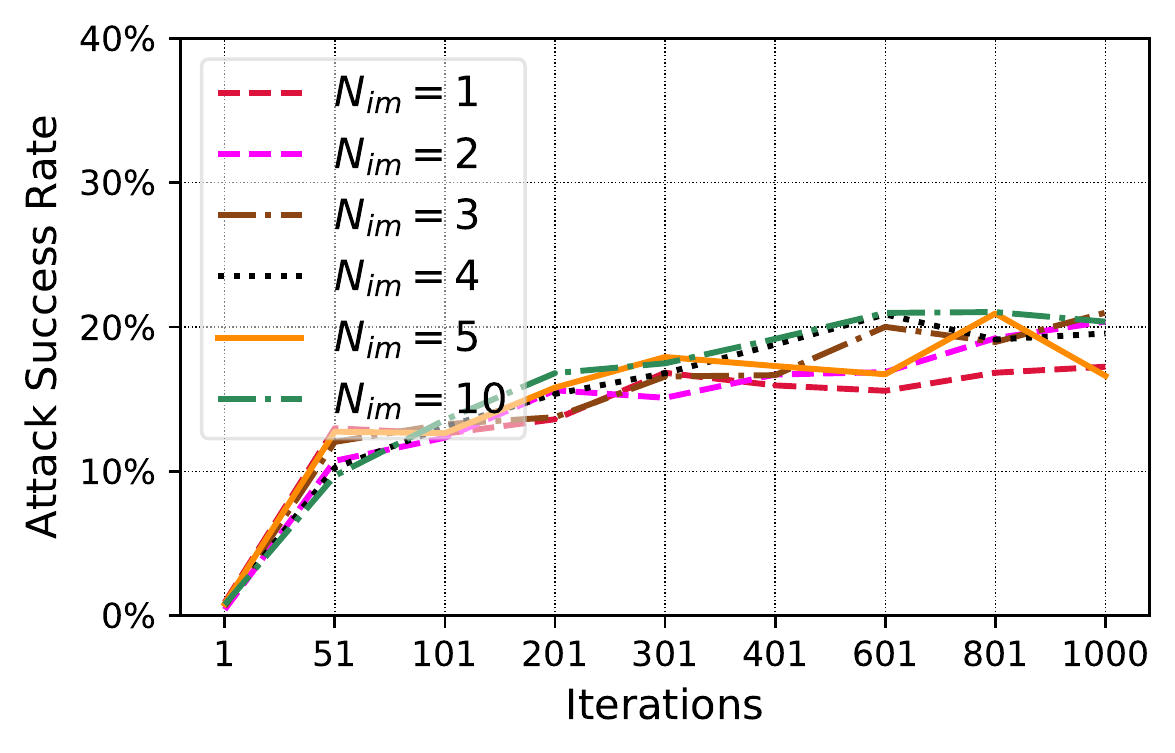}
         \vspace{-0.6cm}
         \caption{CP}
         \label{fig:multiple-end-CP-attacksuccrate}
     \end{subfigure}
     \begin{subfigure}[b]{0.32\textwidth}
         \centering
         \includegraphics[width=\textwidth]{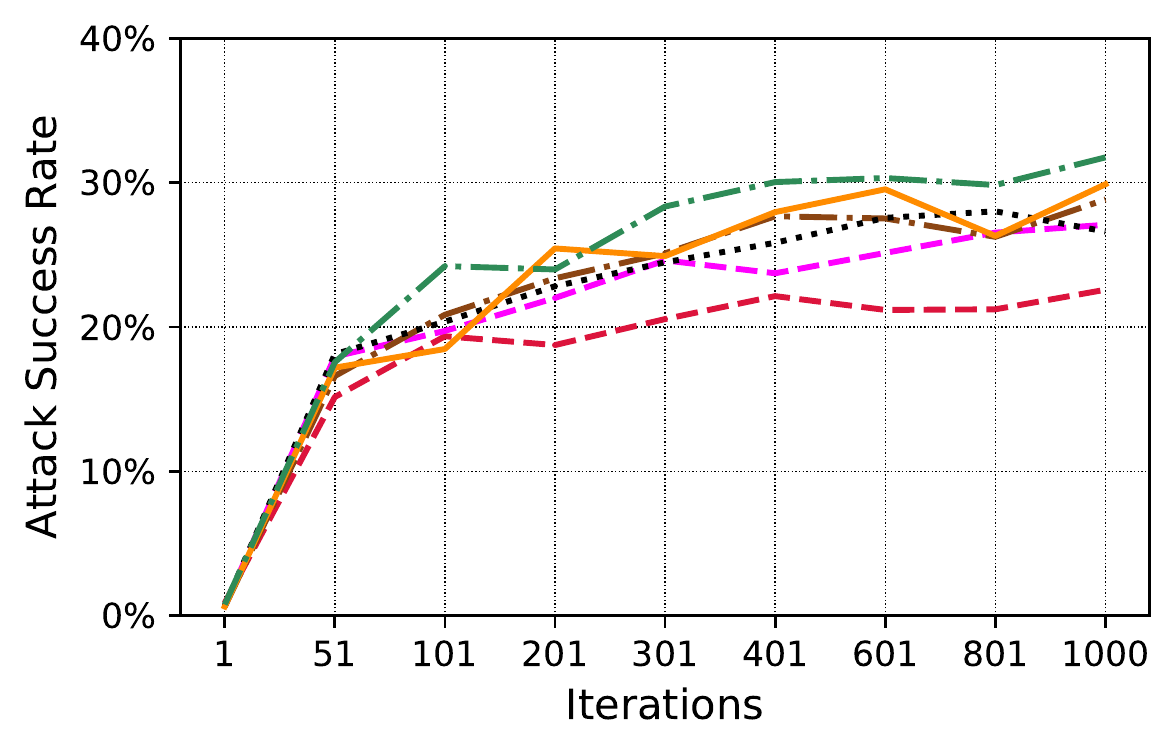}
         \vspace{-0.6cm}
         \caption{\sysshortOne{}}
         \label{fig:multiple-end-BP-attacksuccrate}
     \end{subfigure}
     \begin{subfigure}[b]{0.32\textwidth}
         \centering
         \includegraphics[width=\textwidth]{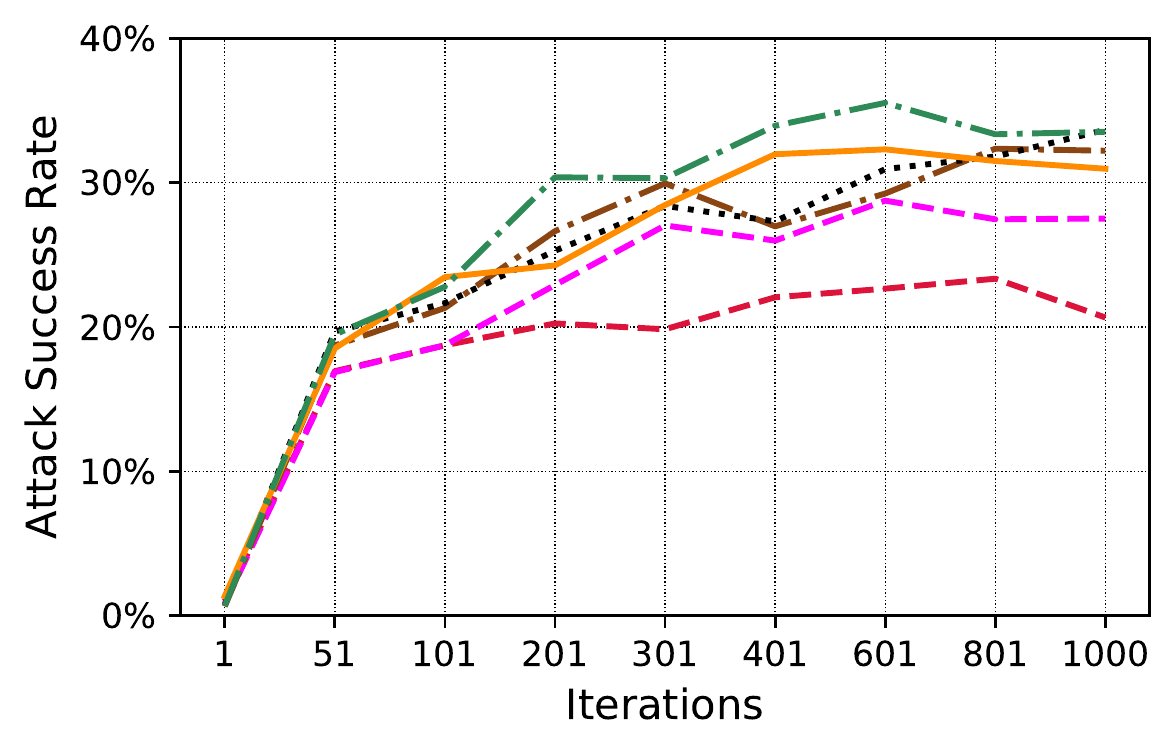}
         \vspace{-0.6cm}
         \caption{\sysshortThree{}}
         \label{fig:multiple-end-BP3-attacksuccrate}
     \end{subfigure}
     \vspace{-0.2cm}
     \caption{Attack transferability to unseen angles in \e2e{}.}
     \label{fig:multiple-end-attacksuccrate}
     \vspace{-0.2cm}
\end{figure*}

We now consider a more realistic setting where the target \textit{object} is known, but there is unpredictable variability in the target \textit{image} at test time (e.g., unknown observation angles).  
This is the first attempt at crafting a \emph{clean-label} and \emph{training-time} dataset poisoning attack that is effective on multiple (unseen) images of the target object at test time.
To this end, we consider a slight variation of \sysshortOne{} that takes multiple images of the target object (capturing as much observation variability as possible), and performs \sysshortOne{} on the averages of their feature vectors. 
We use the Multi-View Car dataset~\cite{ozuysal2009pose}, which contains images from 20 different cars as they are rotated by 360 degrees at increments of 3-4 degrees.
We expect to see lower accuracy when testing the substitute models on the Multi-View Car dataset, as it contains a different distribution of images compared to CIFAR-10.
We observed that images from the car dataset are most commonly misclassified as ``ship,'' therefore to avoid contamination from this inherent similarity, this time we choose ``frog'' as the intended misclassification label, and perform the attacks only for the 14 cars with baseline accuracy of over 90\% to obtain pessimistic results.
We use the same settings as the single-target mode.
We discuss in the Appendix~\ref{appendix:implementationDetails} how the car images of the Multi-View Car dataset are adapted for our models, which are trained on CIFAR-10.


We evaluate both CP and \sysshortOne{} setting the number of target images \(N_\text{im}\) to \(\{1,2,3,4,5,10\}\) to verify the effect of \(N_\text{im}\) on the attack robustness against unseen angles.
Note that when \(N_\text{im}=1\), the attack is in single-target mode.
To select the \(N_\text{im}\) target images, we take one image every \(\frac{360}{N_\text{im}}\) degree rotation of the target car.
Figure~\ref{fig:multiple-tr-attacksuccrate} and Figure~\ref{fig:multiple-end-attacksuccrate} show the attack success rates against \textbf{unseen} images.
In \fix{}, using five targets instead of one improves attack robustness against unseen angles by over 16\%.
In \e2e{}, \sysshortThree{} demonstrates an improvement of 12\%.
When \(N_\text{im}=5\), \sysshortOne{} achieves 14\% higher attack success rate compared to CP, while being 59x faster.
We emphasize that the total number of poison samples crafted for multi-target attacks is the same as single-target mode (i.e., 5).
Figure~\ref{fig:poisons-target-eg-multitarget} in the Appendix depicts poison samples crafted for one particular target car.

\mypar{More Realistic Transfer Learning.} We argue that the setting of this (multi-target) experiment is also relevant for another reason: the source of the fine-tuning set is different from the source of the original training set. 
In particular, we assume that the victim fine-tunes a model -- pre-trained on CIFAR-10 -- on the Multi-View Car dataset.
Our results show that BP achieves comparable success rates in such a more realistic setting.
For example, when \(N_\text{im}\!=\!1\) in \e2e{}, the attack success rates of BP and CP are 51\% and 34\%. 

\subsection{Attack Budget}
\label{sec:attack-budget}
\begin{table*}[t]
\centering
\caption{Evaluation of \sysshortOne{} (after 800 iterations), when different poison budget is used. The first row shows the accuracy that the victim's fine-tuned model classifies poison samples into the poison class label. The second row shows the baseline test accuracy of the model on the standard test set from CIFAR-10. The last row shows the attack success rate. 
}
\label{table:diffpoisonbudget}

\begin{subtable}[t]{0.47\textwidth}
	\centering
	\begin{tabularx}{0.98\linewidth}{L{10.5} | C{3} | C{3} | C{3} | C{3}}
	\multirow{2}{*}{} & \multicolumn{4}{c}{\textbf{\# Poisons}} \\ \cline{2-5}
	 & \textbf{3} & \textbf{5} & \textbf{7} & \textbf{10} \\
	\midrule
	\textbf{Poisons Acc. (\%)} 				& 82.33 		& 84.45 	& 86.57 	& 88.98 		\\
	\textbf{Clean Test Acc. (\%)} 			& 91.92 		& 91.76 	& 91.67 	& 91.60 		\\
	\textbf{Attack Success Rate (\%)} 	& 28.00 		& 42.50 	& 49.50 	& 57.75  		\\
	\end{tabularx}
	
	\vspace{0.5em}
	\caption{Different number of poisons used (\(\epsilon=0.1\)).}
	\label{table:diffnumpoisons}
\end{subtable}
\hspace{2em}
\begin{subtable}[t]{0.47\textwidth}
	\centering

	\addtolength{\tabcolsep}{-3pt}
	\begin{tabularx}{0.98\linewidth}{L{10.5} | C{2.5} | C{2.5} | C{2.5} | C{2.5} | C{2.5} | C{2.5}}
	\multirow{2}{*}{} & \multicolumn{6}{c}{\textbf{Perturbation Budget} \(\bm{\epsilon}\)} \\  \cline{2-7}
	 & \(\bm{0.01}\) & \(\bm{0.03}\) & \(\bm{0.05}\) & \(\bm{0.1}\) & \(\bm{0.2}\) & \(\bm{0.3}\) \\
	\midrule
	\textbf{Poisons Acc. (\%)} 				& 96.05 		& 82.25 	& 82.87 		& 84.45 	& 85.4 	& 86.1		\\
	\textbf{Clean Test Acc. (\%)} 			& 92.01 		& 91.69 	& 91.75 		& 91.76 	& 91.80 	& 91.82	\\
	\textbf{Attack Success Rate (\%)} 	& 4.50 		& 33.00 	& 40.43 		& 42.50 	& 39.75  	& 43.25	\\
	\end{tabularx}
	
	\vspace{0.5em}
	\caption{Different levels of perturbation \(\epsilon\) used (\# poisons = 5).}
	\label{table:pertbudget}
\end{subtable}
\vspace{-0.3cm}
\end{table*}

Until now, we have used five poison samples with an \(\ell_\infty\) perturbation budget of \(\epsilon=0.1\).
Here, we discuss the impact of the number of poison samples and the perturbation amount on the attack success rate.
Since we observed the same trend for single-target and multi-target mode, we only report the numbers for single-target mode.
We limit each attack to 800 iterations to meet time and resource constraints.

Table~\ref{table:diffnumpoisons} shows the attack performance of \sysshortOne{}, when different numbers of poison samples are injected into the victim's fine-tuning dataset.
In general, using more poison samples results in a higher attack success rate, which can be due to two reasons;
First, \sysshortOne{} achieves a lower ``bullseye'' loss (Eq.~\ref{eq:simplecvxLoss}) when more poison samples are used. 
In fact, we confirmed that this is not the case. 
While in some scenarios, the loss value slightly decreases, generally, across different target samples, the loss does not decrease by simply adding more poison samples.
So, if the attack fails to find poison samples shaping a convex polytope around some particular target, increasing the number of poison samples will not help us to find a ``better'' polytope.

Second, having more poison samples in the fine-tuning dataset will cause the classifier to learn the malicious characteristics of the poison samples with a higher probability. 
This indeed contributes to a higher attack success rate.
During our analysis, we noticed that the main reason for the attack failure for a particular target is the following;
In the fine-tuning dataset of the victim, there exist samples from the target's original class that are close ``enough'' to the target so that the victim's model classifies the target into its true class.
In most cases, a few of the poison samples are even classified into the target's original class, which indeed downgrades the malicious effect of poison samples.
Therefore, by adding more poison samples to the fine-tuning dataset, the chance that poison samples in the adjacency of the target outnumber samples from the true class is higher.
Note that we do not consider a white-box threat model in this work, thus the convex polytope created for the substitute networks will not necessarily transfer to the victim's feature space, which means the condition of the mathematical guarantee discussed in Section~\ref{sec:bg} will not always hold.

We also evaluate CP when the number of poison samples is ten. 
As Table~\ref{table:diffnumpoisons-cmp} shows, \sysshortOne{} demonstrates a 6.5\% higher attack success rate than CP.
Running \sysshortOne{} for 800 iterations takes only seven minutes on average, while CP takes 603 minutes, which is 86 times slower.
This happens because CP poorly scales as the number of poison samples increases.
In each iteration of solving Eq.~\ref{eq:cvxLoss}, CP needs to find the optimal set of coefficients for each poison.
If we increase the number of poison samples from five to ten, at each iteration of the attack, ten optimization problems need to be solved to find the best coefficients (instead of five).
This is not the case for BP, as increasing the number of poison samples does not necessarily make solving Eq.~\ref{eq:simplecvxLoss} harder.
The problem is still finding the solution of Eq.~\ref{eq:simplecvxLoss} using backpropagation, with ten poison samples as the parameters, instead of five.
In fact, our evaluation shows that \sysshortOne{} takes roughly the same time as when we use five poison samples.

Table~\ref{table:pertbudget} shows the attack performance of \sysshortOne{}, when five poison samples are crafted, yet with different levels of perturbation.
In general, the ``bullseye'' loss does not change for \(\epsilon\) values greater than 0.05, and increasing \(\epsilon\) further has a negligible impact on the attack success rate.
We argue this happens for the same reason that an attack fails for a particular target when there are some samples from the target class in the victim's fine-tuning dataset that are very close to the target in the victim's feature space.
In such a scenario, increasing the perturbation budget is not enough to move the target from the proximity of its class into the attack zone.
Due to resource and time constraints, we evaluated CP in these settings on a smaller set of targets, and we have observed a trend similar to what we discussed above.

\begin{table}[]
\centering
\caption{Evaluation of \sysshortOne{} and CP (after 800 iterations), when ten poison samples are used, and \(\epsilon\) is set to 0.1.}
\label{table:diffnumpoisons-cmp}
\begin{tabularx}{0.88\linewidth}{L{15} | C{4} | C{4}}
 & \textbf{\sysshortOne{}} & \textbf{CP} \\
\midrule
\textbf{Poisons Acc. (\%)} 					& 88.98 		& 85.20  		\\
\textbf{Clean Test Acc. (\%)} 				& 91.60 		& 91.43  		\\
\textbf{Attack Success Rate (\%)} 		& 57.75 		& 51.25 		\\
\textbf{Attack Execution Time (min.)} & 7				& 	603			\\

\end{tabularx}
\vspace{-0.3cm}
\end{table}

\subsection{Defenses}
\label{sec:eval-defenses}
\begin{table*}[p]
\vspace*{-0.4cm}
\centering
\caption{Evaluation of \sysshortOne{} and CP (after 800 iterations) when the victim employs the \knndef{} defense. 
Note that \(k=0\) means no defense is employed. 
Five and ten poison samples are used in the left and right table, respectively.}
\label{table:knndefense}

\begin{subtable}[]{0.47\textwidth}
	\centering
	\addtolength{\tabcolsep}{-2.4pt}
	\begin{tabularx}{0.98\linewidth}{L{2} | C{3.4} C{3.4} | C{3.4} C{3.4} | C{3.6} C{3.6}}
	\multicolumn{1}{c|}{\multirow{2}{*}{\textbf{k}}} & \multicolumn{2}{c|}{\textbf{\# Deleted Poisons}} & \multicolumn{2}{c|}{\textbf{\# Deleted Samples}} & \multicolumn{2}{c}{\textbf{Adv. Success Rate (\%)}} \\ 
	 & \textbf{\sysshortOne{}} & \textbf{CP} & \textbf{\sysshortOne{}} & \textbf{CP} & \textbf{\sysshortOne{}} & \textbf{CP} \\
	 \midrule
	\textbf{0} & \multicolumn{1}{c|}{-} & - & \multicolumn{1}{c|}{-} & - & \multicolumn{1}{c|}{42.5} & 37.25 \\
	\midrule
	\textbf{1} & \multicolumn{1}{c|}{3.18} & 4.28 & \multicolumn{1}{c|}{36.46} & 37.02 & \multicolumn{1}{c|}{20.50} & 6.75 \\
	\textbf{2} & \multicolumn{1}{c|}{2.42} & 3.86 & \multicolumn{1}{c|}{21.91} & 23.07 & \multicolumn{1}{c|}{24.75} & 8.00 \\
	\textbf{3} & \multicolumn{1}{c|}{3.81} & 4.66 & \multicolumn{1}{c|}{27.86} & 27.87 & \multicolumn{1}{c|}{11.75} & 1.50 \\
	\textbf{4} & \multicolumn{1}{c|}{3.48} & 4.60 & \multicolumn{1}{c|}{25.83} & 26.69 & \multicolumn{1}{c|}{14.75} & 2.50 \\
	\textbf{6} & \multicolumn{1}{c|}{4.22} & 4.85 & \multicolumn{1}{c|}{25.39} & 25.91 & \multicolumn{1}{c|}{8.25} & 1.25 \\
	\textbf{8} & \multicolumn{1}{c|}{4.77} & 4.94 & \multicolumn{1}{c|}{25.69} & 25.80 & \multicolumn{1}{c|}{1.25} & 0.00 \\
	\textbf{10} & \multicolumn{1}{c|}{4.97} & 4.95 & \multicolumn{1}{c|}{26.36} & 26.33 & \multicolumn{1}{c|}{0.00} & 0.25 \\
	\textbf{12} & \multicolumn{1}{c|}{4.98} & 4.96 & \multicolumn{1}{c|}{26.58} & 26.54 & \multicolumn{1}{c|}{0.00} & 0.00 \\
	\textbf{14} & \multicolumn{1}{c|}{4.98} & 4.96 & \multicolumn{1}{c|}{26.21} & 26.21 & \multicolumn{1}{c|}{0.00} & 0.00 \\
	\textbf{16} & \multicolumn{1}{c|}{4.98} & 4.96 & \multicolumn{1}{c|}{26.95} & 26.92 & \multicolumn{1}{c|}{0.00} & 0.00 \\
	\textbf{18} & \multicolumn{1}{c|}{4.98} & 4.96 & \multicolumn{1}{c|}{26.36} & 26.37 & \multicolumn{1}{c|}{0.00} & 0.00 \\
	\textbf{22} & \multicolumn{1}{c|}{4.98} & 4.96 & \multicolumn{1}{c|}{26.62} & 26.59 & \multicolumn{1}{c|}{0.00} & 0.00 \\

	\end{tabularx}
	\vspace{0.3em}
	\caption{\textbf{\# Poisons = 5}}
	\label{table:knndefense-5poisons}

\end{subtable}
\hspace{2em}
\begin{subtable}[]{0.47\textwidth}
	\centering
	\addtolength{\tabcolsep}{-2.4pt}
	\begin{tabularx}{0.98\linewidth}{L{2} | C{3.4} C{3.4} | C{3.4} C{3.4} | C{3.6} C{3.6}}
	\multicolumn{1}{c|}{\multirow{2}{*}{\textbf{k}}} & \multicolumn{2}{c|}{\textbf{\# Deleted Poisons}} & \multicolumn{2}{c|}{\textbf{\# Deleted Samples}} & \multicolumn{2}{c}{\textbf{Adv. Success Rate (\%)}} \\ 
	 & \textbf{\sysshortOne{}} & \textbf{CP} & \textbf{\sysshortOne{}} & \textbf{CP} & \textbf{\sysshortOne{}} & \textbf{CP} \\
	 \midrule
	\textbf{0} & \multicolumn{1}{c|}{-} & - & \multicolumn{1}{c|}{-} & - & \multicolumn{1}{c|}{57.75} & 51.25 \\
	\midrule
	
	\textbf{1} & \multicolumn{1}{c|}{4.30} & 7.56 & \multicolumn{1}{c|}{38.77} & 41.22 & \multicolumn{1}{c|}{49.25} & 14.00 \\
	\textbf{2} & \multicolumn{1}{c|}{2.71} & 6.38 & \multicolumn{1}{c|}{22.75} & 25.77 & \multicolumn{1}{c|}{51.75} & 21.25 \\
	\textbf{3} & \multicolumn{1}{c|}{4.92} & 8.16 & \multicolumn{1}{c|}{30.36} & 31.88 & \multicolumn{1}{c|}{38.75} & 11.00 \\
	\textbf{4} & \multicolumn{1}{c|}{3.94} & 7.76 & \multicolumn{1}{c|}{26.74} & 29.72 & \multicolumn{1}{c|}{46.75} & 12.50 \\
	\textbf{6} & \multicolumn{1}{c|}{4.82} & 8.51 & \multicolumn{1}{c|}{26.57} & 29.44 & \multicolumn{1}{c|}{40.00} & 7.25 \\
	\textbf{8} & \multicolumn{1}{c|}{5.68} & 9.03 & \multicolumn{1}{c|}{27.24} & 29.87 & \multicolumn{1}{c|}{31.25} & 3.25 \\
	\textbf{10} & \multicolumn{1}{c|}{6.53} & 9.31 & \multicolumn{1}{c|}{28.30} & 30.54 & \multicolumn{1}{c|}{26.50} & 2.25 \\
	\textbf{12} & \multicolumn{1}{c|}{7.42} & 9.44 & \multicolumn{1}{c|}{29.19} & 30.82 & \multicolumn{1}{c|}{17.75} & 1.25 \\
	\textbf{14} & \multicolumn{1}{c|}{8.17} & 9.54 & \multicolumn{1}{c|}{29.42} & 30.54 & \multicolumn{1}{c|}{15.25} & 0.25 \\
	\textbf{16} & \multicolumn{1}{c|}{8.86} & 9.59 & \multicolumn{1}{c|}{30.63} & 31.20 & \multicolumn{1}{c|}{8.00} & 0.00 \\
	\textbf{18} & \multicolumn{1}{c|}{9.50} & 9.61 & \multicolumn{1}{c|}{30.60} & 30.63 & \multicolumn{1}{c|}{3.00} & 0.00 \\
	\textbf{22} & \multicolumn{1}{c|}{9.91} & 9.61 & \multicolumn{1}{c|}{31.18} & 30.85 & \multicolumn{1}{c|}{0.25} & 0.00 \\

	\end{tabularx}
	\vspace{0.3em}
	\caption{\textbf{\# Poisons = 10}}
	\label{table:knndefense-10poisons}

\end{subtable}
\vspace*{-0.3cm}
\end{table*}

\begin{table*}[p]
\centering
\caption{Evaluation of \sysshortOne{} and CP when the victim employs the \l2def{} defense. 
}
\label{table:l2defense}

\begin{subtable}[]{0.47\textwidth}
	\centering
	\addtolength{\tabcolsep}{-2.4pt}
	\begin{tabularx}{0.98\linewidth}{L{2} | C{3.4} C{3.4} | C{3.4} C{3.4} | C{3.6} C{3.6}}
	\multicolumn{1}{c|}{\multirow{2}{*}{\(\bm{\mu\)}}} & \multicolumn{2}{c|}{\textbf{\# Deleted Poisons}} & \multicolumn{2}{c|}{\textbf{\# Deleted Samples}} & \multicolumn{2}{c}{\textbf{Adv. Success Rate (\%)}} \\ 
	 & \textbf{\sysshortOne{}} & \textbf{CP} & \textbf{\sysshortOne{}} & \textbf{CP} & \textbf{\sysshortOne{}} & \textbf{CP} \\
	\midrule
	\textbf{0.00} & \multicolumn{1}{c|}{-} & - & \multicolumn{1}{c|}{-} & - & \multicolumn{1}{c|}{42.5} & 37.25 \\
	\midrule
	\textbf{0.02} & \multicolumn{1}{c|}{1.00} & 1.00 & \multicolumn{1}{c|}{10.00} & 10.00 & \multicolumn{1}{c|}{35.00} & 30.25 \\
	\textbf{0.04} & \multicolumn{1}{c|}{2.00} & 2.00 & \multicolumn{1}{c|}{20.00} & 20.00 & \multicolumn{1}{c|}{30.50} & 19.00 \\
	\textbf{0.06} & \multicolumn{1}{c|}{3.00} & 3.00 & \multicolumn{1}{c|}{30.00} & 30.00 & \multicolumn{1}{c|}{17.25} & 7.75 \\
	\textbf{0.08} & \multicolumn{1}{c|}{3.99} & 3.99 & \multicolumn{1}{c|}{40.00} & 40.00 & \multicolumn{1}{c|}{4.75} & 1.75 \\
	\textbf{0.10} & \multicolumn{1}{c|}{4.96} & 4.93 & \multicolumn{1}{c|}{50.00} & 50.00 & \multicolumn{1}{c|}{0.25} & 0.75 \\
	\textbf{0.12} & \multicolumn{1}{c|}{4.99} & 4.98 & \multicolumn{1}{c|}{60.00} & 60.00 & \multicolumn{1}{c|}{0.00} & 0.00 \\
	\textbf{0.14} & \multicolumn{1}{c|}{4.99} & 4.98 & \multicolumn{1}{c|}{70.00} & 70.00 & \multicolumn{1}{c|}{0.00} & 0.00 \\
	\textbf{0.16} & \multicolumn{1}{c|}{5.00} & 4.98 & \multicolumn{1}{c|}{80.00} & 80.00 & \multicolumn{1}{c|}{0.00} & 0.00 \\
	\textbf{0.18} & \multicolumn{1}{c|}{5.00} & 4.99 & \multicolumn{1}{c|}{90.00} & 90.00 & \multicolumn{1}{c|}{0.00} & 0.00 \\
	\textbf{0.20} & \multicolumn{1}{c|}{5.00} & 4.99 & \multicolumn{1}{c|}{100.00} & 100.00 & \multicolumn{1}{c|}{0.50} & 0.00 \\
	\end{tabularx}
	\vspace{0.3em}
	\caption{\textbf{\# Poisons = 5}}
	\label{table:l2defense-5poisons}

\end{subtable}
\hspace{2em}
\begin{subtable}[]{0.47\textwidth}
	\centering
	\addtolength{\tabcolsep}{-2.4pt}
	\begin{tabularx}{0.98\linewidth}{L{2} | C{3.4} C{3.4} | C{3.4} C{3.4} | C{3.6} C{3.6}}
	\multicolumn{1}{c|}{\multirow{2}{*}{\(\bm{\mu\)}}} & \multicolumn{2}{c|}{\textbf{\# Deleted Poisons}} & \multicolumn{2}{c|}{\textbf{\# Deleted Samples}} & \multicolumn{2}{c}{\textbf{Adv. Success Rate (\%)}} \\ 
	 & \textbf{\sysshortOne{}} & \textbf{CP} & \textbf{\sysshortOne{}} & \textbf{CP} & \textbf{\sysshortOne{}} & \textbf{CP} \\
	\midrule
	\textbf{0.00} & \multicolumn{1}{c|}{-} & - & \multicolumn{1}{c|}{-} & - & \multicolumn{1}{c|}{57.75} & 51.25 \\
	\midrule
	\textbf{0.02} & \multicolumn{1}{c|}{1.00} & 1.00 & \multicolumn{1}{c|}{10.00} & 10.00 & \multicolumn{1}{c|}{55.00} & 47.25 \\
	\textbf{0.04} & \multicolumn{1}{c|}{2.00} & 2.00 & \multicolumn{1}{c|}{20.00} & 20.00 & \multicolumn{1}{c|}{53.25} & 45.50 \\
	\textbf{0.06} & \multicolumn{1}{c|}{3.00} & 3.00 & \multicolumn{1}{c|}{30.00} & 30.00 & \multicolumn{1}{c|}{49.50} & 40.25 \\
	\textbf{0.08} & \multicolumn{1}{c|}{4.00} & 4.00 & \multicolumn{1}{c|}{40.00} & 40.00 & \multicolumn{1}{c|}{43.50} & 34.00 \\
	\textbf{0.10} & \multicolumn{1}{c|}{5.00} & 5.00 & \multicolumn{1}{c|}{50.00} & 50.00 & \multicolumn{1}{c|}{37.50} & 21.50 \\
	\textbf{0.12} & \multicolumn{1}{c|}{6.00} & 6.00 & \multicolumn{1}{c|}{60.00} & 60.00 & \multicolumn{1}{c|}{32.50} & 17.00 \\
	\textbf{0.14} & \multicolumn{1}{c|}{7.00} & 7.00 & \multicolumn{1}{c|}{70.00} & 70.00 & \multicolumn{1}{c|}{22.25} & 8.25 \\
	\textbf{0.16} & \multicolumn{1}{c|}{8.00} & 8.00 & \multicolumn{1}{c|}{80.00} & 80.00 & \multicolumn{1}{c|}{11.75} & 4.75 \\
	\textbf{0.18} & \multicolumn{1}{c|}{8.99} & 9.00 & \multicolumn{1}{c|}{90.00} & 90.00 & \multicolumn{1}{c|}{3.00} & 0.75 \\
	\textbf{0.20} & \multicolumn{1}{c|}{9.93} & 9.56 & \multicolumn{1}{c|}{100.00} & 100.00 & \multicolumn{1}{c|}{0.25} & 0.00 \\

	\end{tabularx}
	\vspace{0.3em}
	\caption{\textbf{\# Poisons = 10}}
	\label{table:l2defense-10poisons}

\end{subtable}
\vspace*{-0.2cm}
\end{table*}

\begin{figure*}[p]
     \centering
     \begin{subfigure}[b]{0.4\textwidth}
         \centering
         \includegraphics[width=\textwidth]{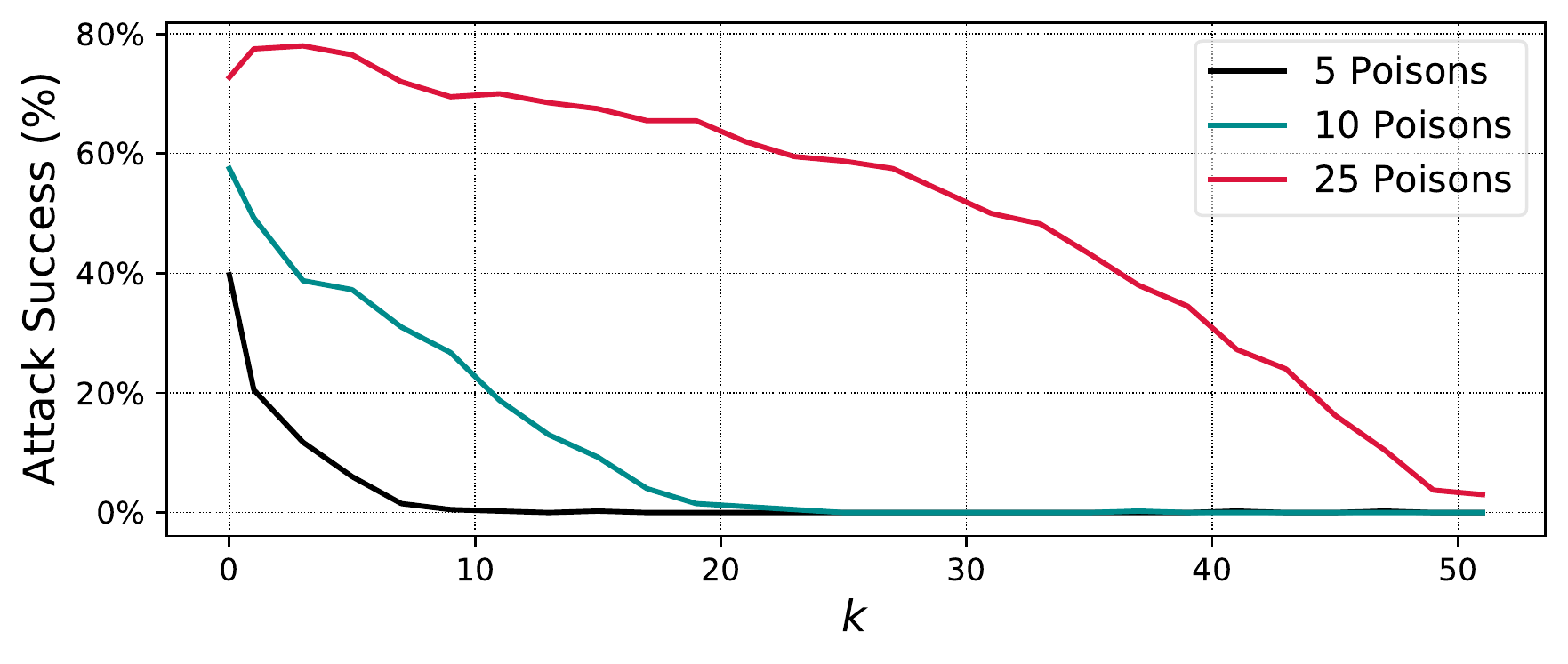}
         \vspace*{-0.6cm}
         \caption{\sysshortOne{} vs. \knndef{}}
         \label{fig:knn-diffnumpoisons-attack-succ-rate}
     \end{subfigure}
     \hfill
     \begin{subfigure}[b]{0.4\textwidth}
         \centering
         \includegraphics[width=\textwidth]{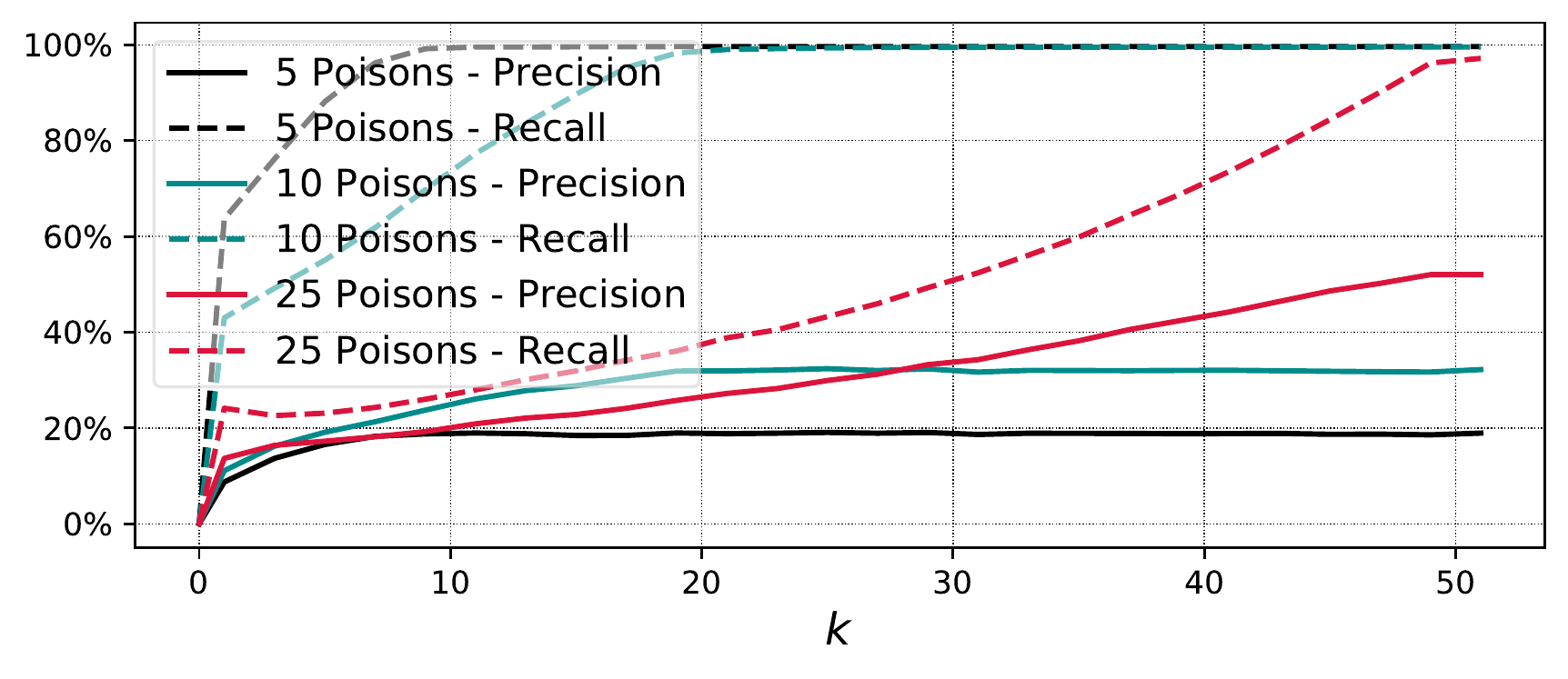}
         \vspace*{-0.6cm}
         \caption{Precision and Recall of the \knndef{} defense.}
         \label{fig:knn-diffnumpoisons-defense-rates}
     \end{subfigure}
     
     \begin{subfigure}[b]{0.4\textwidth}
         \centering
         \includegraphics[width=\textwidth]{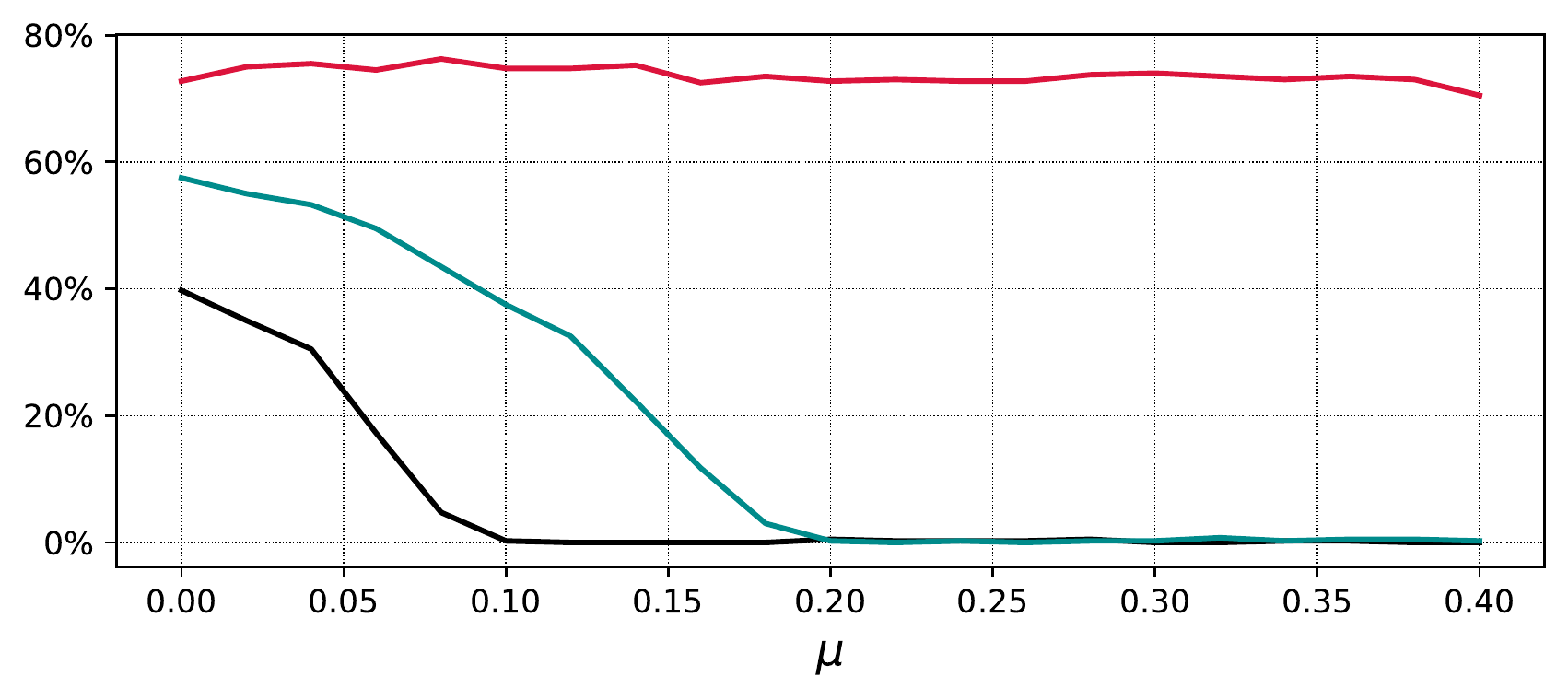}
         \vspace*{-0.6cm}
         \caption{\sysshortOne{} vs. \l2def{}}
        \label{fig:l2def-diffnumpoisons-attack-succ-rate}
     \end{subfigure}
     \hfill
     \begin{subfigure}[b]{0.4\textwidth}
         \centering
         \includegraphics[width=\textwidth]{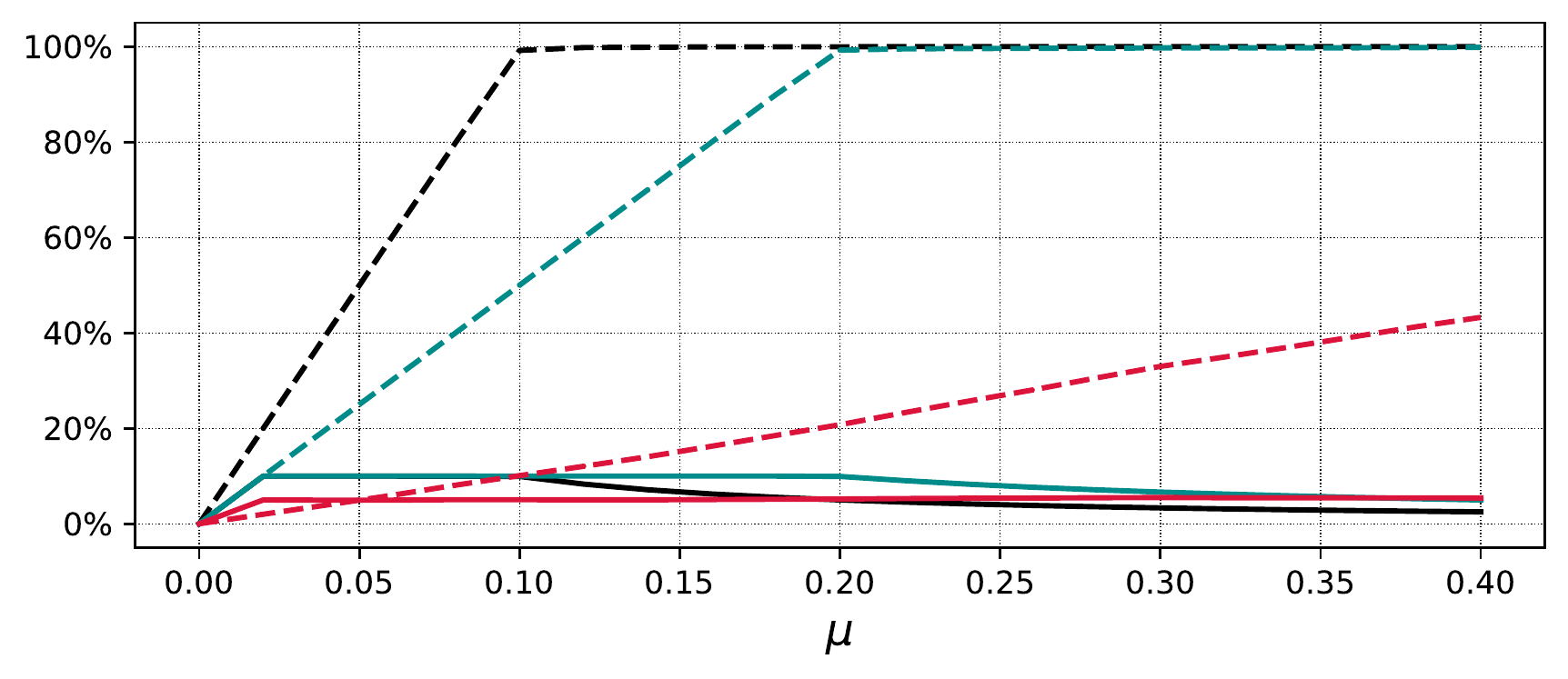}
         \vspace*{-0.6cm}
         \caption{Precision and Recall of the \l2def{} defense.}
         \label{fig:l2def-diffnumpoisons-defense-rates}
     \end{subfigure}
     \vspace*{-0.1cm}
     \caption{Evaluation of \sysshortOne{} against \knndef{} and \l2def{} defenses, when 5, 10, or 25 poison samples are used.}
     \label{fig:defenses}
     \vspace*{-0.3cm}
\end{figure*}
\begin{figure*}[p]
     \centering
     \begin{subfigure}[b]{0.4\textwidth}
         \centering
         \includegraphics[width=\textwidth]{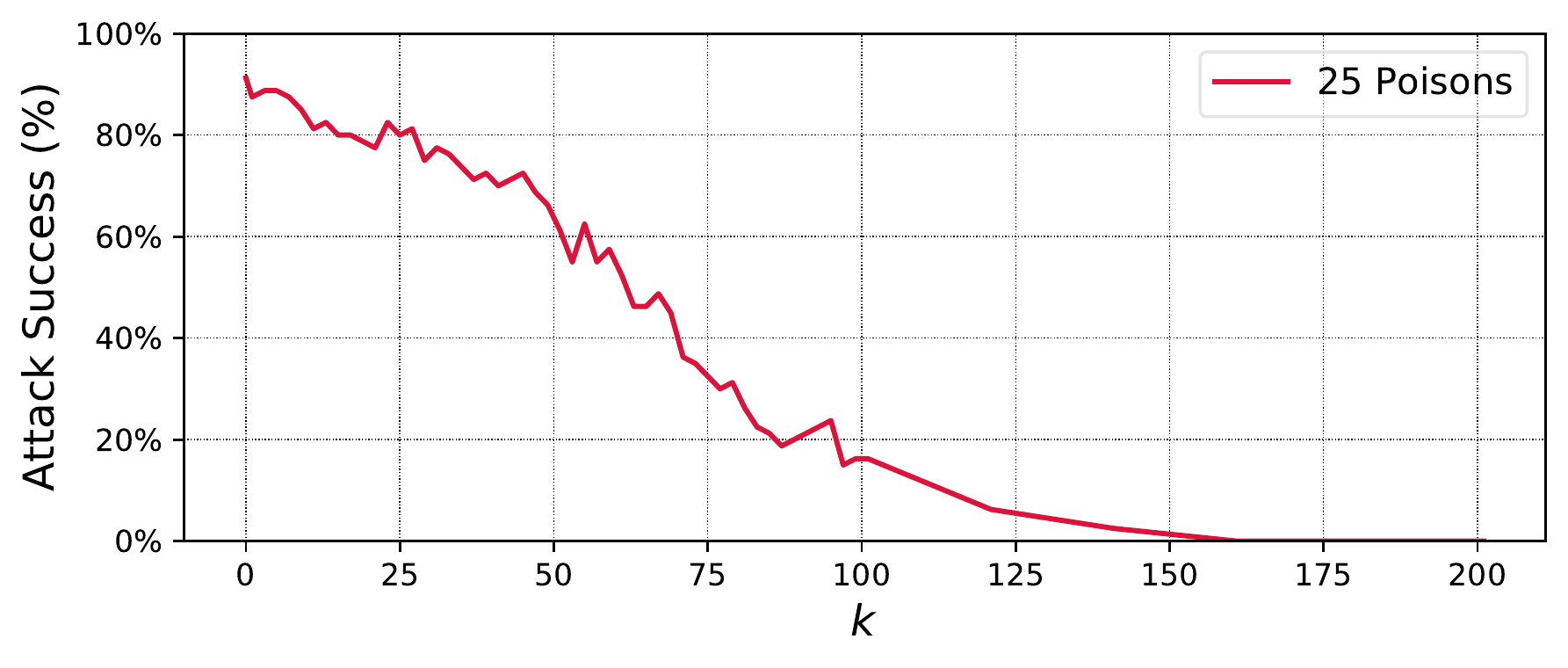}
     \end{subfigure}
     \hfill
     \begin{subfigure}[b]{0.4\textwidth}
         \centering
         \includegraphics[width=\textwidth]{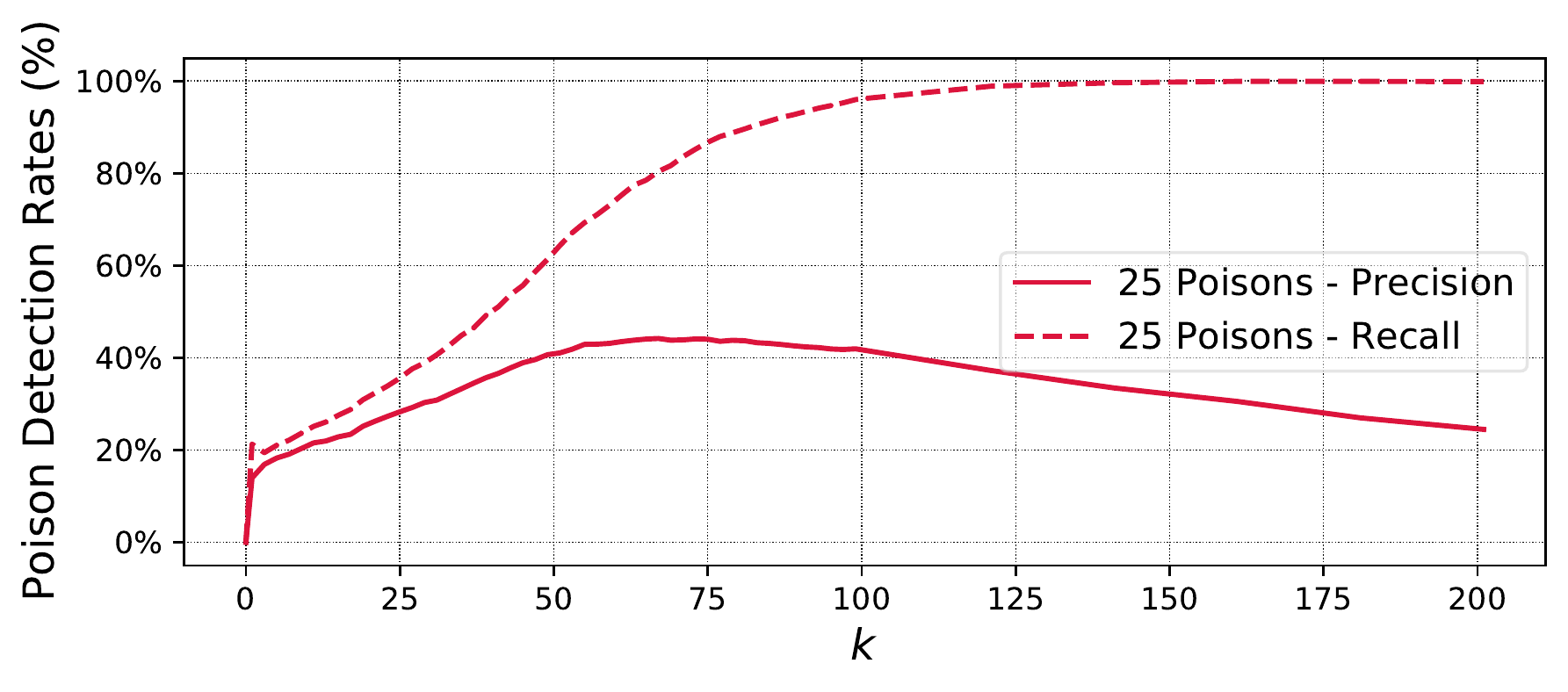}
     \end{subfigure}
     \vspace*{-0.3cm}
     \caption{Evaluation of \sysshortOne{} against the \knndef{} defense, when the target is classless.}
     \label{fig:defenses-noclasstarget}
     \vspace*{-0.3cm}
\end{figure*}

Concurrent to this work, a recent study has been published~\cite{peri2020deep}, which studies defenses against clean-label poisoning attacks, i.e., Feature Collision~\cite{shafahi2018poison} and Convex Polytope~\cite{zhu2019transferable}.
In their evaluation, \knndef{} and \l2def{} defenses generally outperformed other types of defenses, such as adversarial training. 
In this work, we evaluate both \sysshortOne{} and CP against these two defenses.
\textbf{Deep k-NN Defense:} For each sample in the training set, this defense flags the sample as anomalous and discards it from the training set if the point's label is not the mode amongst the labels of its \emph{k} nearest neighbors.
Euclidean distance is used to measure the distance between data points in feature space.
\textbf{\(\mathbf{l_2\textrm{-norm}}\) Outlier Defense:} For each class \(c\), the \l2def{} defense removes a fraction \(\mu\)  of points from class \(c\) that are farthest in feature space from their centroid.

It should be noted that both defenses are vulnerable to data-poisoning attacks.
In the \knndef{} defense, a na\"ive adversary might expand the set of poison samples such that the extra poison samples are close ``enough'' to the old poison samples, so that more poison samples might survive the k nearest neighbor filtration process.
In \l2def{} defense, the position of the centroid can be adjusted towards the poison samples (e.g., by adding more poison samples), especially when the per-class data size is small, which is the case in transfer learning.
While clean-label poisoning attacks can be more powerful by considering neighborhood conformity tests when crafting the poison samples, in this work, we assume the adversary does not know that such defenses will be employed by the victim. 
In particular, we evaluate both \sysshortOne{} and CP against these two defenses.
In our evaluation, we ran \sysshortOne{} and CP against \fix{} for 50 different targets, which results in 50 different sets of poison samples.
We report here the aggregated statistics averaged over these sets of poison samples and eight victim models.
We also evaluated the attacks when the number of poison samples is increased from five to ten.
To meet resource and time constraints, the attacks are limited to 800 iterations.

Table~\ref{table:knndefense} shows the performance of the \knndef{} defense against \sysshortOne{} and CP for various choices of \(k\).
Regardless of how many poison samples are used, the \knndef{} defense becomes more effective against both attacks as \(k\) increases, while eliminating roughly the same number of samples from the training set (i.e., 26 and 31 for when five and ten poison samples are crafted, respectively).
\sysshortOne{} generally demonstrates much higher resilience compared to CP.
For small values of \(k\), the \knndef{} defense discards fewer poison samples of BP compared to CP.
When using five poison samples, setting \(k=1\) is enough to reduce the attack success rate of CP from 37.25\% to 6.75\%, while \sysshortOne{} still achieves an attack success rate of 20.50\%, which is 4.75x higher.
To completely diminish \sysshortOne{}, \(k\) needs to be greater than eight, however, when ten poison samples are crafted, the attack success rate decreases only to 31.25\%.
It is worth noting that CP achieves 1.25\% attack success rate in such a configuration.

Table~\ref{table:l2defense} presents the performance of the \l2def{} defense against \sysshortOne{} and CP for various choices of \(\mu\).
When five poison samples are used, \sysshortOne{} demonstrates a superior resilience against the defense compared to CP for \(\mu < 0.1\).
For larger values of \(\mu\), both attacks are completely thwarted.
However, the larger \(\mu\) is, the more samples are discarded from the training set, which can degrade the model performance on the fine-tuning dataset, and, henceforth, the new task.
For example, when \(\mu=0.1\), the \l2def{} defense eliminates five samples from each class of the dataset.
This represents 10\% of the fine-tuning dataset.
Compared to the \knndef{} defense, the \l2def{} defense tends to eliminate more samples from the dataset to achieve the same level of resilience.
In particular, to completely mitigate the attacks, the \l2def{} defense removes 50 samples, while \knndef{} eliminates 26 samples.
When ten poison samples are used, the \l2def{} defense becomes less effective for small values of \(\mu\).
To completely mitigate the attacks, \(\mu\) needs to be greater than 0.18.
In this setting, the \l2def{} defense removes 90 samples in total, which is 18\% of the fine-tuning dataset.
For smaller values of \(\mu\), \sysshortOne{} is more resilient than CP.
For example, when \(\mu=0.12\) (i.e., 60 samples to be removed from the victim's dataset), the attack success rate of CP reduces to 20\%, while BP demonstrates a 32.50\% attack success rate.

In general, \sysshortOne{} demonstrates higher attack robustness against \knndef{} and \l2def{} defenses compared to CP.
Both defenses completely mitigate the attacks for high values of \(k\) and \(\mu\).
Increasing the number of poison samples makes the \l2def{} defense ineffective, as it needs to aggressively prune the dataset, which will result in lower performance on the victim's task.
This gives \sysshortOne{} a major advantage, as unlike CP, \sysshortOne{} is able to incorporate more poison samples into the attack process, with virtually no cost in attack-execution time (Table~\ref{table:diffnumpoisons-cmp}).
On the other hand, the \knndef{} defense seems to be quite effective, even when more poison samples are used.
Increasing the number of poison samples from five to ten makes this defense remove five more samples on average.
We should note that both attacks are completely mitigated after eliminating 6\% of the victim's dataset, of which 4\% are clean samples.
The precision of poison detection is still low (\(\sim33\%\)).
To further see the effect of the number of poison samples on the precision and recall of poison detection, we evaluated \sysshortOne{}, when crafting 25 poison samples.
Figure~\ref{fig:defenses} shows the performance of \knndef{} and \l2def{} defenses against \sysshortOne{} when the number of poison samples increases from five to ten and then to 25.
Figure~\ref{fig:l2def-diffnumpoisons-attack-succ-rate} demonstrates that the \l2def{} defense is not a plausible choice.
When 25 poison samples are used, removing 40\% of the dataset reduces the success rate of \sysshortOne{} from 75\% to 70\%.
This happens because as more clean samples are removed from the dataset, the poison samples will play a more important role in the training process. 

As Figure~\ref{fig:knn-diffnumpoisons-defense-rates} shows, the poison recall rate of the \knndef{} defense reaches 100\% as \(k\) becomes about two times the number of poison samples.
This is not surprising, as it is almost impossible for the poison label to be identified as the plurality among samples in the neighborhood of the target in such a case. 
On the other hand, this defense is likely to fail if the number of poison samples is large enough to overwhelm the conformity test for each poison sample.
This will happen with high probability when the number of poison samples is larger than the number of data points in the target's true class.
In this case, the majority (or plurality) of points in the neighborhood of each poison sample will likely have the same label as the poison itself.
In fact, we observed that when samples in the target's class are fewer than the number of poisons in the fine-tuning set, the poison samples pass the test undetected in most cases, hence, the attack remains active.
Furthermore, if the target is classless, i.e., does not belong to any of the classes in the training set, the defense becomes less effective, as the poison samples surrounding the target are no longer part of a cluster related to the target's class.
To evaluate this claim, we selected the first ten images of the 102 Category Flower dataset~\cite{nilsback2008automated} as the targets, with ``ship'' being the misclassification class.
As Figure~\ref{fig:defenses-noclasstarget} shows, setting \(k\) to 50 reduces the attack success rate to 66\% for a classless target, whereas for a target from CIFAR-10 the attack is fully mitigated (Figure~\ref{fig:knn-diffnumpoisons-attack-succ-rate}). 
Complete mitigation of the attack requires \(k > 150\), which results in discarding more than 70 samples from the fine-tuning set, of which 45 are clean.

\begin{table*}[t]
\vspace{-0.3cm}
\centering
\caption{Success rates (\%) of six attacks that are evaluated in the benchmark study~\cite{schwarzschild2020just}. The poison samples are not shared for some experiments, thus, we reported the exact numbers from the study. For example, in the black-box scenario of TinyImageNet benchmarks, the original paper reported the attack success rate, averaged over when ResNet-34 or MobileNetV2 networks are used. Therefore, we were not able to present individual attack success rates for these two settings.}
\vspace{-0.1cm}
\label{table:benchmark}
\addtolength{\tabcolsep}{-5.2pt}
\begin{tabular}{l|rrrrrr|rr||r|r}
\multicolumn{1}{l}{} & \multicolumn{8}{c||}{ \textbf{Linear Transfer Learning} } & \multicolumn{2}{c}{\textbf{Training From Scratch} } \\
\cmidrule{2-11}
& & \multicolumn{5}{c|}{\textbf{CIFAR-10} } & \multicolumn{2}{c||}{\textbf{TinyImageNet} } & \multicolumn{1}{c|}{\textbf{CIFAR-10} } & \multicolumn{1}{c}{\textbf{TinyImageNet} } \\
\cmidrule{2-11}
\multicolumn{1}{c|}{} & \multicolumn{1}{c|}{\textbf{White-box}} & \multicolumn{1}{c|}{\textbf{Gray-box} } & \multicolumn{4}{c|}{\textbf{Black-box} } & \multicolumn{1}{c|}{\textbf{White-box} } & \multicolumn{1}{c||}{\textbf{Black-box} } & \multicolumn{1}{c|}{} & \multicolumn{1}{c}{} \\
\multicolumn{1}{c|}{\textbf{Attack} } & \multicolumn{1}{c|}{\textbf{ResNet18}} & \multicolumn{1}{c|}{\textbf{ResNet18} } & \multicolumn{1}{c|}{\textbf{ResNet34} } & \multicolumn{1}{c|}{\textbf{ResNet50} } & \multicolumn{1}{c|}{\textbf{VGG11} } & \multicolumn{1}{c|}{\textbf{MobileNetV2} } & \multicolumn{1}{c|}{\textbf{VGG16} } & \multicolumn{1}{c||}{\textbf{$\frac{\texttt{ResNet34}+\texttt{MobileNetV2}}{\mathbf{2}}$} } & \multicolumn{1}{c|}{\textbf{$\frac{\texttt{VGG16} + \texttt{ResNet34}+\texttt{MobileNetV2}}{\mathbf{3}}$} } & \multicolumn{1}{c}{\textbf{VGG16} } \\
\cmidrule{1-11}
\textbf{FC} & 22 & 6 & 4 & 4 & 7 & 7 & 49 & 2 & 1.33 & 4 \\
\textbf{CP} & 33 & 7 & 5 & 4 & 8 & 7 & 14 & 1 & 0.67 & 0 \\
\textbf{BP} & \textbf{85}& \textbf{10} & \textbf{8} & \textbf{6} & 9 & 7 & \textbf{100} & \textbf{10.5} & 2.33 & \textbf{44} \\
\textbf{WiB} & - & - & - & - & - & - & - & - & \textbf{26} & 32 \\
\textbf{CLBD} & 5 & 5 & 4 & 4 & 7 & 6 & 3 & 1 & 1 & 0 \\
\textbf{HTBD} & 10 & 6 & 6 & 3 & \textbf{14} & 6 & 3 & 0.5 & 2.67 & 0
\end{tabular}
\vspace{-0.3cm}
\end{table*}

\subsection{Comparison On Standardized Benchmarks}
A very recent paper~\cite{schwarzschild2020just} introduced standardized benchmarks for backdoor and poisoning attacks. In particular, the benchmarks include the following attacks.

\begin{itemize}
    \item Clean-label poisoning attacks against transfer learning: FC~\cite{shafahi2018poison}, CP~\cite{zhu2019transferable}, and BP (our attack).
    \item Clean-label and hidden-trigger backdoor attacks: CLBD~\cite{turner2018clean}, and HTBD~\cite{saha2020hidden}.
    \item A from-scratch attack: Witches' Brew (WiB). Unlike transfer learning, this attack assumes that the victim trains a new, randomly initialized model on the poisoned dataset.
\end{itemize}

Our attack was included in this benchmark evaluation, as we had made a pre-print version of our work available on arXiv. In the following, we summarize and expand on these third-party results.


\mypar{Standardized Setup of Benchmarks.} For the benchmarks, poisoning attacks are always restricted to generate poison samples that remain within the \(l_{\infty}\)-ball of radius \texttt{\(\frac{8}{255}\)} centered at the corresponding base images.
On the other hand, backdoor attacks can use any \texttt{\(5 \times 5\)} patch.
Target and base images are chosen from the testing and training sets, respectively, according to a seeded, reproducible random assignment.
This allows the benchmarks to use the same choices for each attack and remove a source of variation from the results.
Each experiment uses 100 independent trials.
In general, two different training modes are considered: (i) linear transfer learning, and (ii) from-scratch training, where the victim's network is trained from random initialization on the poisoned dataset.

Unlike our experiments in Section ~\ref{sec:experiments}, the parameters of only one model are given to the attacker.
In linear transfer learning, the attacks are evaluated in white-box and black-box scenarios.
For white-box tests, the same frozen feature extractor that is given to the attacker is used for evaluation.
In black-box settings, the attacks are evaluated against unseen feature extractor networks.
Benchmarks can be divided into two sets of CIFAR-10 benchmarks and TinyImageNet benchmarks.

In CIFAR-10 benchmarks, for linear transfer learning, models are pre-trained on \textit{CIFAR-100}, and the fine-tuning is done on a subset of CIFAR-10, which has the first 250 images from each class, allowing for 25 poison samples.
The attacker has access to a ResNet-18~\cite{he2016deep} network, and the victim uses either (1) the same ResNet-18 network (white-box scenario) or (2) VGG11~\cite{simonyan2014very} and MobileNetV2~\cite{sandler2018mobilenetv2} networks (black-box scenario).
We extend the benchmarks here by considering a gray-box scenario, where the attacks are evaluated against a ResNet-18 network with unseen parameters. 
Furthermore, for the black-box setting, we evaluate the attacks against ResNet-34 and ResNet-50 networks~\cite{he2016deep} as well.
For these extra evaluations, we have used the poison samples that are shared by Schwarzschild~et~al.~\cite{schwarzschild2020just} in their GitHub repository,\footnote{Accessed Feb. 15 2021.} and here we report the detailed numbers.
When training from scratch, benchmarks use one of ResNet-18, VGG11, and MobileNetV2 networks, and report the average attack success rate.
For this mode, benchmarks use 500 poison samples.

In TinyImageNet benchmarks, for linear transfer learning, models are pre-trained on the first 100 classes of the TinyImageNet dataset~\cite{le2015tiny} and fine-tuned on the second half of the dataset, allowing for 250 poison samples.
The attacker has access to a VGG16 network, and black-box tests are done on ResNet-34 and MobileNetV2 networks.
For the from-scratch setting, the benchmarks are evaluated against a VGG16 model that is trained on the entire dataset with 250 poison samples.


\mypar{Results.} Table~\ref{table:benchmark} shows the success rates of the benchmarks.
In linear transfer learning, our attack outperformed other attacks by a significant margin, especially in white-box settings.
For example, in the TinyImageNet benchmark, \sysshortOne{} achieved an attack success rate of 100\%, while HTBD, CLBD, CP, and FC demonstrated success rates of 3\%, 3\%, 14\%, and 49\%, respectively.
In the gray-box setting, \sysshortOne{} showed only a modest improvement over other attacks.
For the black-box settings in CIFAR-10 benchmarks, \sysshortOne{} showed minimal improvement -- on average 1-2\% -- in comparison to other attacks.
In the black-box scenario of TinyImageNet benchmarks, \sysshortOne{} achieved an attack success rate of 10.5\%, while other attacks were below 2\%.
In general, \sysshortOne{} has shown a superior performance with respect to other contenders in the linear transfer learning mode.\footnote{WiB is not evaluated in the transfer learning mode, as it is not considered in the original work~\cite{geiping2020witches}.}
It is worth noting that backdoor attacks assume stronger threat models compared to poisoning attacks, as they need to manipulate both the training data and the target sample.

Before discussing the results in from-scratch training settings, we emphasize that \sysshortOne{} is designed to attack transfer learning scenarios.
Similar to FC and CP, \sysshortOne{} does not consider from-scratch training scenarios.
We expect the performance of \sysshortOne{} to drop in such a scenario, as the feature space is constantly being altered during training.
On the other hand, CLBD, and WiB attacks are specifically designed to target such scenarios.
On CIFAR-10 benchmarks,  all attacks succeeded less than 3\% of the time. The only exception is WiB, which achieves a success rate of 26\%.
However, in TinyImageNet benchmarks, interestingly, \sysshortOne{} demonstrated a success rate of 44\%, surpassing the runner-up attack (WiB) by 12\%.  
This shows that \sysshortOne{} has the capability to produce poison samples that even survive from-scratch training scenarios for the higher dimensional TinyImageNet dataset.

\section{Discussion}
\label{sec:discussion}
In Section~\ref{sec:eval-defenses}, we have evaluated \sys{} against defenses presented in a (concurrent) paper~\cite{peri2020deep}.
We found that the \knndef{} defense mitigates our attack completely if clean data points from the target's original class outnumber the poison samples.
However, such a defense still suffers from a poor precision rate, i.e., it removes a considerable number of clean samples, which, in turn, might have negative effects on the model performance.
We believe future defenses need to be proposed with higher precision rates.

In our experiments, we have noticed that \sys{} adds noticeable amounts of noise to the poison samples.
In fact, a recent study of clean-label poisoning attacks~\cite{schwarzschild2020just} acknowledges this limitation; poisoning attacks, which claim to be “clean label,” often produce easily visible image artifacts and distortions.
This study advocates using a perturbation budget \(\epsilon\) of 0.03. 
In Section~\ref{sec:attack-budget}, we observed that by using \(\epsilon\!=\!0.03\), our attack produces much fewer distortions, while still achieving an attack success rate of 33.0\% (see Figure~\ref{fig:poisons-target-eg-diffEpsilon} in the Appendix for the visual effect of \(\epsilon\) on poison examples).
In general, work in adversarial ML (in the image domain) suffers from the lack of a clear metric to determine what level of noise is imperceptible by the human eye.
Clean-label poisoning definitely benefits from additional research on this issue to produce less perceptible perturbations.

\section{Conclusions}
\label{sec:conclusion}
In this work, we present a scalable and transferable clean-label poisoning attack, \sys{}, for transfer learning.
\sys{} searches for poison samples that create, in the feature space, a convex polytope around the target image, ensuring that a linear classifier that trains on the poisoned dataset will classify the target into the poison class.
By driving the polytope center close to the target, \sys{} outperforms Convex Polytope---a state-of-the-art attack against transfer learning--- with success rate improvement of \threespImproveSingleTransfer{} and \threespImproveSingleEnd{} for \fix{} and \e2e{}, respectively.
At the same time, 
\sys{} achieves \spFasterCPGeneral{}x faster poison sample generation, which is crucial for enabling future research toward the development of reliable defenses.
Our evaluation of two neighborhood conformity defenses shows that \sys{} is more robust than Convex Polytope against less aggressive defense configurations.
As the number of poison samples increases, the \l2def{} defense becomes ineffective.
The \knndef{} defense also becomes vulnerable when poison samples outnumber the samples from the target's true class.
In general, both defenses demonstrated low detection precision, which indicates further research needs to be done to improve the precision of such defenses.  

\section*{Acknowledgments}
We would like to thank our reviewers for their valuable comments and input to improve our paper.

This material is based on research sponsored by DARPA under agreements FA8750-19-C-0003 and HR0011-18-C-0060, and by a gift from Intel Corp.
The U.S. Government is authorized to reproduce and distribute reprints for Governmental purposes notwithstanding any copyright notation thereon.
This research has also been sponsored by the Amazon Machine Learning Research Awards program and a GPU grant from Nvidia Corp. 
The views and conclusions contained herein are those of the authors and should not be interpreted as necessarily representing the official policies or endorsements, either expressed or implied, of DARPA, the U.S. Government, or the other sponsors.


\medskip
\small
\bibliographystyle{IEEEtranS}
\bibliography{refs}

\appendices

\section{Poison Visualization}
\begin{figure*}[t]
     \centering
     \includegraphics[width=0.8\textwidth]{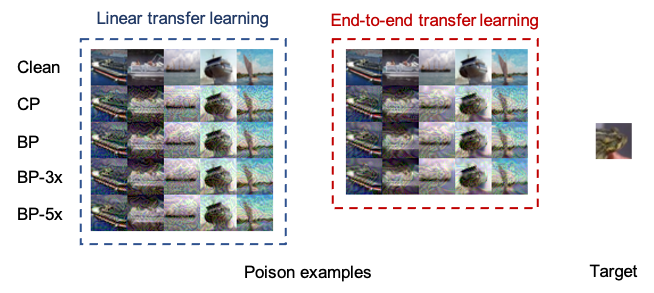}
     \caption{Poison samples crafted by Convex Polytope and \sys{} attacks. The first row shows the original images selected for crafting the poison samples.}
     \label{fig:poisons-target-eg}
\end{figure*}

\begin{figure*}[t]
     \centering
     \includegraphics[width=0.8\textwidth]{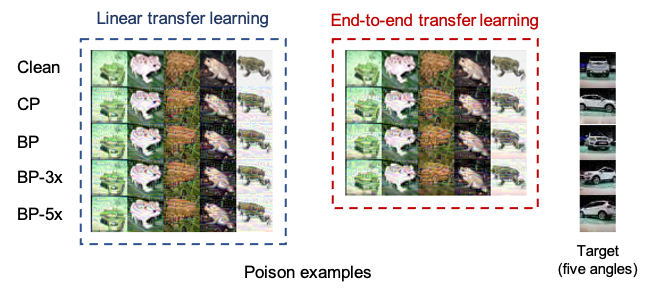}
     \caption{Poison samples crafted by Convex Polytope and \sys{} attacks in multi-target mode. The first row shows the original images selected for crafting the poison samples.}
     \label{fig:poisons-target-eg-multitarget}
\end{figure*}

Figure~\ref{fig:poisons-target-eg} depicts poison samples generated by Convex Polytope and \sys{} for one particular target.
The first row shows the original images that are selected for crafting the poison samples.
Figure!\ref{fig:poisons-target-eg-multitarget} depicts poison samples generated by Convex Polytope and \sys{} in multi-target mode, when multiple images of the target object (from different angles) are considered for crafting poison samples.
Note that we use the Multi-View Car Dataset~\cite{ozuysal2009pose} to select the target images.

\section{Coefficients Optimization Step in Convex Polytope}

As we discussed in Section~\ref{sec:bg}, Convex Polytope performs three steps in each iteration of the attack.
We observed that step one takes a significant amount of time compared to the other two steps.
Algorithm~\ref{alg:cvx-coeffs} shows the details of step one, which searches for the (most) suitable coefficients for the current poison samples at the time.
\begin{algorithm}
\caption{Convex Polytope - Coefficients Updating}
\label{alg:cvx-coeffs}
\begin{algorithmic}[1]
   \STATE {\bfseries Input:} A \(\leftarrow\) \phiPoisons{}
   \STATE \(\alpha \leftarrow \frac{1}{\norm{A^TA}}\)
   \FOR{$i=1$ {\bfseries to} $m$}
   \WHILE {\emph{not converged}}
   \STATE \(\widehat{\boldsymbol c}^{(i)} \leftarrow \boldsymbol c^{(i)} - \alpha A^T (A \boldsymbol c^{(i)} - \phi^{(i)}(\boldsymbol{x_t}))\)
   \IF {\( \textrm{loss}(\widehat{\boldsymbol c}^{(i)}) \geq \textrm{loss}(\boldsymbol c^{(i)}) \)}
   \STATE \(\alpha \leftarrow \frac{1}{\alpha}\)
   \ELSE
   \STATE \( \boldsymbol c^{(i)} \leftarrow \widehat{\boldsymbol c}^{(i)} \)
   \STATE project \(\boldsymbol c^{(i)}\) onto the probability simplex.
   \ENDIF
   \ENDWHILE
   \ENDFOR
\end{algorithmic}
\end{algorithm}

\section{\sys{} vs. Ensemble Feature Collision}
To evaluate Convex Polytope, Zhu et al.~\cite{zhu2019transferable} developed an ensemble version of Feature Collision~\cite{shafahi2018poison} to craft \emph{multiple} poison samples instead of one.
They further used this ensemble version as a benchmark.
The corresponding loss function is defined as:
\begin{align}
L_{FC}=\sum_{i=1}^{m}\sum_{j=1}^{k}\frac{\norm{\phi^{(i)}(x_p^{(j)}) - \phi^{(i)}(x_t)}^2}{\norm{\phi^{(i)}(x_t)}^2}.
\label{eq:fcLoss}
\end{align}
They argue that unlike Feature Collision, Convex Polytope's loss function (Eq.~\ref{eq:cvxLoss}) allows the poison samples to lie further away from the target. 
Experiments showed that Convex Polytope outperforms Feature Collision, especially in black-box settings.
It should be noted that, contrary to what is stated by Zhu et al.~\cite{zhu2019transferable}, the Ensemble Feature Collision attack objective described by Eq.~\ref{eq:fcLoss} is not a special case of Eq.~\ref{eq:cvxLoss} (when the coefficients are set to \(\frac{1}{k}\)), rather, it optimizes completely decoupled objectives for different poison samples.
While centering the target between poison samples allows for more flexibility in poison locations, Eq.~\ref{eq:fcLoss} pushes all poison samples close to the target, which has the same drawbacks of collision attacks, namely, perceptible patterns showing up in poison images and limited transferability.
By exploiting this approach of centering, we show that \sys{} improves both attack transferability and scalability.

\section{Detailed Results for Single-Target Mode}
\label{appendix:experimentDetails}


\subsection{End-to-End Transfer Learning}
\begin{figure*}
     \centering
     \begin{subfigure}[b]{0.45\textwidth}
         \centering
         \includegraphics[width=\textwidth]{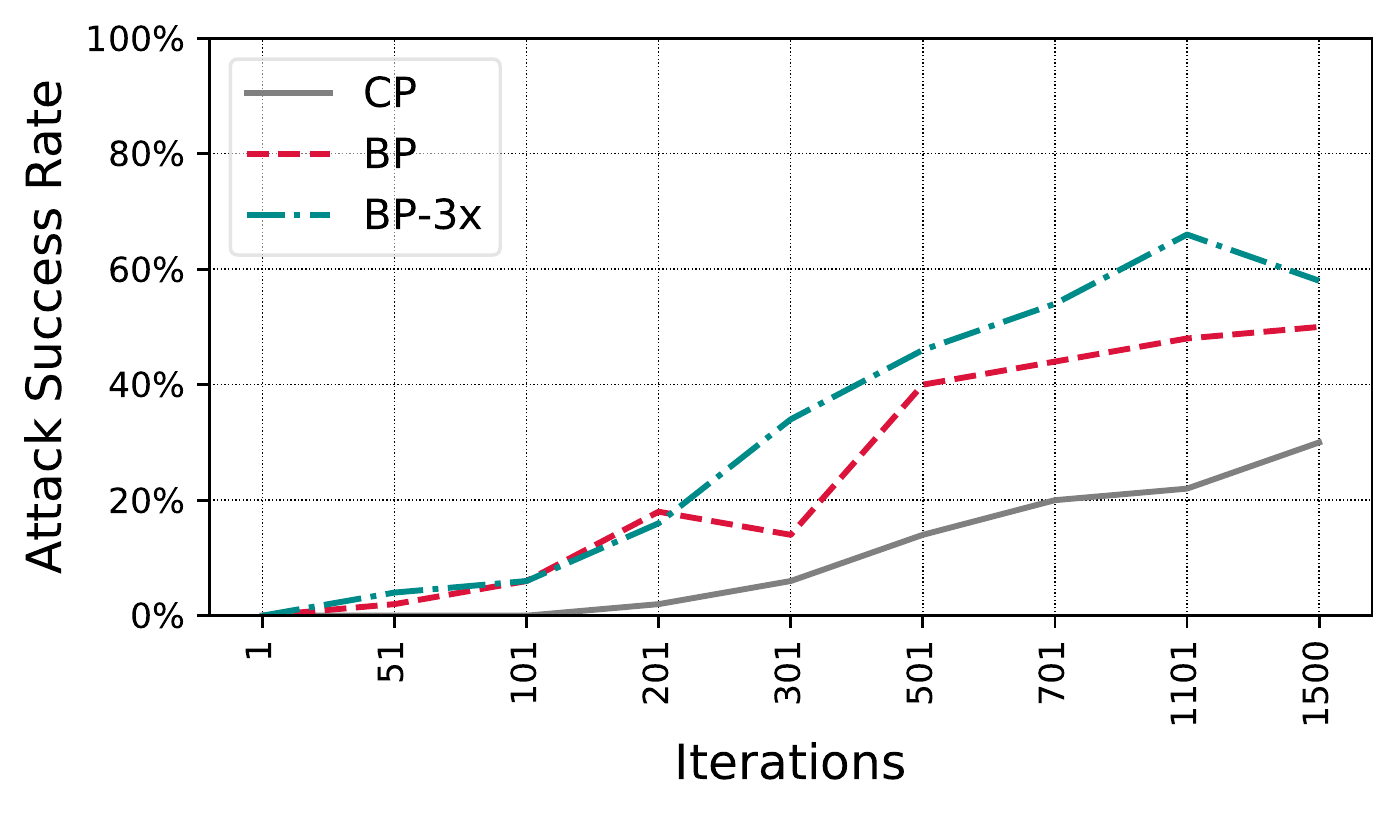}
         \caption{\scriptsize DPN92}
         \label{fig:single-end-DPN92}
     \end{subfigure}
     \hfill
     \begin{subfigure}[b]{0.45\textwidth}
         \centering
         \includegraphics[width=\textwidth]{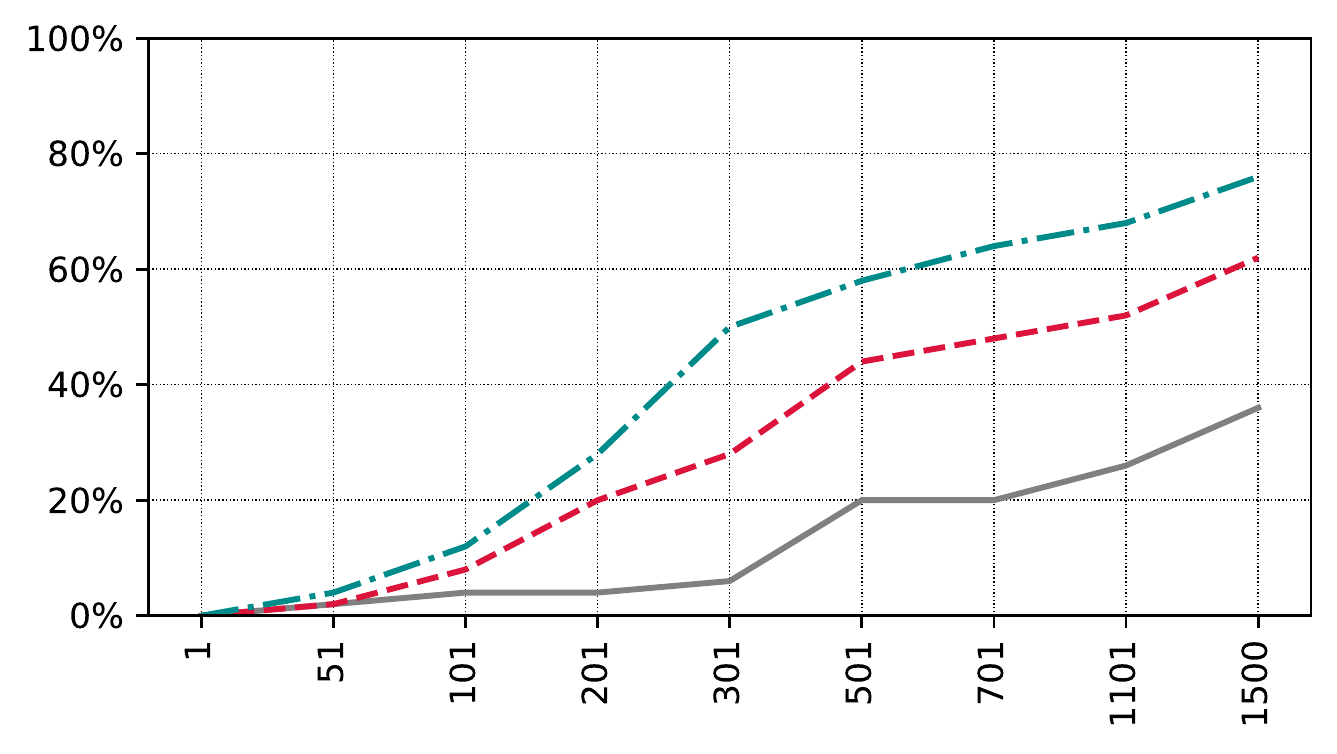}
         \caption{\scriptsize SENet18}
         \label{fig:single-end-SENet18}
     \end{subfigure}
     
     \begin{subfigure}[b]{0.45\textwidth}
         \centering
         \includegraphics[width=\textwidth]{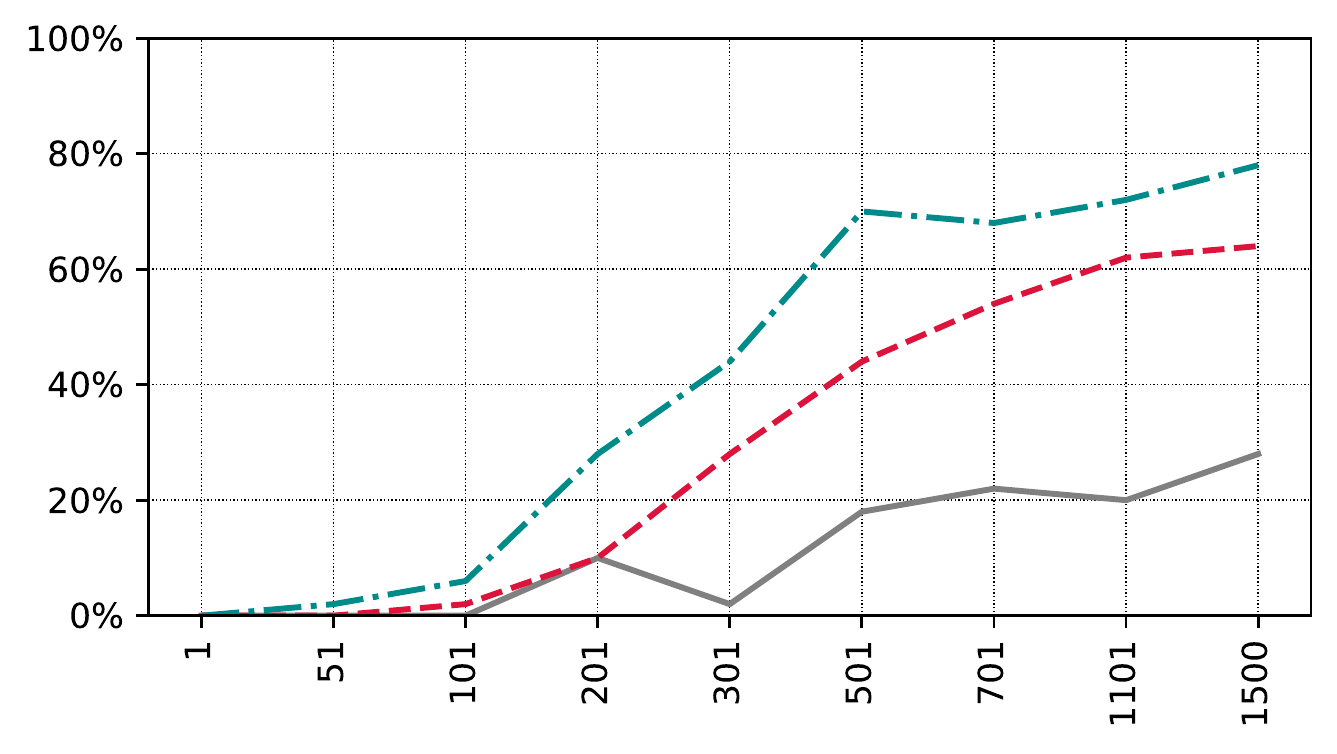}
         \caption{\scriptsize ResNet50}
         \label{fig:single-end-ResNet50}
     \end{subfigure}
     \hfill
     \begin{subfigure}[b]{0.45\textwidth}
         \centering
         \includegraphics[width=\textwidth]{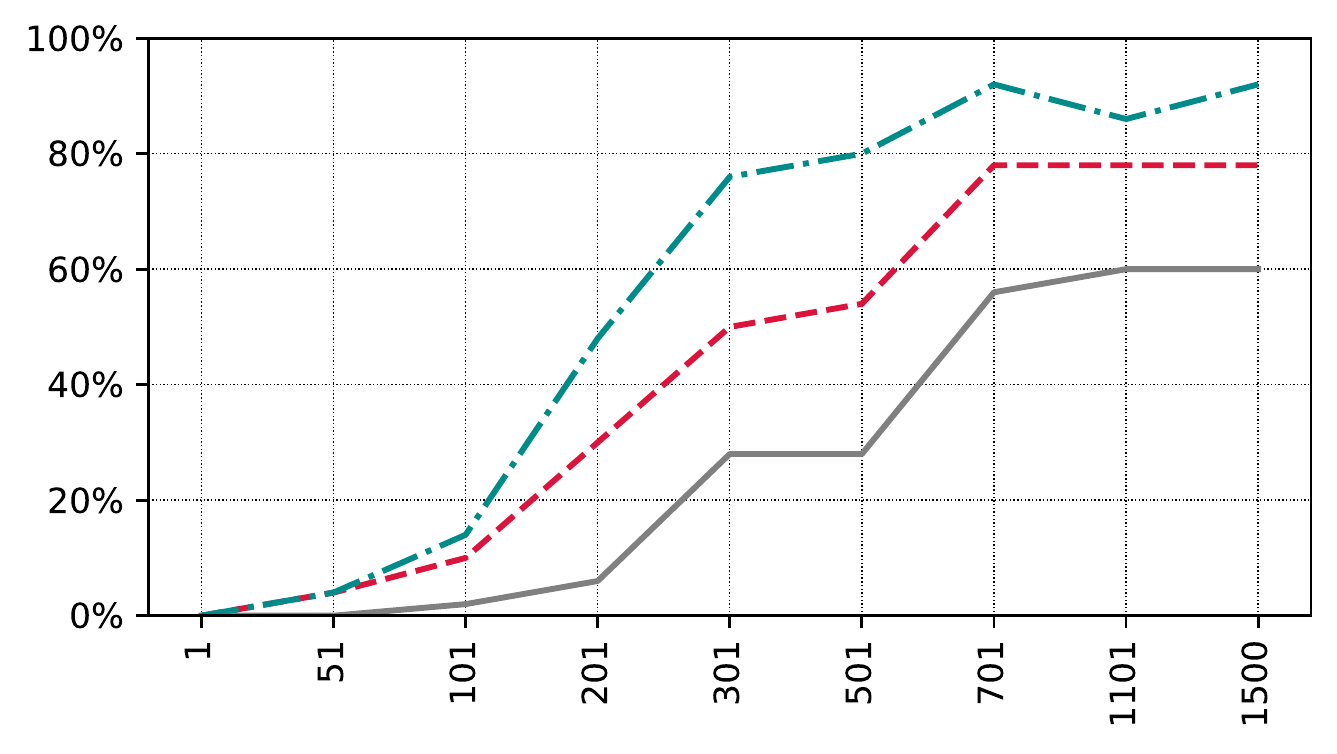}
         \caption{\scriptsize ResNeXt29\_2x64d}
         \label{fig:single-end-ResNeXt29_2x64d}
     \end{subfigure}
     
     \begin{subfigure}[b]{0.45\textwidth}
         \centering
         \includegraphics[width=\textwidth]{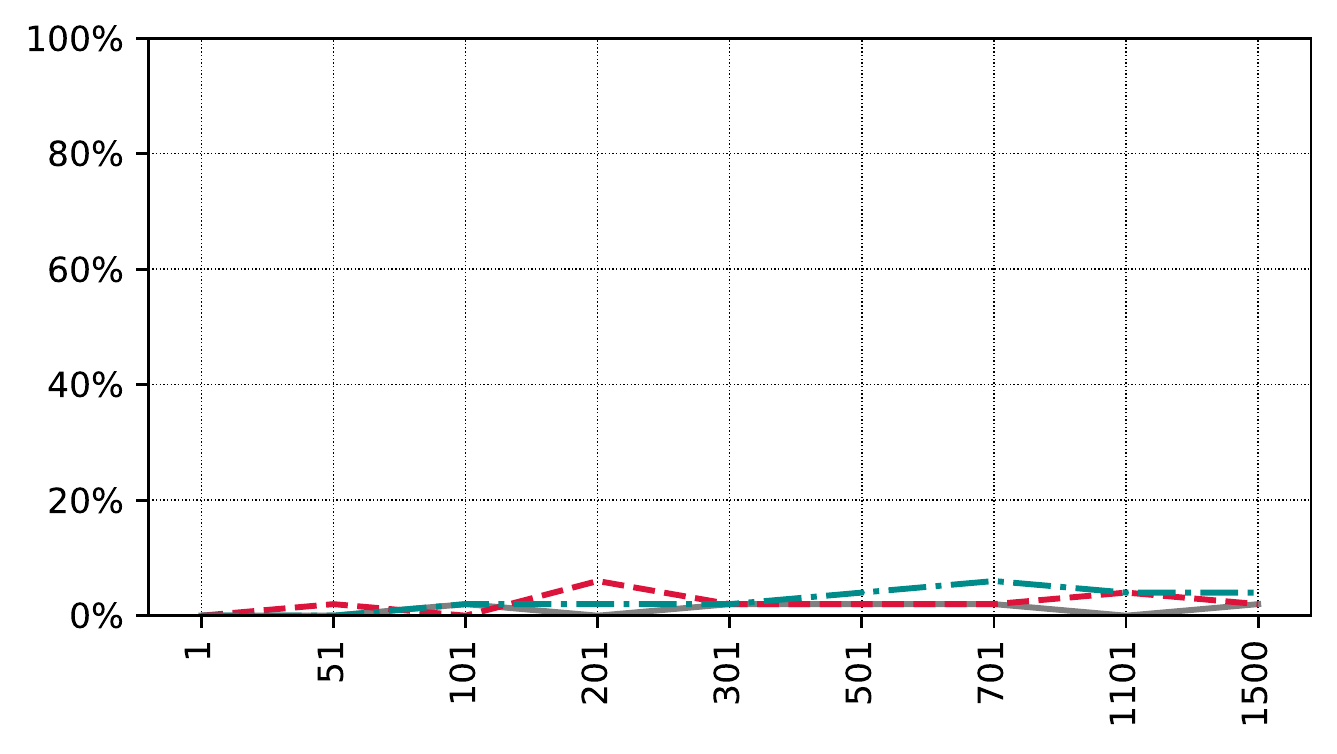}
         \caption{\scriptsize GoogLeNet}
         \label{fig:single-end-GoogLeNet}
     \end{subfigure}
     \hfill
     \begin{subfigure}[b]{0.45\textwidth}
         \centering
         \includegraphics[width=\textwidth]{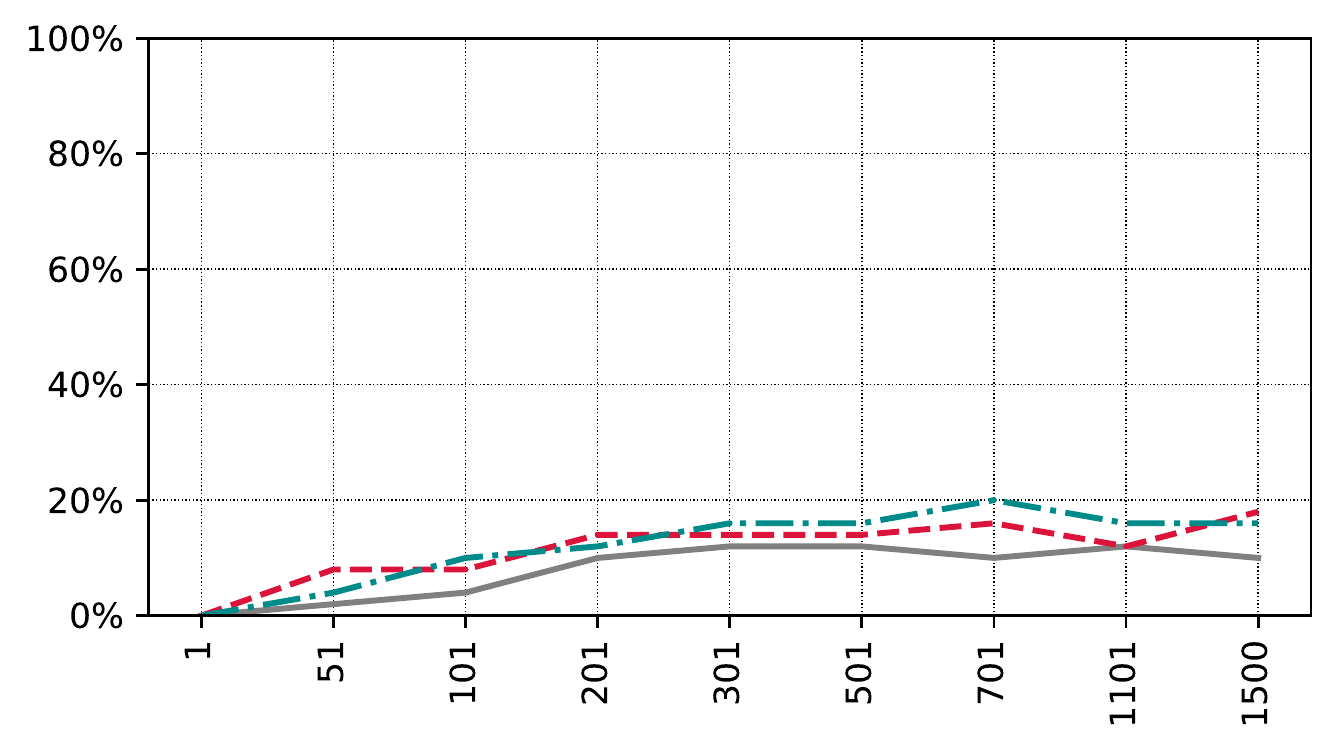}
         \caption{\scriptsize MobileNetV2}
         \label{fig:single-end-MobileNetV2}
     \end{subfigure}
     
     \begin{subfigure}[b]{0.45\textwidth}
         \centering
         \includegraphics[width=\textwidth]{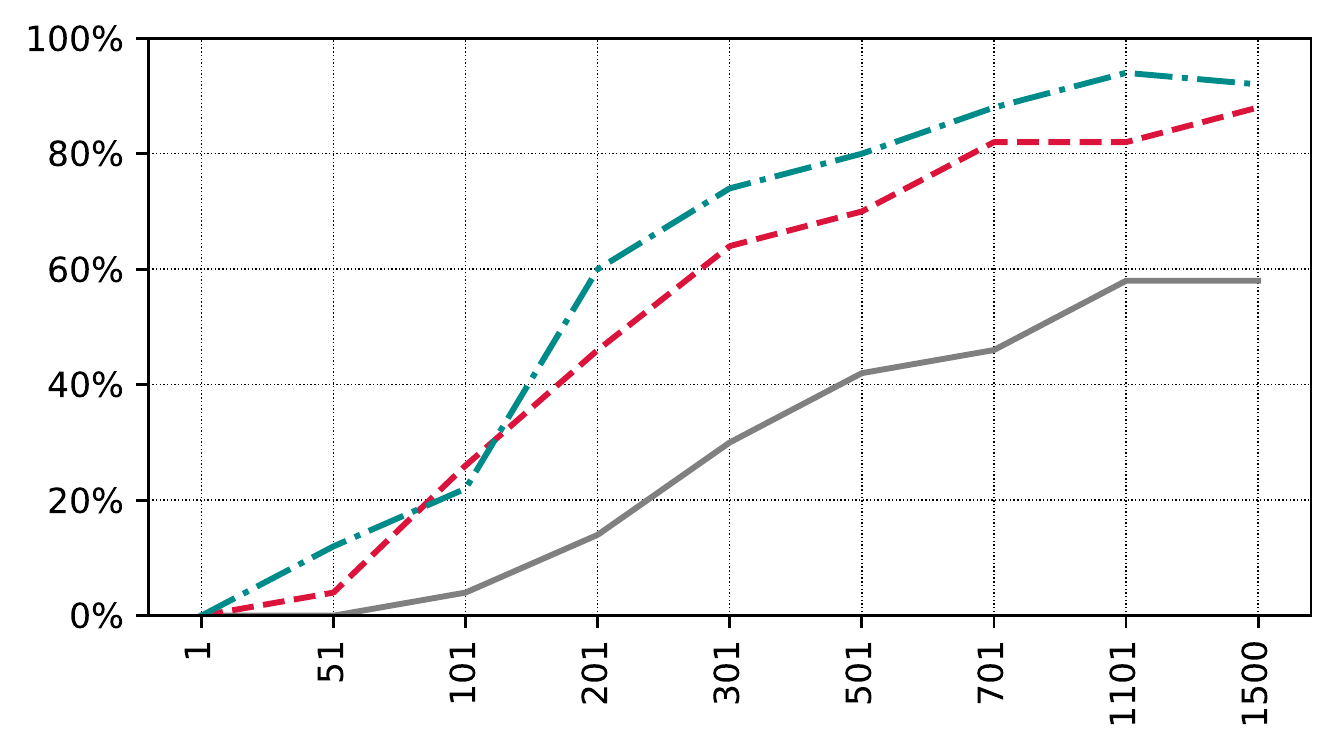}
         \caption{\scriptsize ResNet18}
         \label{fig:single-end-ResNet18}
     \end{subfigure}
     \hfill
     \begin{subfigure}[b]{0.45\textwidth}
         \centering
         \includegraphics[width=\textwidth]{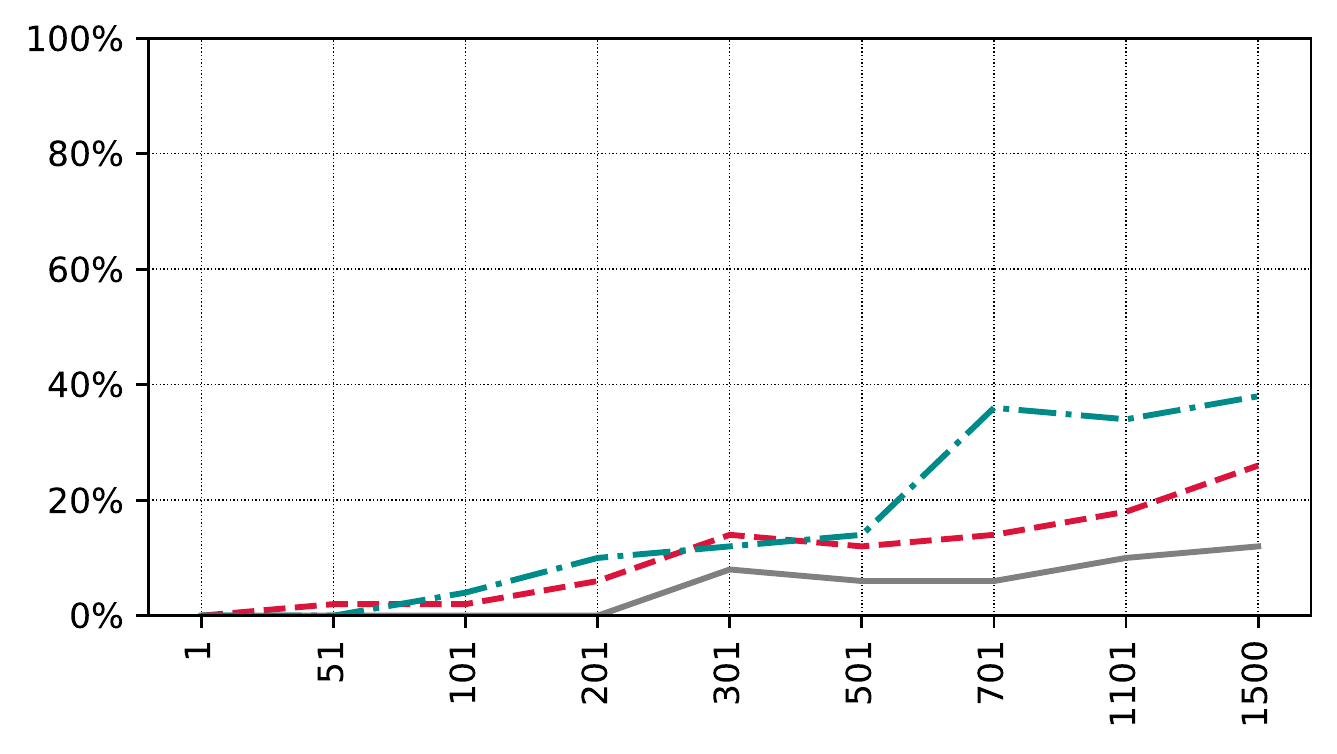}
         \caption{\scriptsize DenseNet121}
         \label{fig:single-end-DenseNet121}
     \end{subfigure}
     \caption{End-to-end transfer learning: Success rates of CP, \sysshortOne{}, \sysshortThree{}, and \sysshortFive{}, against each individual victim model. Notice \texttt{GoogLeNet}, \texttt{MobileNetV2}, \texttt{ResNet18} and \texttt{DenseNet121} are the black-box setting.}
     \label{fig:single-end-individualVictim}
\end{figure*}

Figure~\ref{fig:single-end-individualVictim} shows the attack success rates of CP, \sysshortOne{}, \sysshortThree{} and \sysshortFive{}, against each individual victim model when the victim employs \e2e{}.
Among them, the last row presents the black-box setting.
We note that none of CP, \sysshortOne{}, and \sysshortThree{} shows attack transferability for GoogLeNet.
Zhu et al.~\cite{zhu2019transferable} have made a similar observation.
They argued that since GoogLeNet has a more different architecture than the substitute models, it is more difficult for the ``attack zone'' to survive \e2e{}.

\subsection{Different Pairs of <original class, poison class>}
\begin{figure}
    	\centering
        \includegraphics[width=0.8\linewidth]{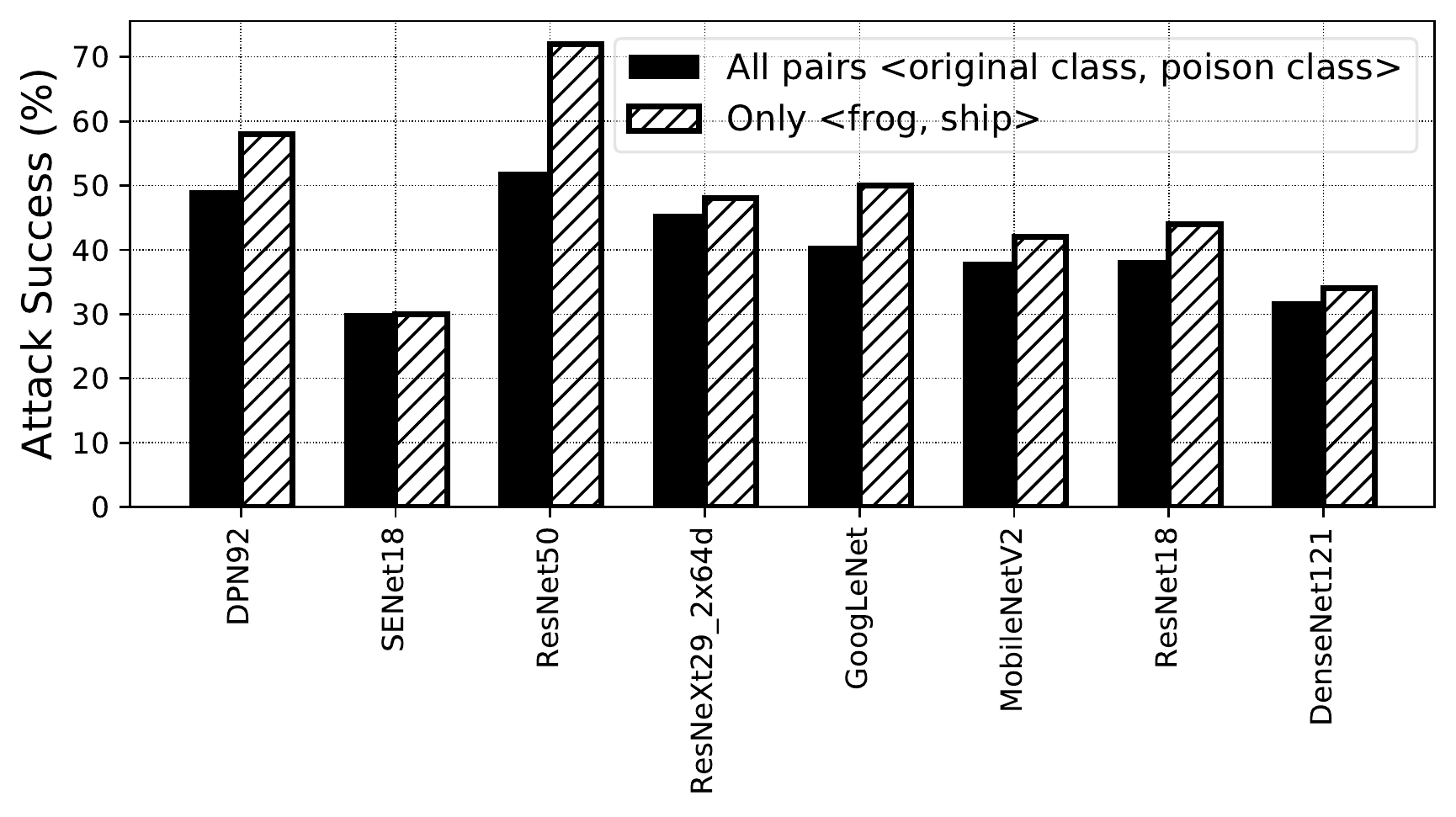}
		\caption{Attack success rates of \sysshortThree{} for all 90 pairs of <original class, poison class> in \fix{} as well as the original setting <frog, ship> (for 50 target images indexed from 4,851 to 4,900).}
		\label{fig:diffpairs-vs-frogship}
\end{figure}

To assess the effect of original and poison classes on the attack performance, we evaluate \sysshortThree{} for all 90 pairs of <original class,  poison class>; each with 5 different target images (indexed from 4,851 to 4,855 in the original class), resulting in a total of 450 attack instances.
We focus on linear transfer learning and limit each attack to 800 iterations to meet time and resource constraints.
Figure~\ref{fig:diffpairs-vs-frogship} shows the attack performance against individual victim networks in this setting as well as the original setting of <frog, ship>.
Table~\ref{table:diffpairs} shows the average attack performance for individual class pairs.
In particular, we have found the attack much less successful when our targeted misclassification is one of \texttt{airplane} or \texttt{deer} classes.

\begin{figure}[t]
     \centering
     \includegraphics[width=0.5\textwidth]{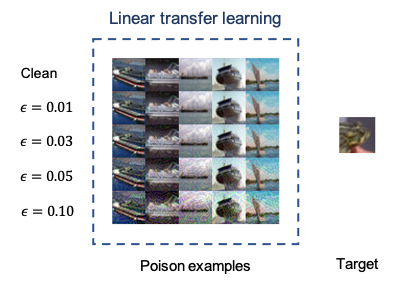}
     \caption{Poison samples crafted by \sys{} attacks in \fix{} using different values of \(\epsilon\).}
     \label{fig:poisons-target-eg-diffEpsilon}
\end{figure}

%

\begin{figure}
\vspace{-0.4cm}
\centering
	\begin{subfigure}[b]{0.4\textwidth}
    		\includegraphics[width=0.9\textwidth]{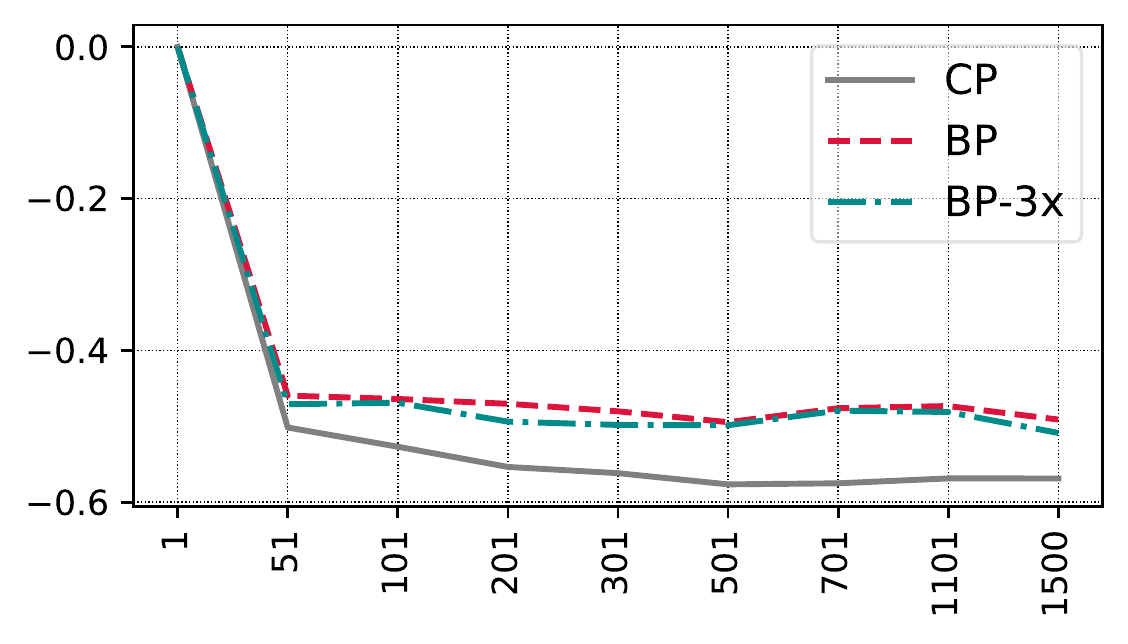}
    		\vspace{-0.3cm}
    		\caption{Zero overlap}
    		\label{fig:diff0-single-tr-meanVictim-cleantestacc}
    	\end{subfigure}
    	
    	\begin{subfigure}[b]{0.4\textwidth}
    		\centering
    		\includegraphics[width=0.9\textwidth]{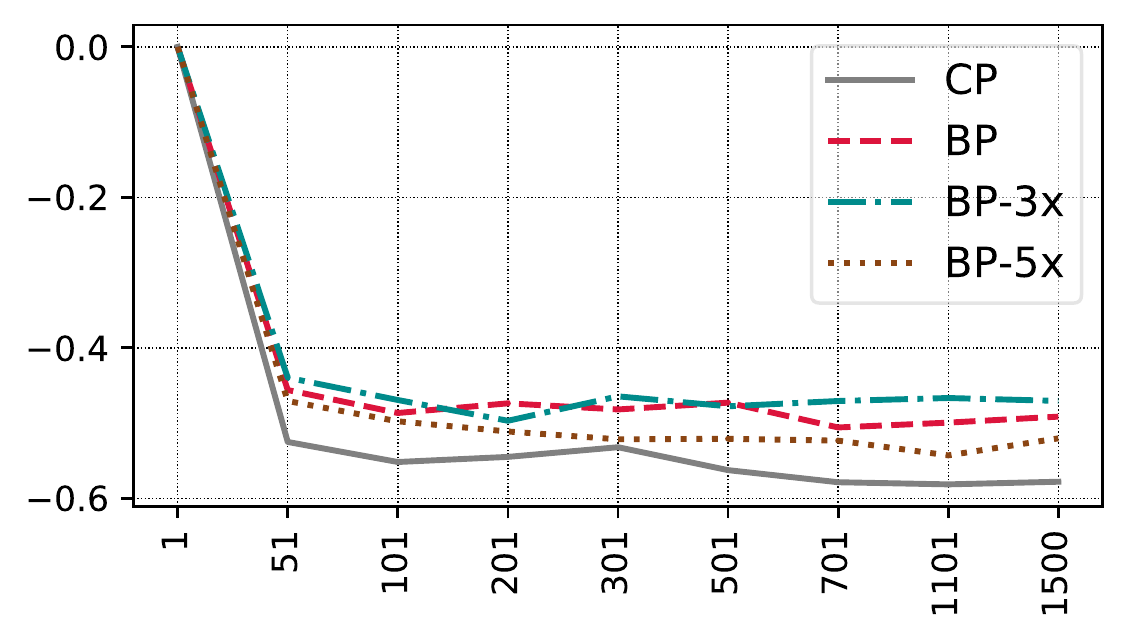}
    		\vspace{-0.3cm}
    		\caption{50\% overlap}
    		\label{fig:diff50-single-tr-meanVictim-cleantestacc}
    	\end{subfigure}
    	\vspace{-0.1cm}
    	\caption{Average variation in baseline test accuracy of models in linear transfer learning, when there is zero or 50\% overlap between training sets of the victim and substitute networks.}
    	\label{fig:diff-single-tr-meanVictim-cleantestacc}
    	\vspace{-0.3cm}
\end{figure}




\section{Implementation Details}
\label{appendix:implementationDetails}
The authors of Convex Polytope released the source code of CP along with the substitute networks. All models are trained with the same architecture and hyperparameters defined in \texttt{https://github.com/kuangliu/}, except for dropout.
We used their implementation directly for comparison.
For all experiments, we used \texttt{PyTorch-v1.3.1} over \texttt{Cuda 10.1}.
We ran all the attacks using \texttt{NVIDIA Titan RTX} graphics cards.
For solving Eq. 1 (Convex Polytope) and Eq. 3 (Bullseye Polytope), we used similar settings and parameters to what is practiced by Zhu et al.~\cite{zhu2019transferable}.

\mypar{Processing the Multi-View Car Dataset.} The resolutions of the Multi-View Car dataset are \texttt{376\(\times\)250}. To resize the images of this dataset to \texttt{32\(\times\)32} (the resolution of the CIFAR-10 images), we have used the \texttt{opencv-python} library.
While resizing the images, we achieved the best performance of the models on the Multi-View Car dataset using the \texttt{cv2.INTER\_AREA} interpolation.
It should be noted that the Multi-View Car dataset provides the exact location of the cars in the images.

\section{Defenses Against Evasion and Backdoor Attacks}
\label{appendix:defensesRelatedWork}
Most adversarial defenses are proposed for mitigating evasion attacks, where a targeted input is perturbed by imperceptible amounts during inference to enable misclassification.
Such perturbations are calculated using the gradients of the loss function on the victim network, or a set of surrogate networks if the victim network is unknown~\cite{goodfellow2014explaining, biggio2013evasion, szegedy2013intriguing}.
Many defenses against evasion attacks focus on obfuscating the gradients~\cite{athalye2018obfuscated}. 
They achieve this in several ways, e.g., introducing randomness during test time, or using non-differentiable layers.
Athalye et al.~\cite{athalye2018obfuscated} demonstrate that such defenses can be easily defeated by introducing techniques to circumvent the absence of gradient information, like replacing non-differentiable layers with approximation differentiable layers. 
Robust defenses to evasion attacks must avoid relying on obfuscated gradients and provide a ``smooth'' loss surface in the data manifold.
Variants of adversarial training~\cite{madry2017towards, shafahi2019adversarial, xie2019feature} and linearity or curvature regularizers~\cite{moosavi2019robustness, qin2019adversarial} are proposed to achieve this property. 
These defenses provide modest accuracy against strong multi-iteration PGD attacks~\cite{madry2017towards}. 
Papernot et al.~\cite{papernot2018deep} proposed the \knndef{} classifier, which combines the k-nearest neighbors algorithm with representations of the data learned by each layer of the neural network, as a way to detect outlier examples in feature space, with the hope that adversarial examples are the outliers.

Several defenses are proposed against backdoor attacks, primarily focusing on neighborhood conformity tests to sanitize the training data.
Steinhardt et al.~\cite{steinhardt2017certified} exploited variants of \l2def{} defense, where a data point is anomalous if it falls outside of a parameterized radius in feature space.
Chen et al.~\cite{chen2018detecting} employed feature clustering to detect and remove the poison samples, with the assumption that backdoor triggers will cause poison samples to cluster in feature space.

\section{Deep Sets}
\label{appendix:DeepSets}
One of the contributions of the "Deep Sets" paper is a characterization of all functions that take a set as input, which says that any such function f can be written as another function \(\rho\) of certain mean embedding \(\phi\) of the elements of the sets. 
We are instantiating this theorem in the following way: (1) The set input is the set of poison samples \(\{x^{(j)}_{p}\}_{j=1}^k\).
(2) \(\mathrm{f}\) is a prediction function: 
\begin{equation}
    \mathrm{f(\{x^{(j)}_{p}\}_{j=1}^k) = Predict(Train(X_C + \{x^{(j)}_{p}\}_{j=1}^k)}, x_t)\notag
\end{equation}
where \(\mathrm{Predict(h,x)}\) applies a classifier h to data point x. \(\mathrm{X_C}\) denotes the clean data. This is a set-function due to the permutation-invariant training procedure (e.g., Shuffle + SGD) that is typically adopted.
By the theorem, this function has an alternative representation \(\rho(\frac{1}{k}\sum_{j=1}^k\phi(x^{(j)}_{p}))\) that depends only on a certain mean embedding \(\phi\) of the poison samples.
For this reason, it motivates us to set the \(\{c^{(i)}\}\) in Eq.~\ref{eq:cvxLoss} to \(\frac{1}{k}\), which results in Eq.~\ref{eq:simplecvxLoss}. 
We acknowledge that this is not a formal theorem statement because the feature map \(\phi\) that we used might not be the same as the feature map that is required in applying Deep Sets theory, but given the flexibility of neural networks, we believe if we end-to-end optimize over \(\phi\) too, it is a reasonable approximation.

\begin{figure*}
     \centering
     \begin{subfigure}[b]{0.45\textwidth}
         \centering
         \includegraphics[width=\textwidth]{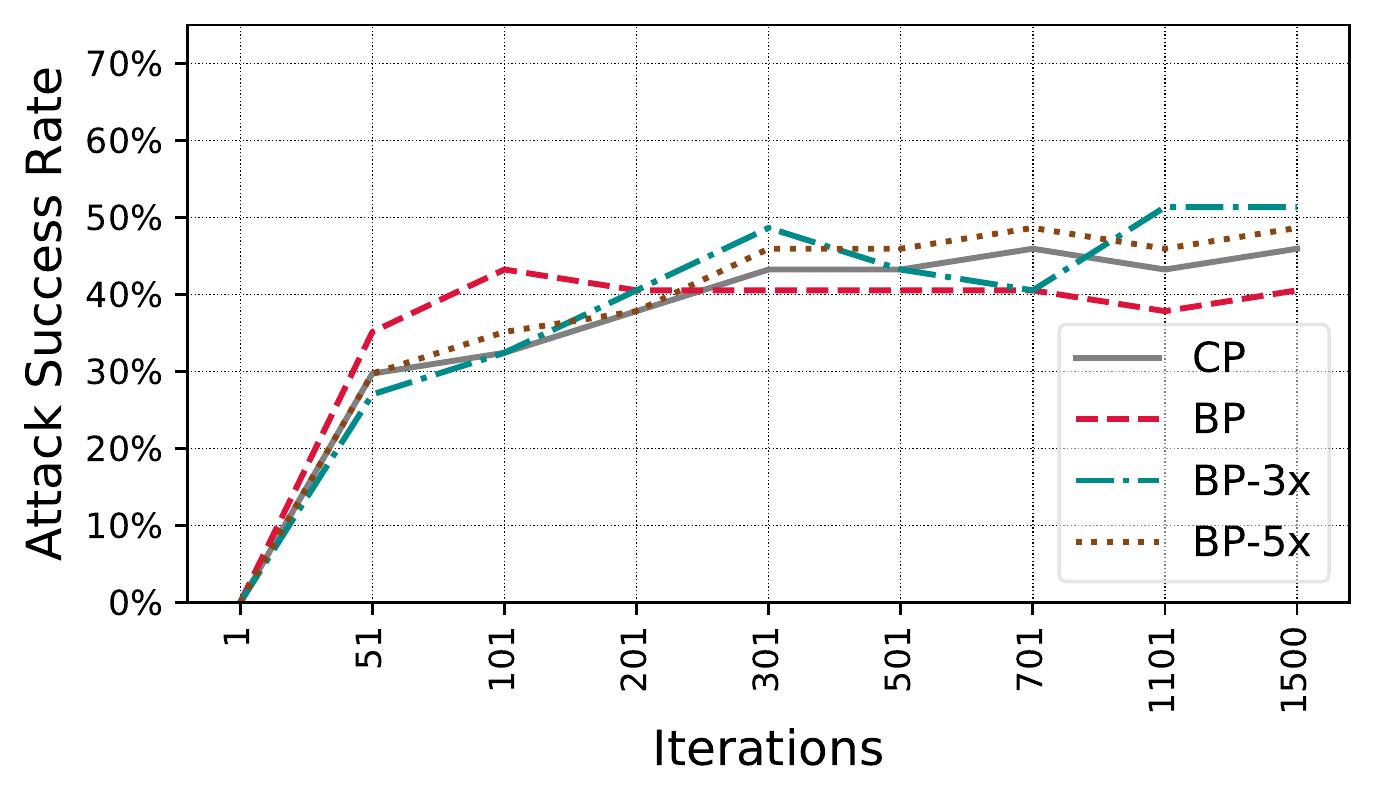}
         \caption{\scriptsize DPN92}
         \label{fig:difftraining-50-single-tr-DPN92}
     \end{subfigure}
     \hfill
     \begin{subfigure}[b]{0.45\textwidth}
         \centering
         \includegraphics[width=\textwidth]{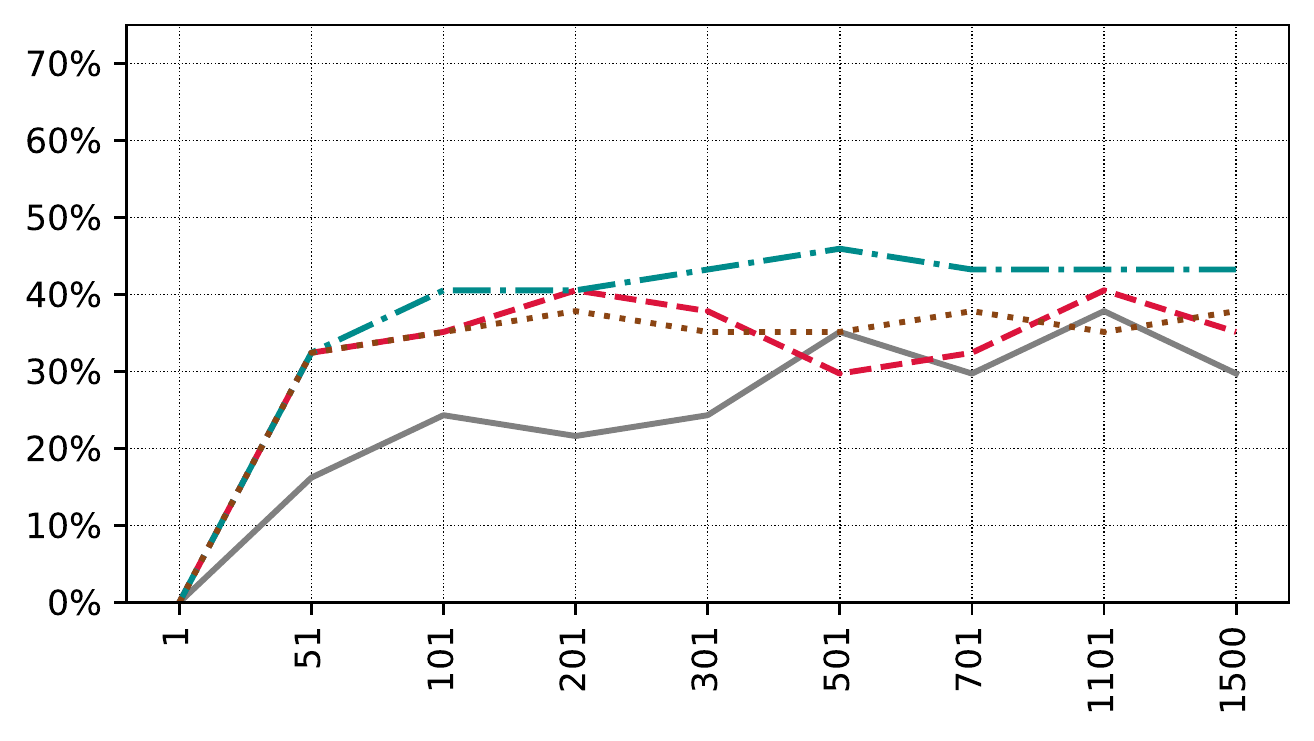}
         \caption{\scriptsize SENet18}
         \label{fig:difftraining-50-single-tr-SENet18}
     \end{subfigure}
     
     \begin{subfigure}[b]{0.45\textwidth}
         \centering
         \includegraphics[width=\textwidth]{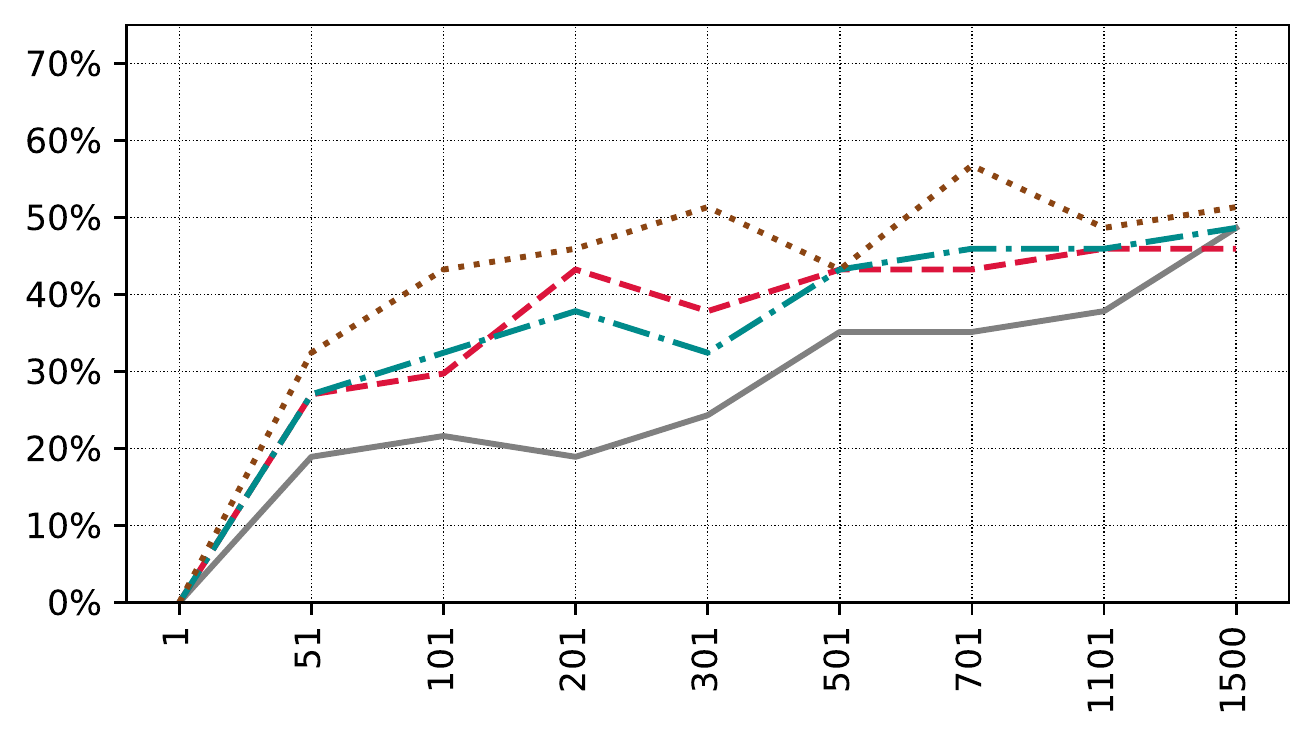}
         \caption{\scriptsize ResNet50}
         \label{fig:difftraining-50-single-tr-ResNet50}
     \end{subfigure}
     \hfill
     \begin{subfigure}[b]{0.45\textwidth}
         \centering
         \includegraphics[width=\textwidth]{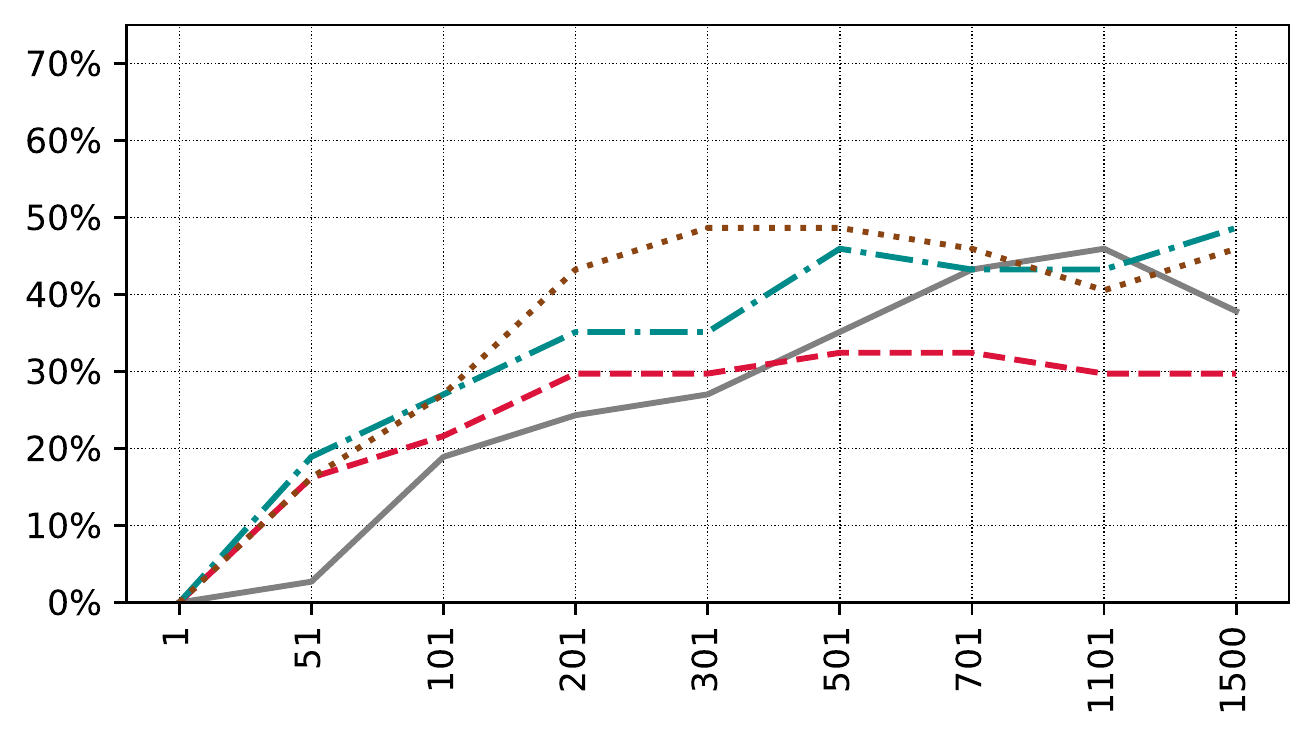}
         \caption{\scriptsize ResNeXt29\_2x64d}
         \label{fig:difftraining-50-single-tr-ResNeXt29_2x64d}
     \end{subfigure}
     
     \begin{subfigure}[b]{0.45\textwidth}
         \centering
         \includegraphics[width=\textwidth]{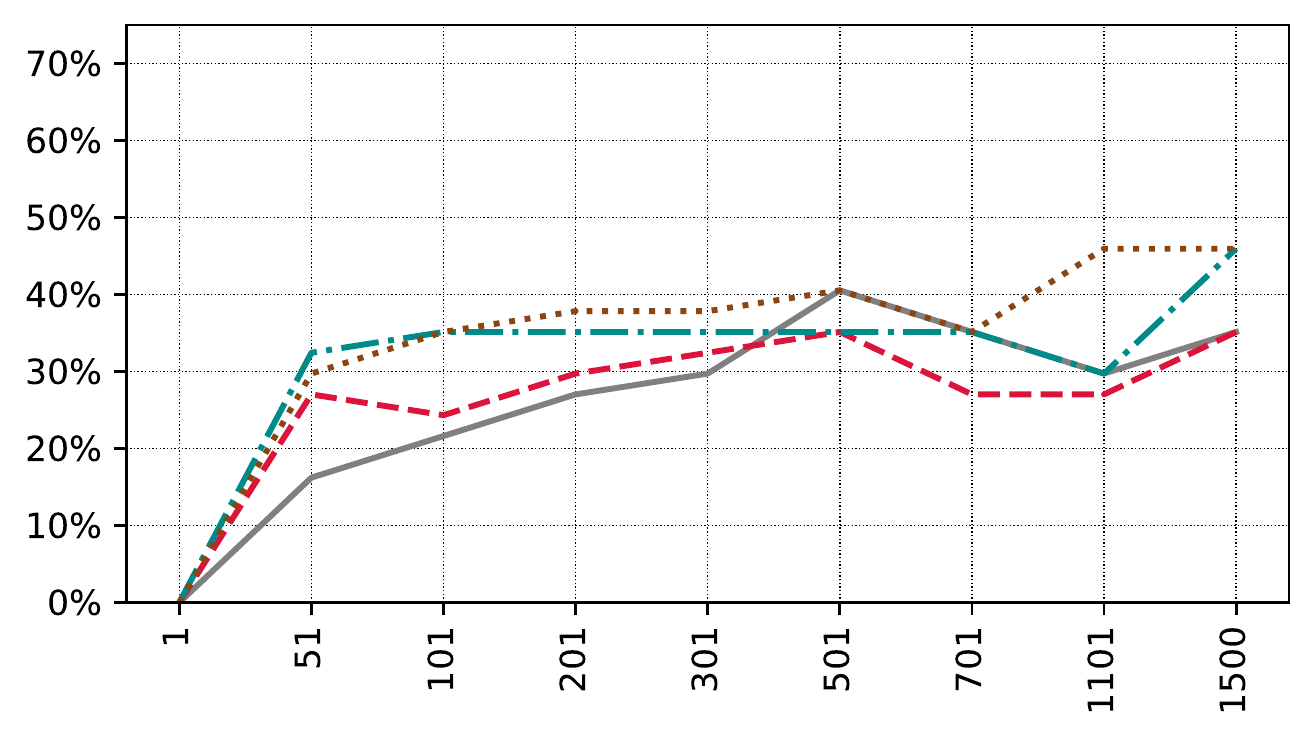}
         \caption{\scriptsize GoogLeNet}
         \label{fig:difftraining-50-single-tr-GoogLeNet}
     \end{subfigure}
     \hfill
     \begin{subfigure}[b]{0.45\textwidth}
         \centering
         \includegraphics[width=\textwidth]{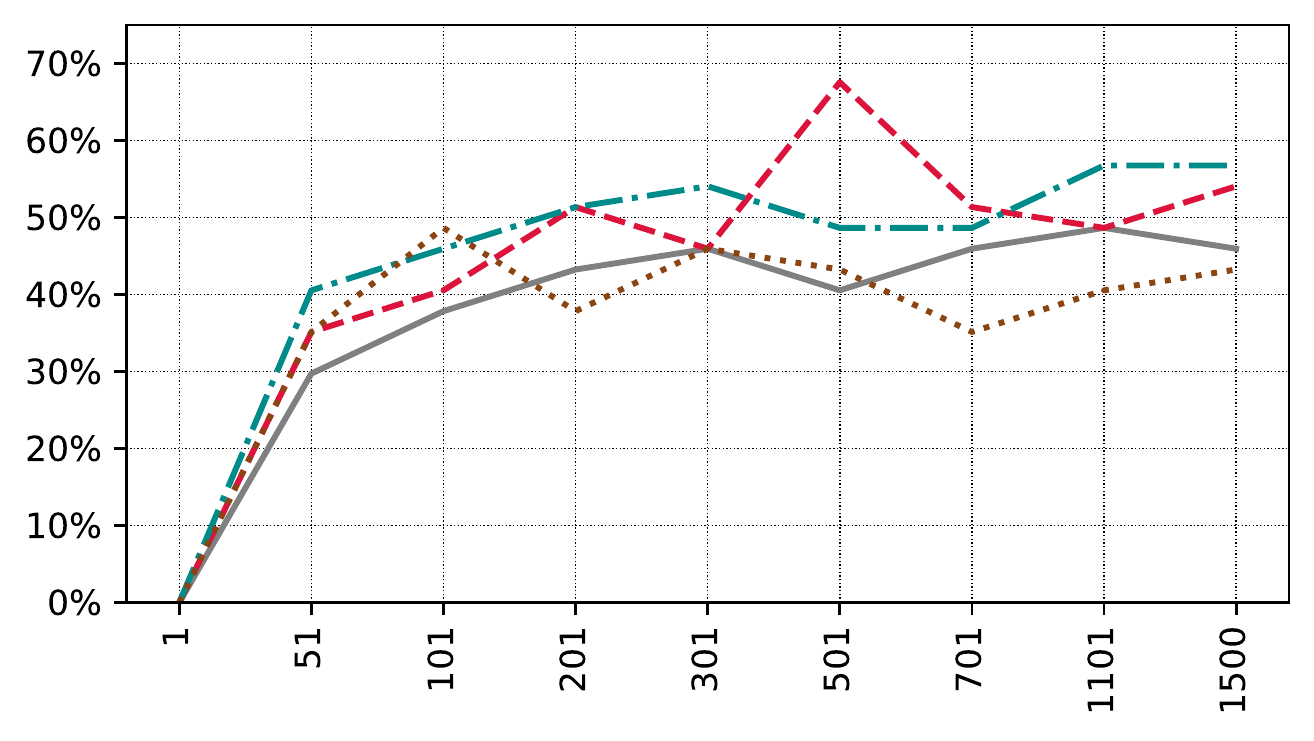}
         \caption{\scriptsize MobileNetV2}
         \label{fig:difftraining-50-single-tr-MobileNetV2}
     \end{subfigure}
     
     \begin{subfigure}[b]{0.45\textwidth}
         \centering
         \includegraphics[width=\textwidth]{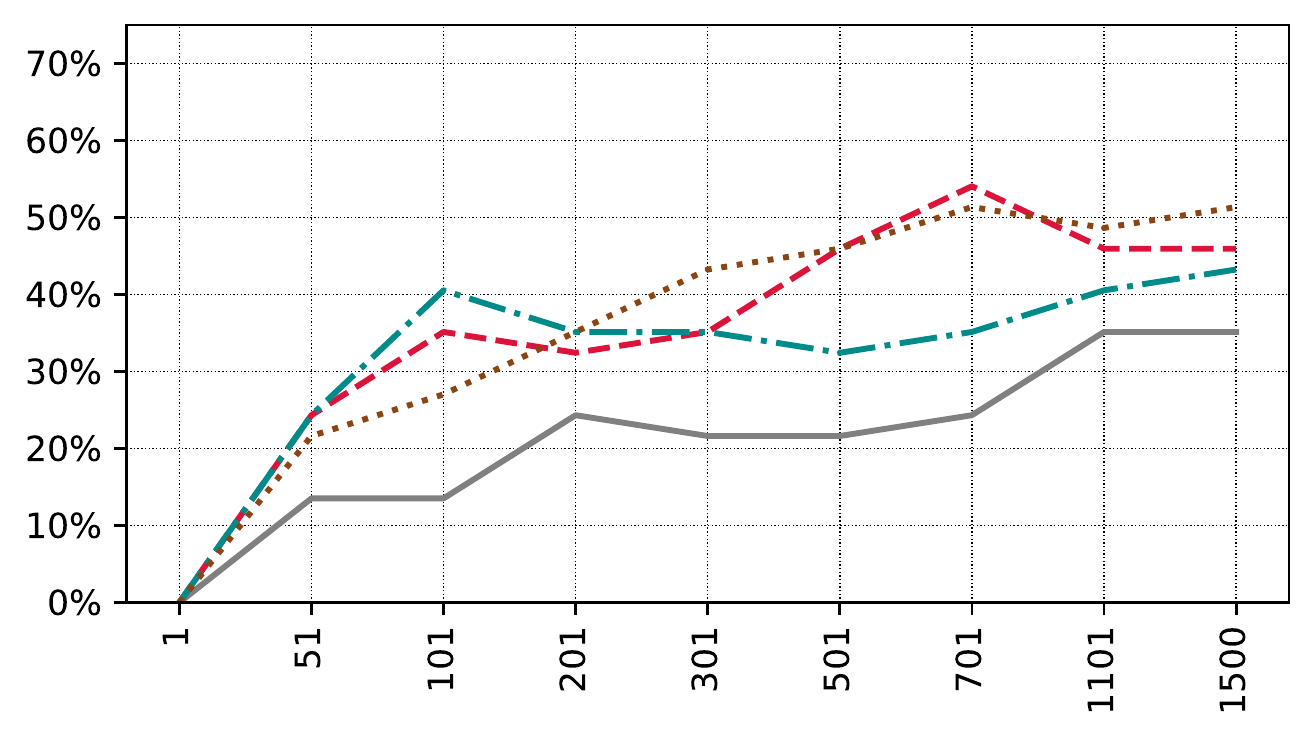}
         \caption{\scriptsize ResNet18}
         \label{fig:difftraining-50-single-tr-ResNet18}
     \end{subfigure}
     \hfill
     \begin{subfigure}[b]{0.45\textwidth}
         \centering
         \includegraphics[width=\textwidth]{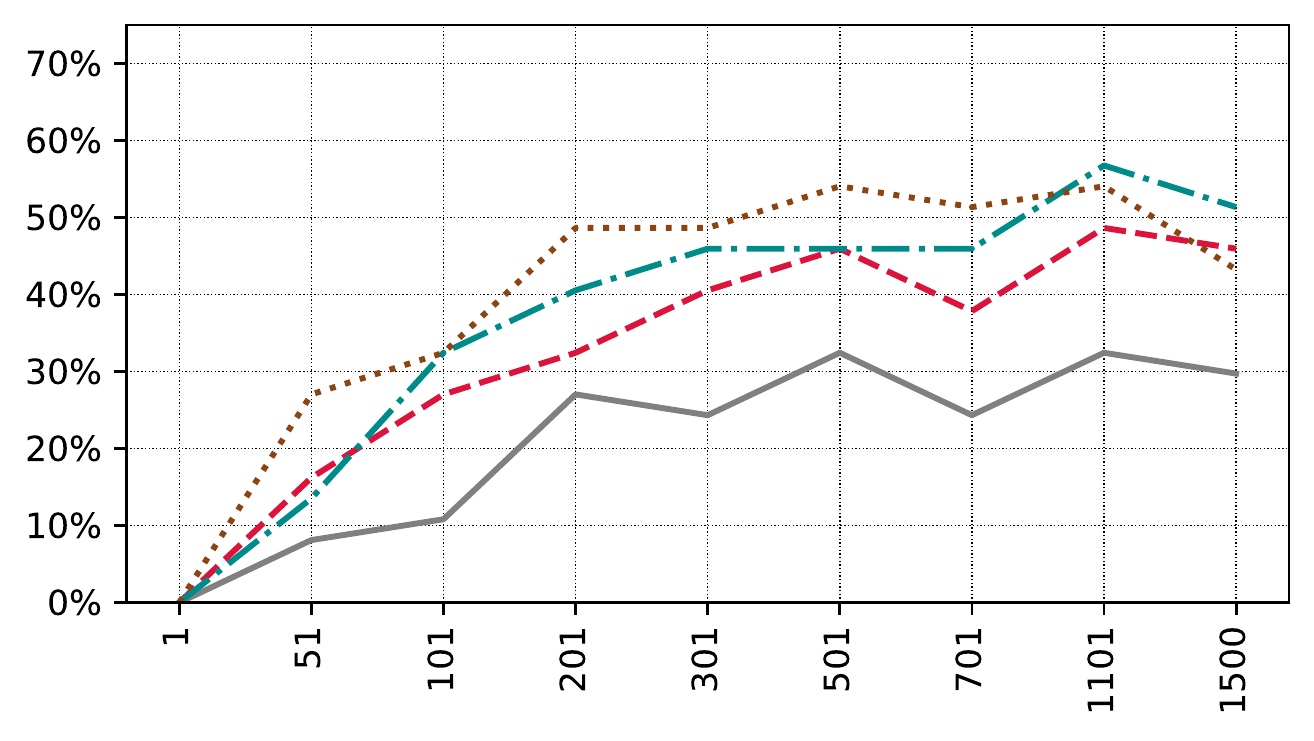}
         \caption{\scriptsize DenseNet121}
         \label{fig:difftraining-50-single-tr-DenseNet121}
     \end{subfigure}
     \caption{Linear transfer learning when we have 50\% overlap between the training sets of substitute and victim's networks: Success rates of CP, \sysshortOne{}, and \sysshortThree{}, against each individual victim model.}
     \label{fig:difftraining-50-single-tr-individualVictim}
\end{figure*}

\begin{table*}[]
    \centering
    \caption{Evaluation of \sysshortThree{} against linear transfer learning for individual class pairs. Attacks are limited to 800 iterations. Each individual pair is tested using five different target images. Having considered eight victim networks, in total, we evaluate each pair against the victim's network 40 times. Each cell shows the number of times that the attack succeeded for each pair.}
    \label{table:diffpairs}
    \begin{tabularx}{\textwidth}{ccccccccccccc}
        \multicolumn{2}{l}{\multirow{2}{*}{}} & \multicolumn{10}{c}{\textbf{Poison Class}} &  \\ \cline{3-12}
\multicolumn{1}{l}{} & \multicolumn{1}{l|}{} & \multicolumn{1}{c|}{\textbf{airplane}} & \multicolumn{1}{c|}{\textbf{automobile}} & \multicolumn{1}{c|}{\textbf{bird}} & \multicolumn{1}{c|}{\textbf{cat}} & \multicolumn{1}{c|}{\textbf{deer}} & \multicolumn{1}{c|}{\textbf{dog}} & \multicolumn{1}{c|}{\textbf{frog}} & \multicolumn{1}{c|}{\textbf{horse}} & \multicolumn{1}{c|}{\textbf{ship}} & \multicolumn{1}{c|}{\textbf{truck}} & \textbf{Total (of 360)} \\ \cline{2-13} 
\multicolumn{1}{c|}{\multirow{10}{*}{\textbf{Original Class}}} & \multicolumn{1}{c|}{\textbf{airplane}} & \multicolumn{1}{c|}{-} & \multicolumn{1}{c|}{14} & \multicolumn{1}{c|}{13} & \multicolumn{1}{c|}{17} & \multicolumn{1}{c|}{9} & \multicolumn{1}{c|}{18} & \multicolumn{1}{c|}{17} & \multicolumn{1}{c|}{20} & \multicolumn{1}{c|}{16} & \multicolumn{1}{c|}{24} & 148 \\ \cline{2-13} 
\multicolumn{1}{c|}{} & \multicolumn{1}{c|}{\textbf{automobile}} & \multicolumn{1}{c|}{7} & \multicolumn{1}{c|}{-} & \multicolumn{1}{c|}{17} & \multicolumn{1}{c|}{19} & \multicolumn{1}{c|}{12} & \multicolumn{1}{c|}{15} & \multicolumn{1}{c|}{20} & \multicolumn{1}{c|}{16} & \multicolumn{1}{c|}{25} & \multicolumn{1}{c|}{24} & 155 \\ \cline{2-13} 
\multicolumn{1}{c|}{} & \multicolumn{1}{c|}{\textbf{bird}} & \multicolumn{1}{c|}{6} & \multicolumn{1}{c|}{13} & \multicolumn{1}{c|}{-} & \multicolumn{1}{c|}{24} & \multicolumn{1}{c|}{7} & \multicolumn{1}{c|}{14} & \multicolumn{1}{c|}{20} & \multicolumn{1}{c|}{22} & \multicolumn{1}{c|}{20} & \multicolumn{1}{c|}{22} & 148 \\ \cline{2-13} 
\multicolumn{1}{c|}{} & \multicolumn{1}{c|}{\textbf{cat}} & \multicolumn{1}{c|}{6} & \multicolumn{1}{c|}{9} & \multicolumn{1}{c|}{11} & \multicolumn{1}{c|}{-} & \multicolumn{1}{c|}{10} & \multicolumn{1}{c|}{18} & \multicolumn{1}{c|}{15} & \multicolumn{1}{c|}{13} & \multicolumn{1}{c|}{14} & \multicolumn{1}{c|}{22} & 118 \\ \cline{2-13} 
\multicolumn{1}{c|}{} & \multicolumn{1}{c|}{\textbf{deer}} & \multicolumn{1}{c|}{6} & \multicolumn{1}{c|}{17} & \multicolumn{1}{c|}{15} & \multicolumn{1}{c|}{24} & \multicolumn{1}{c|}{-} & \multicolumn{1}{c|}{15} & \multicolumn{1}{c|}{17} & \multicolumn{1}{c|}{24} & \multicolumn{1}{c|}{17} & \multicolumn{1}{c|}{23} & 158 \\ \cline{2-13} 
\multicolumn{1}{c|}{} & \multicolumn{1}{c|}{\textbf{dog}} & \multicolumn{1}{c|}{8} & \multicolumn{1}{c|}{15} & \multicolumn{1}{c|}{13} & \multicolumn{1}{c|}{31} & \multicolumn{1}{c|}{7} & \multicolumn{1}{c|}{-} & \multicolumn{1}{c|}{17} & \multicolumn{1}{c|}{16} & \multicolumn{1}{c|}{15} & \multicolumn{1}{c|}{22} & 144 \\ \cline{2-13} 
\multicolumn{1}{c|}{} & \multicolumn{1}{c|}{\textbf{frog}} & \multicolumn{1}{c|}{5} & \multicolumn{1}{c|}{17} & \multicolumn{1}{c|}{16} & \multicolumn{1}{c|}{20} & \multicolumn{1}{c|}{10} & \multicolumn{1}{c|}{15} & \multicolumn{1}{c|}{-} & \multicolumn{1}{c|}{15} & \multicolumn{1}{c|}{35} & \multicolumn{1}{c|}{23} & 156 \\ \cline{2-13} 
\multicolumn{1}{c|}{} & \multicolumn{1}{c|}{\textbf{horse}} & \multicolumn{1}{c|}{10} & \multicolumn{1}{c|}{11} & \multicolumn{1}{c|}{14} & \multicolumn{1}{c|}{22} & \multicolumn{1}{c|}{12} & \multicolumn{1}{c|}{17} & \multicolumn{1}{c|}{23} & \multicolumn{1}{c|}{-} & \multicolumn{1}{c|}{19} & \multicolumn{1}{c|}{26} & 154 \\ \cline{2-13} 
\multicolumn{1}{c|}{} & \multicolumn{1}{c|}{\textbf{ship}} & \multicolumn{1}{c|}{5} & \multicolumn{1}{c|}{20} & \multicolumn{1}{c|}{17} & \multicolumn{1}{c|}{24} & \multicolumn{1}{c|}{9} & \multicolumn{1}{c|}{19} & \multicolumn{1}{c|}{0} & \multicolumn{1}{c|}{18} & \multicolumn{1}{c|}{-} & \multicolumn{1}{c|}{23} & 135 \\ \cline{2-13} 
\multicolumn{1}{c|}{} & \multicolumn{1}{c|}{\textbf{truck}} & \multicolumn{1}{c|}{6} & \multicolumn{1}{c|}{16} & \multicolumn{1}{c|}{13} & \multicolumn{1}{c|}{23} & \multicolumn{1}{c|}{12} & \multicolumn{1}{c|}{18} & \multicolumn{1}{c|}{20} & \multicolumn{1}{c|}{23} & \multicolumn{1}{c|}{23} & \multicolumn{1}{c|}{-} & 154 \\ \cline{2-13} 
 & \multicolumn{1}{c|}{\textbf{Total (of 360)}} & \multicolumn{1}{c|}{\textbf{59}} & \multicolumn{1}{c|}{142} & \multicolumn{1}{c|}{129} & \multicolumn{1}{c|}{204} & \multicolumn{1}{c|}{\textbf{88}} & \multicolumn{1}{c|}{149} & \multicolumn{1}{c|}{149} & \multicolumn{1}{c|}{167} & \multicolumn{1}{c|}{184} & \multicolumn{1}{c|}{209} & 
    \end{tabularx}
\end{table*}

\end{document}